\documentclass[11pt]{article}

\usepackage[final]{acl}

\usepackage{times}
\usepackage{latexsym}
\usepackage{tablefootnote}
\usepackage[table]{xcolor}
\usepackage{tikz}

\usepackage[T1]{fontenc}

\usepackage[utf8]{inputenc}

\usepackage{microtype}

\usepackage{inconsolata}

\usepackage{graphicx}

\usepackage{pifont}
\usepackage{hyperref}
\usepackage{url}
\usepackage{amsmath}
\usepackage{booktabs} 
\usepackage{adjustbox}
\usepackage{multirow} 
\usepackage{arydshln}
\usepackage{listings}
\usepackage{xcolor}
\usepackage{cuted}
\usepackage{capt-of}

\newcommand{\metric}{XQDT}
\newcommand{\metricw}{XQDT$_{W}$}
\newcommand{\metrice}{XQDT$_{E}$}
\newcommand{\cmark}{\ding{51}}
\newcommand{\xmark}{\ding{55}}
\title{\metric: eXplainable and Quantitative Data-Text \\Alignment Metric with Feedback Signals}

\newcommand{\bad}[1]{\textcolor{red!70!black}{#1}}
\newcommand{\fix}[1]{\textcolor{green!50!black}{#1}}

\newcommand{\softbox}[1]{%
\fcolorbox{black}{gray!8}{%
\parbox[t]{\dimexpr\linewidth-2\fboxsep-2\fboxrule\relax}{\raggedright #1}}}

\author{Kun Efimov-Zhang, Yifei Song, Claire Gardent\\
        CNRS/LORIA and Université de Lorraine\\
  \texttt{kun.zhang@inria.fr} \quad
  \texttt{\{yifei.song,claire.gardent\}@loria.fr}
}

\begin{document}
\maketitle
\begin{abstract}
Evaluating data-text alignment remains challenging: existing metrics often provide limited explanations for the scores, while prompt-based LLM-as-Judge methods can be expensive and unreliable. We present an end-to-end explainable evaluation metric that fine-tunes a language model to identify omitted, extra, incorrect, and correct data units in a data-text pair. These local judgements are aggregated into precision, recall, and F1 scores, providing both fine-grained diagnostic feedback and an interpretable measure of alignment quality. Across benchmarks, our fine-tuned models outperform LLM-as-Judge methods in error prediction and achieve competitive precision, recall, and F1 scores, while maintaining strong correlation with human judgements. Beyond evaluation, our verifier outputs also provide useful feedback signals for downstream correction and refinement, supporting alignment-oriented improvement of data-to-text and text-to-data. Code and resources are available at
\url{https://github.com/guihuzhang/xqdt}.%

\end{abstract}

\section{Introduction}
\label{sec:context}
\begin{figure*}[t]
\centering
\small
\begin{tikzpicture}[
    node distance=0.15cm,
    box/.style={rectangle, draw, align=left, font=\footnotesize, inner sep=4pt},
    label/.style={font=\small\bfseries},
    correct/.style={fill=green!15},
    incorrect/.style={fill=orange!15},
    omitted/.style={fill=purple!15},
    extra/.style={fill=blue!15}
]

\node[label, anchor=north west] (input_label) at (0, 0) {Input};

\node[box, text width=0.31\textwidth, anchor=north west, minimum height=5.2cm] (input_box) at (0, -0.4) {
    {\sc Data} ${D}${\sc :}\\
    1. $\langle$Marie Curie, birth place, Warsaw$\rangle$\\
    2. $\langle$Marie Curie, discovered, radium$\rangle$\\
    3. $\langle$Marie Curie, birth year, 1867$\rangle$\\
    4. $\langle$Marie Curie, field, radioactivity$\rangle$\\
    5. $\langle$Marie Curie, death year, 1934$\rangle$\\[0.5cm]
    
    {\sc Text} ${T}${\sc :}\\
    ``{Marie Curie was born in} {Krakow} in {1867}. {After moving to Paris}, {she conducted research in radioactivity and discovered radium} {with her husband Pierre Curie}.''
};

\node[label, anchor=north west] (pred_label) at (0.35\textwidth, 0) {Model Prediction};

\node[box, text width=0.39\textwidth, anchor=north west, minimum height=5.2cm] (pred_box) at (0.35\textwidth, -0.4) {};

\node[incorrect, text width=0.37\textwidth, anchor=north west, font=\footnotesize, inner sep=3pt] at (0.36\textwidth, -0.5) {
    \textcolor{red}{{\sc [Incorrect]}} Data item 1:\\
    Warsaw $\rightarrow$ Krakow (birth place error)
};

\node[correct, text width=0.37\textwidth, anchor=north west, font=\footnotesize, inner sep=3pt] at (0.36\textwidth, -1.3) {
    \textcolor{green!50!black}{{\sc [Correct]}} Data item 2:
    discovered radium
};

\node[correct, text width=0.37\textwidth, anchor=north west, font=\footnotesize, inner sep=3pt] at (0.36\textwidth, -1.7) {
    \textcolor{green!50!black}{{\sc [Correct]}} Data item 3:
    birth year 1867
};

\node[correct, text width=0.37\textwidth, anchor=north west, font=\footnotesize, inner sep=3pt] at (0.36\textwidth, -2.1) {
    \textcolor{green!50!black}{{\sc [Correct]}} Data item 4:
    field radioactivity
};

\node[omitted, text width=0.37\textwidth, anchor=north west, font=\footnotesize, inner sep=3pt] at (0.36\textwidth, -2.5) {
    \textcolor{violet}{{\sc [Omitted]}} Data item 5: 
    $\langle$Marie Curie, death year, 1934$\rangle$ is not in text
};

\node[extra, text width=0.37\textwidth, anchor=north west, font=\footnotesize, inner sep=3pt] at (0.36\textwidth, -3.3) {
    \textcolor{blue}{{\sc [Extra]}} Text content: 
    $\langle$Marie Curie, moved to, Paris$\rangle$ is not in data
};

\node[extra, text width=0.37\textwidth, anchor=north west, font=\footnotesize, inner sep=3pt] at (0.36\textwidth, -4.1) {
    \textcolor{blue}{{\sc [Extra]}} Text content: 
    $\langle$Marie Curie, husband,  Pierre Curie$\rangle$ is not in data
};

\node[label, anchor=north west] (metrics_label) at (0.78\textwidth, 0) {Metrics};

\node[box, text width=0.205\textwidth, anchor=north west, fill=yellow!10, minimum height=5.2cm] (metrics_box) at (0.78\textwidth, -0.4) {
    {\sc Error Counts:}\\
    \textcolor{green!50!black}{$|\text{Correct}| = 3$}\\
    \textcolor{red}{$|\text{Incorrect}| = 1$}\\
    \textcolor{violet}{$|\text{Omitted}| = 1$}\\
    \textcolor{blue}{$|\text{Extra}| = 2$}\\ [0.1cm]
    
    $\text{Precision}_{\text{D2T}} = \text{Recall}_{\text{T2D}}$\\
    $\dfrac{\textcolor{green!50!black}{3}}{\textcolor{green!50!black}{3}+\textcolor{blue}{2}+\textcolor{red}{1}} = {0.50}$\\ [0.2cm]
    
    $\text{Recall}_{\text{D2T}} = \text{Precision}_{\text{T2D}}$\\
    $\dfrac{\textcolor{green!50!black}{3}}{\textcolor{green!50!black}{3}+\textcolor{violet}{1}+\textcolor{red}{1}} = {0.60}$\\ [0.2cm]
    
    $\text{F1}_{\text{D2T}} = \text{F1}_{\text{T2D}}$\\
    $2 \times \frac{0.50 \times 0.60}{0.50 + 0.60} = {0.55}$
};

\draw[-latex, thick] (input_box.east) -- (pred_box.west);
\draw[-latex, thick] (pred_box.east) -- ([yshift=0cm]metrics_box.west);

\end{tikzpicture}
\caption{\textbf{Example of our fine-grained explainable evaluation for data-text alignment.} The model assigns labels (Correct, Incorrect, Omitted, Extra) to each input data unit and identifies extra content units in the text. These fine-grained local judgments are then aggregated into precision, recall, and F1 scores as shown in Equations~\ref{eq:precision_d2t_recall_t2d}, \ref{eq:recall_d2t_precision_t2d}. %
}
\label{fig:example}
\end{figure*}

Reliable bidirectional evaluation of data-text alignment is essential in data-to-text and text-to-data tasks, where structured data and text should faithfully express the same information. We propose \metric\ (E\textbf{X}plainable and \textbf{Q}uantitative \textbf{D}ata-\textbf{T}ext Alignment), a novel metric for explainable evaluation of consistency between structured data \(D\) and English text \(T\). We assume that \(D\) is composed of atomic data items, and that \(T\) is either a verbalisation of \(D\) (for data-to-text) or a text from which \(D\) is derived (for text-to-data).

Unlike overlap-based or reference-based metrics~\citep{papineni-etal-2002-bleu,banerjee-lavie-2005-meteor}, classification-based~\citep{dusek-kasner-2020-evaluating,zhang-etal-2023-factspotter} and regression-based metrics~\citep{martinez-etal-2025-semantic}, which ultimately produce only scores, or methods that rely on multiple expensive LLM calls to generate explanations~\citep{wang-etal-2023-chatgpt,xu-etal-2023-instructscore,huidrom-etal-2025-assessing}, our approach is both \textit{explainable} and \textit{quantitative}. It predicts explicit labels (omitted, extra, incorrect, correct) for individual data units, and aggregates these local predictions into precision, recall, and F1 scores. 

Figure~\ref{fig:example} illustrates the output of \metric\ on an example data-text pair. By grounding evaluation in the structured data \(D\), each input unit can be directly inspected to determine whether it is correctly verbalised, omitted, or incorrectly realised, while unsupported text content is separately identified as {\sc Extra}. The resulting predictions are explicit and localised, and can also be reused as signals for downstream correction and refinement.

We make three contributions:

\begin{itemize}
    \item We introduce \metric, a metric that provides both fine-grained explanations of data-text mismatches and quantitative scores in terms of precision, recall, and F1.
    \item We evaluate \metric\ on ground-truth error labels and against human annotations across multiple datasets (KELM, WebNLG, and E2E) and settings, including synthetic data, model outputs, and manually validated noisy data-text pairs.
    \item We show that the verifier outputs produced by \metric\ can be used as feedback signals for downstream correction and refinement for both data-to-text and text-to-data.
\end{itemize}

\begin{table}[h]
\centering
\small
\setlength{\tabcolsep}{4.5pt}
\begin{adjustbox}{width=\columnwidth}
\begin{tabular}{lccc|cc}
\toprule
& \multicolumn{3}{c}{Localisation} & \multicolumn{2}{c}{Scores} \\
\cmidrule(lr){2-4}\cmidrule(lr){5-6}
Method & Omitted & Extra & Incorrect & Precision & Recall \\
\midrule
NLI    & $\triangle$ & \xmark & $\triangle$ & \cmark & \cmark \\
DQE    & $\triangle$ & $\triangle$ & $\triangle$ & \cmark & \cmark \\
FS     & $\triangle$ & \xmark & $\triangle$ & \xmark & \cmark \\
MonoLR & \xmark & \xmark & \xmark & \cmark & \cmark \\
XQDT   & \cmark & \cmark & \cmark & \cmark & \cmark \\
\bottomrule
\end{tabular}
\end{adjustbox}
\caption{\textbf{Comparison of metrics in terms of scoring and localisation capabilities.} $\triangle$ denotes indirect localisation signals without explicit unit-aligned labels.}
\label{tab:metric-feature-comparison}
\end{table}

\section{Related Work}
\label{section:related}

\paragraph{Metrics for Data-Text Alignment.}
Prior work evaluates data-text alignment with NLI-based entailment \citep{dusek-kasner-2020-evaluating}, question generation and answering (DQE, \citealp{rebuffel-etal-2021-data}), triple-level recall-oriented scoring (FactSpotter, \citealp{zhang-etal-2023-factspotter}), and direct regression of precision, recall, and F1 (MonoLR, \citealp{martinez-etal-2025-semantic}). DQE is effective but relies on multi-step question generation and answering, FactSpotter is better suited to recall than to unsupported content, and MonoLR predicts aggregate scores without explicit unit-level localisation. Recent work also explores LLM-based evaluation, including prompted judges, distillation of LLM feedback into learned metrics, and meta-evaluation of judge protocols \citep{wang-etal-2023-chatgpt,xu-etal-2023-instructscore,huidrom-etal-2025-assessing}, while separate evaluations identify limitations in LLM compositional relation
reasoning \citep{zhao2024large}. As summarised in Table~\ref{tab:metric-feature-comparison}, our method differs in jointly providing unit-level localisation of omitted, extra, and incorrect content with precision, recall, and F1 estimates.

Atomic verification also underlies long-form factuality metrics.
FActScore \citep{min-etal-2023-factscore} decomposes generations into atomic facts and verifies each fact
against a knowledge source, while SAFE \citep{wei2024longform} combines atomic decomposition with
search-augmented verification. Structured discourse representations further enrich atomic facts with discourse
relations for interpretable cross-text factual consistency verification
\citep{zhang-etal-2025-structured}. 
XQDT instead addresses bidirectional alignment with a given structured input:
Correct, Omitted, and Incorrect units are anchored in the input, whereas Extra
units must be discovered and localised from the text because they are absent
from the input.

\paragraph{Feedback for Data-to-Text and Text-to-Data.}
Alignment signals can also be used as feedback to improve both data-to-text and text-to-data systems. Prior work studies iterative refinement with LLM feedback \citep{madaan2023selfrefine,gero2023selfverification}, fine-grained step-level
preference optimisation \citep{zhao-etal-2026-step-pruning}, omission-oriented recovery for text-to-data generation \citep{han-etal-2024-pive}, and preference- or reward-based optimisation for structured prediction and data-to-text generation \citep{li-etal-2025-generating,song-gardent-2025-mucal}. These results suggest that a verifier that localises omitted, extra, and incorrect content can provide practical, fine-grained  supervision signals beyond evaluation.

\section{Method}
\label{sec:label_assignment}

\subsection{Fine-Grained Explainable Evaluation} 
\label{subsec:labels}

As illustrated in Figure~\ref{fig:example}, given a data-text pair $(D, T)$, our model outputs the differences between the two by predicting triples annotated with a label indicating: whether a triple in $D$ is correctly verbalised in $T$ (Correct); whether it is in $D$ but incorrectly verbalised in $T$ (Incorrect); whether it is in $D$ but not verbalised by $T$ (Omitted); or whether a triple expressed in $T$ is absent from $D$ (Extra). Correct, Omitted, and Incorrect units are anchored in $D$, whereas Extra units must be recovered from $T$ because no corresponding input unit exists. See Appendix~\ref{app:labelsdef} for detailed label definition.

\subsection{Quantifying Recall, Precision, and F1} Quantitative scores are obtained by aggregating the labels on each data unit into {\sc precision, recall,} and {\sc F1} scores as follows.%
\begin{align}
    \text{P}%
    =  \frac{|\text{Correct}|}{|\text{Correct}| + |\text{Extra}| + |\text{Incorrect}|}
    \label{eq:precision_d2t_recall_t2d} \\
    \text{R}%
    = \frac{|\text{Correct}|}{|\text{Correct}| + |\text{Omit}| + |\text{Incorrect}|}
    \label{eq:recall_d2t_precision_t2d}
\end{align}

The F1 score is $F1=\frac{2 \times \text{P} \times \text{R}}{\text{P} + \text{R}}$. %
When evaluating a text with respect to some data ($D$ as input, $T$ as output), \(\text{P}\) measures how much of $T$ is faithful to $D$ (penalizing hallucinations and inaccuracies), while \(\text{R}\) measures how much of $D$ is captured in $T$ (penalizing omissions and inaccuracies). %
An {\sc Incorrect} unit affects both scores: it is not a correctly covered input unit, lowering recall, and it is an inaccurate realised unit in the output, lowering precision. By contrast, {\sc Omitted} units affect recall only, and {\sc Extra} units affect precision only. Equivalently, the recall denominator counts all input units
($|\text{Correct}|+|\text{Incorrect}|+|\text{Omitted}|$), whereas the precision
denominator counts all realised output units
($|\text{Correct}|+|\text{Incorrect}|+|\text{Extra}|$).

\subsection{Data Sources}
\label{subsec:datasets}
We use three datasets of aligned data-English text pairs in our experiments: \textbf{WebNLG 3.0}\footnote{\url{https://gitlab.com/shimorina/webnlg-dataset/-/tree/master/release_v3.0}} \citep{webnlg2020}, a dataset of 45K data-text pairs where the data is a DBPedia graph and the text has been crowdsourced and verified to match the content of the corresponding graph; %
\textbf{{KELM}} ~\citep{agarwal-etal-2021-knowledge}, a silver corpus of 18M generated data--text pairs produced by a T5 model verbalising Wikidata triples; and \textbf{E2E} \cite{novikova-etal-2017-e2e}, 42K data-text pairs describing restaurant recommendations. We use the cleaned version of this dataset \cite{dusek-etal-2019-semantic}. 
WebNLG 3.0 and cleaned E2E provide the aligned pairs used to construct the synthetic training and error-detection test sets. KELM contributes additional data units to one Omitted perturbation and serves as the naturally noisy silver corpus in our human evaluation. For correlation analyses, we use 4L-RP-Human, the WebNLG 2020 and 2017 human evaluations, and E2E human judgements; Appendix~\ref{app:data} describes these resources in detail. 

\subsection{Implementation} 
The \metric\ model is a Transformer-based generative model fine-tuned on synthetic data composed of $(D, T, E)$ triples where $D$ is a set of triples, $T$ is a text and $E$ is a set of labeled triples highlighting the differences between $D$ and $T$ as specified in section~\ref{subsec:labels}. We train and test \metric\ evaluation models on two benchmarks: WebNLG %
and E2E. We also use KELM silver data for an evaluation comparing the model predictions with human annotations of the difference between text and data.  %

\subsection{Data Construction}
\label{subsec:synthetic-data}
We create the training data by applying perturbations on data-text pairs from two benchmarks where data and texts are semantically aligned: WebNLG and E2E. The gold data-text pairs provide {\sc Correct} examples. We then create labelled data for {\sc Incorrect, Omitted} and {\sc Extra} cases using perturbations designed to simulate each error type. Across all perturbations, we preserve a semantic connection to the original pair so that negative examples remain plausible and cannot be detected from topical mismatch alone. These perturbations are summarised in Table~\ref{tab:perturbations} and described in more detail as follows. %

\begin{table*}[]
\small
    \centering
    \begin{tabular}{llll}\toprule
    Gold Data & Data Perturb & Text Perturb & Description \\\midrule
       \bf Correct  & &&\\
       $(D,T)$ & $\tilde{D } =  D$ & $\tilde{T} = T$& Keep gold data \\\midrule
        \bf Omitted & &&\\
       $(D_1,T_1), D_2$ &  $\tilde{ D} =  D_1 \cup  D_2$ & $\tilde{T} = T$  & Merge two data, keep one text\\
        $(D,T)$ &  $\tilde{ D} =  D \cup \{d^+\}$& $\tilde{ T} =  T$ & Fabricate data items, keep the text\\\midrule
       \bf  Extra & &&\\
        $(D_1,T_1), (D_2, T_2)$ & $\tilde{ D} =  D_1$ &  $\tilde{ T} =  T_1 \|  T_2$ & Merge two texts, keep one data\\
        $(D,T)$&  $\tilde{ D} =  D \setminus \{d^{-}\}$ & $\tilde{ T} =  T$&  Drop data items, keep the text\\\midrule
        \bf Incorrect & &&\\
          $(D,T)$&  $\tilde{ D} = ( D\setminus \{d_i\}) \cup \{\tilde d_i\}$ & $\tilde{ T} =  T$ & Modify data items, keep the text\\\bottomrule
    \end{tabular}
    \caption{\textbf{Perturbations used to create the synthetic data.} $(D,T),(D_1,T_1), (D_2, T_2)$ are data-text pairs from either E2E or WebNLG. 
    $D_2$ in the first \textbf{Omitted} construction is sampled from E2E, WebNLG, or KELM.
    }
    \label{tab:perturbations}
\end{table*}

\paragraph{Omitted.}
To create data-text pairs where some data is {\sc Omitted} i.e., present in $ D$ but not in $ T$, we apply two types of perturbations: (1) extending a gold data-text pairs with related data; (2) adding newly fabricated triples. 

(1) \textit{Extending a gold data instance with related data from existing dataset.}  Given  $\langle ( D_1, T_1), D_2 \rangle$ where $( D_1, T_1)$ is sampled from either WebNLG or E2E, $D_2$ is a data unit from WebNLG,  KELM or E2E such that there is no triple in $D_2$ that is also in $ D_1$, we create a new instance $(D_1 \cup D_2, T_1)$ and label all triples \( d_i \in  D_2 \) as {\sc Omitted}.
    To ensure that $ D_2$ is semantically related to $ D_1$, we adopt a relatedness-controlled selection strategy. First, we retrieve a candidate graph $ D_2$ that shares at least one entity or predicate with $ D_1$. This mimics realistic scenarios where data-to-text models fail to verbalise related but distinct information, or text-to-data models which generate extra but related data units. 
    Second, we apply an overlap-filtering mechanism to avoid near-duplicate triples. For each triple in the candidate graph, we compute its element-wise overlap with all triples in $ D_1$, defined as the number of matching elements across subject, predicate, and object. Any triple sharing two or three elements with any triple in $ D_1$ is discarded. 
    Finally, to maintain realistic data sizes, we cap $D_2$ to at most four triples.
    If more than four eligible triples remain after filtering, we randomly sample between one and four to prevent artificially large omission sets. 
    The resulting $D_2$ distribution is relatively uniform across sizes: 29.5\% containing 1 triple, 24.8\% containing 2 triples, 24.4\% containing 3 triples, and 21.3\% containing 4 triples.

    \color{black}
(2) \textit{Adding newly fabricated triples to a gold data-text pair.} 
    This perturbation introduces \emph{counterfactual} triples that are not present in $ T_1$ but are semantically plausible. 
    To generate such triples, we first randomly select a triple from $ D_1$, and then modify two of its fields while keeping a semantic anchor in the remaining field: either by fixing the predicate and altering both subject and object, or by fixing one entity and altering the other entity together with the predicate. This ensures that each new triple maintains a connection to $ D_1$ while still diverging from the text.

    Candidate replacements for entities and predicates are selected from a semantic neighbourhood constructed through co-occurrence statistics: two predicates are considered related if they ever share a subject or an object in WebNLG, E2E or KELM,
    and two entities are related if they have ever been connected with the same predicate. 
    This yields a pool of alternatives from similar semantic domains. To prevent replacements that are overly close to the original element, we further filter these candidates using BGE-1.5-Large \citep{xiao23bge}, retaining only the top 30 neighbours whose cosine similarity to the original element is below 0.65. If an element has no neighbours under the co-occurrence criterion, we compute its similarities with the pool of all predicates or all entities 
    and select from the top 30 candidates below the threshold. The resulting triple $d^+$ remains semantically coherent with $ D_1$ while not being expressed in $ T_1$, and is therefore labelled as {\sc Omitted}.
    
    \color{violet}

\color{black}

\paragraph{Extra.}
Symmetrically to {\sc Omitted} triples, {\sc Extra} triples are expressed in $T$ but absent from $D$. We construct these cases in two complementary ways from gold WebNLG or E2E pairs: either remove triples from the gold data while keeping its text unchanged, or concatenate two related texts while retaining only one graph as input. Triples expressed in the resulting text but absent from the resulting data are labelled as {\sc Extra}.

\paragraph{Incorrect.}
To create instances with {\sc Incorrect} triples, we modify only the data side and keep the text unchanged. Given a gold pair $(D,T)$, we select $d_i\in D$, replace it with a modified triple $\tilde d_i$, and create $((D\setminus\{d_i\})\cup\{\tilde d_i\},T)$. The modified triple $\tilde d_i$ is labelled as {\sc Incorrect}. 
We obtain $\tilde d_i$ in two ways. We either replace one element of $d_i$ with a semantically related but distinct alternative from the same candidate pool used to synthesise Omitted triples, or swap its subject and object while keeping the predicate unchanged.
This operation mimics a common failure mode of data-to-text and text-to-data models assigning the arguments to the wrong positions.  
Triples with symmetric predicates\footnote{Symmetric relation examples:  
\textit{“synonym”, “opposite of”, “coincident with”, “interacts with”, “partner”, 
“relative”, “related category”, “connects with”, “twinned body”,  
“different from”, “the same as”, “sibling”, “adjacent station”, “shares border with”}} are excluded from this perturbation  
as reversing their arguments does not meaningfully alter the semantics of the triple.

\paragraph{Error Combination.}
Each deviation type ({\sc Omitted, Extra, Incorrect}) is sampled independently through 
Bernoulli trials \citep{feller1957introduction}, allowing multiple perturbations 
to co-occur within the same example. To prevent duplicated or logically 
incompatible modifications, we enforce the following ordering across 
operations. 
First, triples introduced during the construction of {\sc Omitted} cases are 
treated as external additions and subsequent {\sc Extra} and {\sc Incorrect} 
operations operate exclusively on the original data graph, never removing or 
altering the omitted triples. Similarly, triples removed during {\sc Extra} 
construction are treated as permanently unavailable and are excluded from 
consideration in later {\sc Incorrect} operations. When a triple is modified 
under {\sc Incorrect}, its altered form is not used as a semantic reference 
for generating related content in {\sc Omitted} or {\sc Extra} perturbations: all semantic relations are computed with respect to the original unmodified graph. 
After all perturbations are applied, the final text $\tilde{ T}$ is 
shuffled at the sentence level and the final data $\tilde{ D}$ at the 
triple level. This removes positional cues that could simplify the verification 
task, while preserving the semantic validity of both modalities.
\color{black}
Table~\ref{tab:dataset_stats} and Table~\ref{tab:error_cooccurrence} in Appendix~\ref{app:synth-stats} present the statistics of the synthesized datasets and %
 the distribution of error numbers per instance. %

We also manually audited 180 synthetic examples:
90 from each dataset and 60 for each target error type.
The intended perturbation and its assigned error type were valid in
169/180 cases (93.9\%),
while 154/180 cases (85.6\%) contained no additional unlabelled alignment
errors.
Appendix~\ref{app:synth-stats} reports the audit criteria,
per-type results,
and representative examples.

\section{Experiments}
\label{sec:xp}

\subsection{Models and Baselines}
\label{subsec:models}

\paragraph{\metric.} We experiment with open source models from different sizes and families: {Gemma3} \citep{gemmateam2025gemma3technicalreport} (270M--12B), {Qwen3} \citep{yang2025qwen3technicalreport} (0.6B--14B), %
and {Llama~3} \citep{grattafiori2024llama3herdmodels} (1B--8B).
We fine-tune \metricw\ on synthetic data constructed from WebNLG 3.0 using the method described in Section~\ref{subsec:synthetic-data}. 
For \metrice, we combine the WebNLG and E2E synthetic training data because
E2E contains only seven distinct properties.
All fine-tuned verifiers use LoRA \citep{hu2022lora}. 
Model descriptions and hyperparameter settings are provided in
Appendices~\ref{appendix:model_v} and~\ref{sec:lora}.

\paragraph{Baselines.} 
We compare our fine-tuned \metric\ verifiers with prompted LLM-as-judges using the same input--output pairs and label schema; the prompts are shown in Appendix~\ref{appendix:prompt_baseline}. When comparing quantitative scores with human judgements, we also consider baseline metrics that output precision and recall.
\textbf{Data-QuestEval (DQE)} \citep{rebuffel-etal-2021-data} employs a question-answering  and question-generation framework: for precision, it generates questions from text and checks if the data can answer them; for recall, it generates questions from data and checks if the text can answer them. \textbf{NLI-Based} \citep{dusek-kasner-2020-evaluating} uses natural language inference: for precision, it concatenates all triples into a meaning representation and computes entailment probability from this representation to text; for recall, it checks whether the text entails each individual triple and averages the probabilities. It uses hand-crafted predicate templates to convert triples into text before applying NLI, making it particularly well suited to WebNLG and E2E datasets. \textbf{FactSpotter (FS)} \citep{zhang-etal-2023-factspotter} is a self-supervised recall-based model and cannot produce precision. 
\textbf{MonoLR} \citep{martinez-etal-2025-semantic} regresses precision and recall scores, trained on synthetic examples with incorrectness, omissions and additions. In all cases, F1 is the harmonic mean of precision and recall.

\begin{table}[!t]
\centering
\small
\setlength{\tabcolsep}{5pt}
\begin{adjustbox}{width=\columnwidth}
\begin{tabular}{lcccc}
\toprule
Model & Extra & Omitted & Incorrect & Macro F1 \\
\midrule
GPT-5.1-Prompt & 18.0 & 81.7 & 78.1 & 59.2 \\
GPT-4.1-Prompt & 16.8 & 30.8 & 64.1 & 37.2 \\
G3-27B-Prompt & 4.6 & 14.4 & 43.2 & 20.7 \\
Q3-32B-Prompt & 2.8 & 1.2 & 47.6 & 17.2 \\
L3.3-70B-Prompt & 13.2 & 51.8 & 52.3 & 39.1 \\
\midrule
G3-1B & 66.8 & 94.5 & 92.3 & 84.5 \\
G3-4B & 78.3 & \textbf{97.6} & 96.5 & 90.8 \\
G3-12B & \underline{81.3} & 96.5 & \underline{97.0} & \underline{91.6} \\
\midrule
Q3-0.6B & 69.7 & 96.2 & 95.2 & 87.1 \\
Q3-1.7B & 73.8 & \underline{96.9} & 96.1 & 88.9 \\
Q3-4B & 77.5 & \underline{97.2} & 96.6 & 90.4 \\
Q3-8B & 79.2 & 96.8 & \underline{96.9} & \underline{90.9} \\
Q3-14B & \underline{81.3} & 96.8 & \textbf{97.1} & \textbf{91.7} \\
\midrule
L3.2-1B & 71.6 & 92.4 & 93.1 & 85.7 \\
L3.2-3B & 78.6 & 93.7 & 95.0 & 89.1 \\
L3.1-8B & \textbf{82.2} & 93.8 & 95.6 & 90.5 \\
\bottomrule
\end{tabular}
\end{adjustbox}
\caption{\textbf{Triple-level F1 (\%) for prompt-based and fine-tuned \metric\ verifier models on synthetic data constructed from WebNLG.} %
Here and below, \textbf{bold} indicates the highest value per column; \underline{underline} indicates the second- and third-highest. G3, L3, and Q3 refer to the Gemma3, Llama3, and Qwen3 implementation of our \metric\ metric.}
\label{tab:all-models-f1-tri}
\end{table}

\begin{table}[!t]
\centering
\small
\setlength{\tabcolsep}{5pt}
\begin{tabular}{lcccc}
\toprule
Model & Extra & Omitted & Incorrect & Macro \\
\midrule
GPT-5.1-Prompt & 35.7 & 82.4 & 78.7 & 65.6 \\
GPT-4.1-Prompt & 33.1 & 41.2 & 62.1 & 45.5 \\
G3-27B-Prompt & 4.8 & 16.4 & 36.6 & 19.3 \\
Q3-32B-Prompt & 6.6 & 1.7 & 43.3 & 17.2 \\
L3.3-70B-Prompt & 19.8 & 57.1 & 54.3 & 43.7 \\
\midrule
G3-1B & 93.5 & 97.3 & 96.7 & 95.8 \\
G3-4B & 97.1 & 99.0 & \textbf{97.9} & 98.0 \\
G3-12B & \underline{97.7} & \textbf{99.3} & 97.8 & \textbf{98.3} \\
\midrule
Q3-0.6B & 96.3 & 98.7 & 97.7 & 97.6 \\
Q3-1.7B & 96.5 & 99.0 & 97.7 & 97.7 \\
Q3-4B & \underline{97.7} & \underline{99.1} & \textbf{97.9} & \underline{98.2} \\
Q3-8B & 97.5 & 99.0 & 97.7 & 98.1 \\
Q3-14B & \textbf{97.9} & \underline{99.2} & \textbf{97.9} & \textbf{98.3} \\
\midrule
L3.2-1B & 93.9 & 97.0 & 96.3 & 95.7 \\
L3.2-3B & 97.1 & 98.1 & 97.1 & 97.4 \\
L3.1-8B & \underline{97.8} & 98.0 & 97.3 & 97.7 \\ 
\bottomrule
\end{tabular}
\caption{\textbf{Triple-level F1 (\%) for prompt-based and fine-tuned
\metric\ models on the E2E synthetic data.}}
\label{tab:e2e-all-models-f1-tri}
\end{table}

\subsection{Evaluation Methods and Results} 
\label{subsec:evln-methods}
We evaluate \metric\ in three ways. First, we evaluate the explainable labels generated by our metric  by computing F1 scores on the test part of the synthetic data described in Section~\ref{subsec:synthetic-data} and evaluate: does \metric\ correctly identify erroneous and correct triples? 
Second, we evaluate the quantitative aspect of our metric by computing correlations between its quantified P/R/F1 scores and human judgements related to semantic adequacy across datasets. 
Third, we examine the consistency between human annotations and \metric's predictions for the three error types on silver data--text pairs. Element-level matching gives partial credit for recovering individual triple components, while similarity-based matching recognises semantically equivalent units that differ in surface form.

\subsubsection{How well does \metric\  detect the different types of errors?} 
We first evaluate on the test split of the synthetic data described in Section~\ref{subsec:synthetic-data} using 
three evaluation levels, i.e., \textit{triple-level, element-level}, and
\textit{similarity-based} metrics.  
Tables~\ref{tab:all-models-f1-tri}
and~\ref{tab:e2e-all-models-f1-tri} present the triple-level F1 results on
WebNLG and E2E where triple-level evaluation measures exact matches between predicted and gold triples: a triple is counted as correct only when the predicted subject, predicate, and object match the gold triple exactly.
Definitions of evaluation metrics and full results are provided in Appendix~\ref{appendix:pr_scores}, Tables~\ref{tab:all-models-F1}--\ref{tab:e2e-recall}.

\begin{table}[t]
\centering
\small
\begin{adjustbox}{max width=\columnwidth}
\begin{tabular}{lccc}
\toprule
Model & $P_m$ vs $P_h$ & $R_m$ vs $R_h$ & $F1_m$ vs $F1_h$ \\
\midrule
DQE & 61.2 & 58.3 & 62.5 \\
MonoLR         & 71.0 & 63.7 & 70.6 \\
NLI      & 73.4 & 73.2 & 73.3 \\
FS    & ---  & 86.5 & ---  \\
\midrule
G3-27B-Prompt  & 28.9 & 35.1 & 36.1 \\
Q3-32B-Prompt  & 35.9 & 30.4 & 35.0 \\
L3.3-70B-Prompt & 32.4 & 35.5 & 32.8 \\
GPT-5.1-Prompt & 80.9 & 76.8 & 78.5 \\
\midrule
G3-1B   & 74.3 & 81.5 & 81.3 \\
G3-4B   & 79.0 & \underline{87.6} & \underline{87.4} \\
G3-12B  & 61.5 & 55.7 & 62.7 \\
\midrule
Q3-0.6B  & 71.8 & 79.3 & 78.4 \\
Q3-1.7B  & 66.3 & 74.0 & 72.5 \\
Q3-4B    & \underline{88.5} & \underline{86.9} & \underline{88.1} \\
Q3-8B    & \textbf{88.6} & \textbf{88.9} & \textbf{89.6} \\
Q3-14B   & 60.2 & 64.8 & 69.5 \\
\midrule
L3.2-1B & 71.0 & 77.0 & 76.7 \\
L3.2-3B & \underline{85.0} & 80.4 & 85.9 \\
L3.1-8B & 82.4 & 78.3 & 84.3 \\
\bottomrule
\end{tabular}
\end{adjustbox}
\caption{\textbf{Spearman correlation} between $P_m$, $R_m$, $F1_m$ metric scores and  \textbf{4L-RP-Human} $P_h$, $R_h$, $F1_h$ annotations. Pearson's $r$ and Kendall's $\tau$ are reported in Appendix \ref{sec:webnlg_further}. }
\label{tab:4lrphuman_correlation}
\end{table}

\begin{table}[ht]
\centering
\small
\setlength{\tabcolsep}{4pt}
\begin{tabular}{lcccccc}
\toprule
& \multicolumn{2}{c}{$P_m$ vs $P_a$}
& \multicolumn{2}{c}{$R_m$ vs $R_a$}
& \multicolumn{2}{c}{$F1_m$ vs $F1_a$} \\
\cmidrule(lr){2-3}\cmidrule(lr){4-5}\cmidrule(lr){6-7}
Model
& $\rho_{\mathrm{sys}}$ & $\rho_{\mathrm{txt}}$
& $\rho_{\mathrm{sys}}$ & $\rho_{\mathrm{txt}}$
& $\rho_{\mathrm{sys}}$ & $\rho_{\mathrm{txt}}$ \\
\midrule
DQE & 66.3 & 56.2 & 72.3 & 56.6 & 74.3 & 62.8 \\
MonoLR         & 77.1 & 58.5 & 90.8 & 61.0 & 85.7 & \underline{65.1} \\
NLI      & 66.0 & 62.4 & 85.3 & 62.1 & 80.5 & 63.2 \\
FS    & {---} & {---} & 89.5 & 63.7 & {---} & {---} \\
\midrule
G3-27B-Prompt   & 60.8 & 48.0 & 56.2 & 52.1 & 58.2 & 53.2 \\
Q3-32B-Prompt   & 59.0 & 56.6 & 66.0 & 58.9 & 62.6 & 53.6 \\
L3.3-70B-Prompt & 65.8 & 57.2 & 73.5 & 54.5 & 70.4 & 54.4 \\
GPT-5.1-Prompt  & 71.2 & 61.9 & 80.4 & 62.3 & 76.9 & 62.4 \\
\midrule
G3-1B   & 78.8 & 62.6 & 91.1 & 63.7 & 86.4 & 64.0 \\
G3-4B   & 73.3 & \underline{63.2} & 88.7 & 62.3 & 83.1 & 64.3 \\
G3-12B  & 72.5 & 62.8 & 87.6 & \textbf{65.5} & 81.3 & \underline{64.9} \\
\midrule
Q3-0.6B  & \underline{80.9} & 63.0 & \underline{91.8} & 63.7 & 87.5 & 63.9 \\
Q3-1.7B  & 80.0 & \underline{63.2} & 91.3 & \underline{64.3} & \underline{87.6} & \textbf{65.3} \\
Q3-4B    & 65.9 & 62.5 & 83.8 & 62.8 & 76.2 & 64.5 \\
Q3-8B    & 76.0 & \underline{63.2} & 91.2 & \underline{64.1} & 85.6 & 63.2 \\
Q3-14B   & 71.1 & 60.5 & 90.0 & 62.7 & 84.3 & 63.7 \\
\midrule
L3.2-1B & \textbf{84.8} & 62.4 & \textbf{94.2} & 63.5 & \textbf{90.8} & 63.9 \\
L3.2-3B & 69.2 & \textbf{63.7} & 85.4 & 62.7 & 78.9 & 63.8 \\
L3.1-8B & 73.7 & 62.4 & 87.8 & 62.4 & 82.2 & 63.6 \\
\bottomrule
\end{tabular}
\caption{\textbf{Spearman correlation} at system level ($\rho_{\mathrm{sys}}$) and
text level ($\rho_{\mathrm{txt}}$) between $P_m$, $R_m$, $F1_m$ metric scores  and $P_a$, $R_a$, $F1_a$ scores derived from \textbf{WebNLG 2020 human-judgments.}}
\label{tab:webnlg20-spearman-combined}
\end{table}

\begin{table}[ht]
\centering
\small
\setlength{\tabcolsep}{3pt}
\begin{adjustbox}{width=\columnwidth}
\begin{tabular}{lccccccccc}
\toprule
& \multicolumn{3}{c}{{Is-ok}} & \multicolumn{3}{c}{{Has-miss.}} & \multicolumn{3}{c}{{Has-add.}} \\
Model & $P$ & $R$ & $F1$ & $P$ & $R$ & $F1$ & $P$ & $R$ & $F1$ \\
\midrule
{NLI} & \underline{92.7} & 86.4 & \textbf{89.4} & 68.0 & \underline{90.1} & \textbf{77.5} & 30.9 & {32.7} & \textbf{31.8} \\
{DQE}  & 73.7 & \textbf{91.0} & 81.5 & 37.2 & 14.4 & 20.8 &  7.1 &  5.7 &  6.3 \\
{FS} & {93.0} & 17.9 & 30.1 & 28.5 & {98.0} & 44.2 & ---  & ---  & ---  \\
\midrule
G3-27B-Prompt
& 0.0 & 0.0 & 0.0
& 25.1 & \textbf{99.8} & 40.1
& 3.6 & \textbf{92.8} & 7.0 \\
Q3-32B-Prompt
& 87.3 & 21.6 & 34.6
& 27.2 & 86.9 & 41.4
& 4.0 & \underline{86.1} & 7.6 \\
L3.3-70B-Prompt
& \textbf{93.4} & 46.0 & 61.6
& 35.1 & 82.4 & 49.2
& 4.7 & \underline{79.8} & 9.0 \\
\midrule
G3-1B   & 90.2 & 88.1 & 89.1 & 68.6 & 83.7 & 75.4 & \textbf{43.6} & 16.3 & 23.8 \\
G3-4B   & 89.9 & 88.5 & 89.2 & \textbf{69.6} & 82.8 & 75.6 & 29.1 & 15.4 & 20.1 \\
G3-12B  & 90.2 & 88.2 & 89.2 & 69.0 & \underline{83.9} & 75.7 & 29.2 & 19.2 & 23.2 \\
\midrule
Q3-0.6B  & \underline{90.3} & 88.3 & \underline{89.3} & 69.0 & \underline{83.9} & 75.7 & 34.4 & 14.9 & 20.8 \\
Q3-1.7B  & 90.1 & 88.4 & \underline{89.3} & 69.3 & 83.4 & 75.7 & 31.2 & 18.8 & 23.4 \\
Q3-4B    & \underline{90.3} & 88.4 & \textbf{89.4} & 69.3 & 83.7 & \underline{75.8} & 31.4 & 15.9 & 21.1 \\
Q3-8B    & 90.0 & \underline{88.7} & \textbf{89.4} & \textbf{69.6} & 83.0 & 75.7 & 24.1 & \underline{22.6} & 23.3 \\
Q3-14B   & \underline{90.4} & 87.8 & 89.1 & 68.4 & \underline{84.2} & 75.5 & 30.8 & 13.5 & 18.7 \\
\midrule
L3.2-1B & 89.9 & 88.4 & 89.1 & 69.2 & 82.6 & 75.3 & \underline{37.4} & 17.8 & \underline{24.1} \\
L3.2-3B & 90.2 & \underline{88.6} & \textbf{89.4} & \underline{69.5} & 83.7 & \underline{76.0} & 31.5 & 16.8 & 21.9 \\
L3.1-8B & \underline{90.4} & 88.3 & \underline{89.3} & \underline{69.5} & \underline{83.9} & \underline{76.0} & 35.1 & 16.3 & 22.3 \\
\bottomrule
\end{tabular}
\end{adjustbox}
\caption{\textbf{Precision, recall, and F1} on \textbf{E2E coarse label}. }
\label{tab:e2e-coarse}
\end{table}

Our fine-tuned models substantially outperform prompt-based approaches across all error types and metrics on both datasets. Among prompt-based models, GPT-5.1 achieves the best performance, though still below our fine-tuned models.
The results also show the {\sc Extra} error type is the most challenging for the fine-tuned models. Unlike {\sc Omitted} and {\sc Incorrect} errors, where the model can copy triples directly from the input data, detecting extra content requires the model to identify and decompose information from the text into atomic data units, making it a fundamentally harder task. Comparing results on E2E \textit{vs} WebNLG, we see that our fine-tuned models achieve notably higher scores on E2E. We attribute this to the much lower data diversity in E2E: it has only 7 unique predicates and 78 unique entities, compared to 411 predicates and 3,622 entities in WebNLG. The best-performing models across model families achieve comparable macro-averaged F1 scores, and larger models bring marginal improvements.

\subsubsection{Comparison with  Human Judgements}
\label{sec:comparison_human}
\label{sec:eval4l_rp_human}

\paragraph{Correlation with Human Judgements of Recall and Precision.} \citet{martinez-etal-2025-semantic}'s 4L-RP-Human dataset 
provides human annotations of precision and recall scores for data-text pairs from WebNLG, making it well suited to evaluate \metricw\ quantitative scores. %
Table~\ref{tab:4lrphuman_correlation}, which reports correlation results %
on this dataset\footnote{The MonoLR baseline score is higher than reported by \citet{martinez-etal-2025-semantic} because we removed invalid negative precision annotations.}, shows that the best fine-tuned verifier, Qwen3-8B, achieves the highest correlations for all three scores and outperforms GPT-5.1 used as a prompted judge. 
Recall correlations tend to remain consistently high across models, whereas precision correlations exhibit greater variability. We conjecture that this is due to the increased difficulty of determining what should not be included in the text, compared to identifying what should be present. %
Model size does not consistently predict agreement with human judgements.
Qwen3-8B outperforms Qwen3-14B, while Llama3.2-3B outperforms Llama3.1-8B.
By contrast, Qwen3-14B slightly outperforms Qwen3-8B on the synthetic test set
(91.7 vs.\ 90.9 F1), showing that synthetic error detection and agreement
with human judgements capture different aspects of performance. One plausible explanation is that controlled synthetic perturbations and
naturally occurring errors are related but not identically distributed.
Beyond 8B in our experiments, additional model capacity may capture more
construction-specific patterns without improving generalisation to independent
human judgements.

\color{black}

\paragraph{Correlation with WebNLG Human Judgements of Semantic Adequacy}
\label{sec:corr_webnlg}
WebNLG 2020 \citep{webnlg2020} provides human judgements for 3 semantic dimensions: data correctness (CR), relevance (RV), and coverage (CV), which correspond to precision- and recall-related aspects of semantic adequacy. Following \citet{martinez-etal-2025-semantic}, we derive human precision as $P_a = CR \times RV$, recall as $R_a = CV$, and $F1_a$ as the harmonic mean of $P_a$ and $R_a$, and correlate metric outputs ($P_m$, $R_m$, $F1_m$) with the targets at system and text levels. Table~\ref{tab:webnlg20-spearman-combined} reports the Spearman results.
At system level, our models achieve strong Spearman correlations with $P_a$, $R_a$, $F1_a$.
At text level, our models attain the highest precision and recall correlations among all compared methods, with marginal gains on F1. Compact verifiers already achieve high correlations, and scaling up yields no consistent gains.
We report all correlations on WebNLG 2017 in Appendix~\ref{sec:webnlg2017}.

\paragraph{Comparison with E2E Human Judgements}
\label{sec:corr_e2e}

The E2E human judgements \citep{DUSEK2020123} provide both coarse error-type labels (\textit{ok}, \textit{missing}, and \textit{added}) and global quality scores between 0--100. We focus here on the coarse labels and report precision, recall, and F1 for the three corresponding binary classification tasks in Table~\ref{tab:e2e-coarse}. To evaluate the metrics against these labels, we map each metric output to \textit{is-ok}, \textit{has-missing}, and \textit{has-added}. For XQDT, we map an output to \textit{has-missing} if it contains at least one omitted or incorrect unit, to \textit{has-added} if it contains at least one extra unit, and to \textit{is-ok} only when all predicted units are correct. Performance on \textit{is-ok} and \textit{has-missing} is consistently strong, whereas \textit{has-added} is markedly weaker, consistent with the very low inter-annotator agreement for the \textit{added} label (Krippendorff's $\alpha = 0.104$, Table~\ref{tab:e2e-iaa}). Global-quality correlations and the mappings for the other metrics are discussed in Appendix~\ref{app:e2e_details}.

NLI performs comparatively well on this benchmark for two reasons. First, E2E evaluation collapses XQDT’s fine-grained labels into three coarse categories, which partially obscures the benefits of explicit error localisation. Second, the NLI baseline uses hand-crafted templates that are particularly effective in E2E’s closed-domain setting with limited schema diversity. The prompted judges generally favour recall over precision for \textit{has-missing} and \textit{has-added}, suggesting that they over-predict errors; consequently, their F1 scores are substantially lower than those of the fine-tuned verifiers.

\paragraph{Cross-dataset generalisation.}
We compare Qwen3 verifiers trained on WebNLG 3.0 synthetic data only
with verifiers of the same size trained on the combined WebNLG 3.0 and E2E
synthetic training sets.
All models are evaluated on the same E2E examples. 
The 4B and 8B models are the most stable:
their \textit{is-ok} and \textit{has-missing} F1 scores differ by at most
0.6 points between the two training-data settings,
and their text-level correlations differ by at most 1.0 point.
The WebNLG-only models also achieve triple-level F1 above 92.0 for both Omitted and Incorrect errors.
For Extra errors, however,
strict F1 is much lower than similarity F1,
which remains above 80.0.
Full results are shown in Appendix~\ref{app:cross_dataset}.

\subsubsection{Consistency with Human Annotations on Noisy Data}
\label{sec:kelm_human_verification}

A natural downstream application of alignment error detection is automated quality auditing of silver-standard corpora, enabling both the selection of high-quality subsets for downstream training and the identification of misaligned instances for targeted post-processing. 
We evaluate this capability on KELM \citep{agarwal-etal-2021-knowledge}, a large-scale silver corpus of approximately 18M automatically generated data--text pairs, where alignment errors are expected but exhaustive manual annotation is prohibitively costly. 
For each data--text pair, annotators are asked to verify the error items reported by \metric : (i) judging whether each is a false alarm or a genuine error, (ii) whether the error type and the target triple are correct, and (iii) indicating any errors the model fails to report. 
Appendix~\ref{sec:data_selection} describes data selection, Appendix~\ref{sec:annotation_protocol} details the annotation protocol, and Figure~\ref{fig:anno_interface} shows the interface.
Table~\ref{tab:kelm_prf} reports instance-averaged precision, recall, and F1 for fine-grained error-item detection.
Qwen3-8B achieves the highest precision and the highest F1, whereas GPT-5.1 achieves the highest recall but has lower precision.
The two models obtain similar F1 scores, but Qwen3-8B provides a more balanced precision--recall trade-off and substantially outperforms Qwen3-32B prompting.
Detailed consistency results and further analyses on KELM are reported in Appendix~\ref{sec:kelm_appendix}.

\begin{table}[t]
\centering
\small
\begin{tabular}{lcccc}
\toprule
Model & P & R & F1 & F1 95\% CI \\
\midrule
Q3-32B-Prompt & 35.4 & 38.1 & 33.2 & [25.2, 42.9] \\
GPT-5.1-Prompt & \underline{63.8} & \textbf{86.5} & \underline{68.1} & [58.8, 77.3] \\
\midrule
Q3-0.6B & 43.0 & 53.0 & 45.3 & [34.9, 55.9] \\
Q3-4B   & 56.7 & 63.1 & 57.6 & [47.3, 67.7] \\
Q3-8B   & \textbf{69.4} & {74.1} & \textbf{68.9} & [59.3, 78.4] \\
Q3-14B  & 46.6 & 52.8 & 47.6 & [36.1, 58.3] \\
\bottomrule
\end{tabular}
\caption{\textbf{Instance-averaged P/R/F1} for fine-grained error-item detection on KELM (all values in \%, 95\% bootstrap confidence intervals (CI) for F1) \textit{How well does \metricw\ predict errors ?}}
\label{tab:kelm_prf}
\end{table}

\section{Using \metric\ for Data-to-Text and Text-to-Data Improvement}

We next study whether \metric\ can serve as a feedback signal for improving data-to-text and text-to-data outputs. We use the Qwen3-8B verifier as the default verifier in downstream experiments, motivated by its strong performance on the 4L-RP-Human and KELM evaluations. Starting from a base system output, we apply the verifier to the input-output pair, convert predicted omitted, extra, and incorrect items into feedback, and provide it to a revision LLM for iterative correction. 
The corresponding repair prompts are provided in Appendix~\ref{sec:appendix_repair_results}.
We run up to three rounds of revision, yielding R1, R2, and R3 outputs. Across tasks, we compare XQDT-based feedback with baselines including self-refine, NLI-based feedback, and FactSpotter-based feedback, and additionally with PiVe for text-to-data.

\subsection{Text-to-Data Improvement}

For text-to-data improvement, the verifier compares a predicted triple set against the source text and identifies omitted, extra, and incorrect triples, which the revision LLM uses to add, remove, or revise triples. We evaluate BT5 \citep{agarwal-etal-2020-machine} and ReGen \citep{dognin-etal-2021-regen} on WebNLG and GenWiki using the Exact, Strict, and Partial metrics of \citet{webnlg2020}. Tables~\ref{tab:maintext_genwiki_exact_compact} and~\ref{tab:maintext_webnlg_exact_compact} show representative Exact F1 results for R1 and R3, while full results for all sizes and three metrics are reported in
Appendix~\ref{sec:appendix_repair_results}, Tables~\ref{tab:appendix_webnlg_bt5_r123}--\ref{tab:appendix_genwiki_regen_r123}.
Relative to the unrevised base outputs, XQDT improves Exact F1 by up to
9.9 points on GenWiki and 6.6 points on WebNLG. 
On WebNLG, XQDT performs best across the representative BT5 and ReGen settings. On GenWiki, the appendix shows the same pattern in most settings under Exact and Strict, while NLI and PiVe are more competitive under Partial F1, likely due to metric sensitivity rather than a reversal of the overall ranking. Qualitative examples are in Appendix~\ref{sec:appendix_repair_results}. Figures~\ref{fig:appendix_t2d_case_id1287} and~\ref{fig:appendix_t2d_case_id1296} show representative round-by-round text-to-data repair trajectories.

\begin{table}[t]
\centering
\scriptsize
\begin{adjustbox}{max width=\columnwidth}
\begin{tabular}{lcccccc}
\toprule
& \multicolumn{3}{c}{BT5} & \multicolumn{3}{c}{ReGen} \\
\cmidrule(lr){2-4} \cmidrule(lr){5-7}
Method & G3-12B & L3.1-8B & Q3-14B & G3-12B & L3.1-8B & Q3-14B \\
\midrule
Baseline & 40.89 & 40.89 & 40.89 & 47.53 & 47.53 & 47.53 \\
\midrule
XQDT-R1 & 48.60 & 44.65 & 43.26 & 50.67 & 49.67 & 48.45 \\
XQDT-R3 & \textbf{50.79} & \textbf{47.80} & \textbf{47.08} & \textbf{51.95} & \textbf{50.48} & \textbf{49.48} \\
\midrule
SR-R1 & 42.13 & 40.26 & 41.49 & 44.08 & 47.18 & 44.23 \\
SR-R3 & 41.96 & 40.01 & 40.27 & 43.79 & 45.56 & 41.01 \\
\midrule
FS-R1 & 37.73 & 40.13 & 33.67 & 44.56 & 46.49 & 43.41 \\
FS-R3 & 36.96 & 38.92 & 33.10 & 44.24 & 45.57 & 43.18 \\
\midrule
NLI-R1 & 47.04 & 41.09 & 38.33 & 49.35 & 46.71 & 45.33 \\
NLI-R3 & 49.36 & 41.58 & 40.14 & 51.25 & 46.37 & 45.69 \\
\midrule
PiVe-R1 & 40.73 & 40.66 & 40.73 & 47.34 & 47.25 & 47.34 \\
PiVe-R3 & 40.89 & 40.83 & 40.89 & 47.53 & 47.35 & 47.53 \\
\bottomrule
\end{tabular}
\end{adjustbox}
\caption{\textbf{Improved Text-to-Data Scores on GenWiki.} Exact F1 (\%) on GenWiki with BT5 and ReGen as base text-to-data models. SR denotes Self-Refine; FS denotes FactSpotter; and R1/R3 denote 1 or 3 rounds of revision.} 
\label{tab:maintext_genwiki_exact_compact}
\end{table}

\begin{table}[t]
\centering
\scriptsize
\begin{adjustbox}{max width=\columnwidth}
\begin{tabular}{lcccccc}
\toprule
& \multicolumn{3}{c}{BT5} & \multicolumn{3}{c}{ReGen} \\
\cmidrule(lr){2-4} \cmidrule(lr){5-7}
Method & G3-12B & L3.1-8B & Q3-14B & G3-12B & L3.1-8B & Q3-14B \\
\midrule
Baseline & 67.47 & 67.47 & 67.47 & 73.15 & 73.15 & 73.15 \\
\midrule
XQDT-R1 & 72.01 & 69.60 & 72.30 & 74.98 & 74.21 & 75.33 \\
XQDT-R3 & \textbf{72.49} & \textbf{70.65} & \textbf{74.07} & \textbf{75.34} & \textbf{74.47} & \textbf{75.67} \\
\midrule
SR-R1 & 58.29 & 62.01 & 62.04 & 60.85 & 66.00 & 65.50 \\
SR-R3 & 55.00 & 57.90 & 58.66 & 56.48 & 60.41 & 60.12 \\
\midrule
FS-R1 & 65.46 & 67.38 & 64.83 & 73.06 & 73.59 & 73.23 \\
FS-R3 & 65.08 & 66.98 & 64.50 & 72.87 & 73.23 & 72.99 \\
\midrule
NLI-R1 & 68.07 & 66.09 & 66.21 & 72.35 & 71.85 & 71.98 \\
NLI-R3 & 68.81 & 65.77 & 66.89 & 72.74 & 71.06 & 71.98 \\
\midrule
PiVe-R1 & 67.63 & 67.52 & 67.63 & 73.37 & 73.23 & 73.37 \\
PiVe-R3 & 67.47 & 67.34 & 67.47 & 73.15 & 72.90 & 73.15 \\
\bottomrule
\end{tabular}
\end{adjustbox}
\caption{\textbf{Improved Text-to-Data Scores on WebNLG.} {Exact F1} (\%) for improved text-to-data results on WebNLG with BT5 and ReGen as base extractors. }
\label{tab:maintext_webnlg_exact_compact}
\end{table}

\subsection{Data-to-Text Improvement}

For data-to-text improvement, the verifier compares a generated text against the data and identifies omitted, extra, and incorrect contents, which are converted into feedback for a revision LLM to revise the text while preserving source-supported content. We evaluate on WebNLG 2020 using Control Prefix \citep{clive-etal-2022-control} as the base generator and compare XQDT feedback against the same baselines except PiVe. Representative results are shown in Table~\ref{tab:webnlg_controlprefix20_two_models}, and full results across all repair models on the WebNLG Control Prefix baseline are in Appendix Table~\ref{tab:appendix_webnlg_control_prefix_r13}.
Because Control Prefix is already strong on WebNLG, room for post-hoc repair is limited and the gains are modest. Nevertheless, XQDT still provides measurable improvements on several automatic metrics.
Additional data-to-text examples are shown in Section~\ref{sec:appendix_repair_results}, Figures~\ref{fig:appendix_d2t_case_701} and~\ref{fig:appendix_d2t_case_400}.

\begin{table}[t]
\centering
\scriptsize

\begin{adjustbox}{max width=\columnwidth}
\begin{tabular}{llccccc}
\toprule
Model & Method & BLEU & chrF++ & METEOR & PARENT-F1 & BERTS-F1 \\
\midrule
\multirow{9}{*}{G3-4B}
& Baseline & \textbf{53.81} & 69.18 & 41.58 & 65.60 & 95.72 \\
\cmidrule(lr){2-7}
& SR-R1 & 51.99 & 67.95 & 40.98 & 65.70 & 95.55 \\
& SR-R3 & 51.54 & 67.62 & 40.82 & 65.65 & 95.51 \\
\cmidrule(lr){2-7}
& FS-R1 & 53.43 & 69.05 & 41.57 & 65.44 & \textbf{95.73} \\
& FS-R3 & 53.35 & 69.11 & 41.63 & 65.50 & 95.72 \\
\cmidrule(lr){2-7}
& NLI-R1 & 52.67 & 68.38 & 41.20 & 65.09 & 95.64 \\
& NLI-R3 & 52.48 & 68.38 & 41.22 & 65.18 & 95.64 \\
\cmidrule(lr){2-7}
& XQDT-R1 & 53.73 & 69.45 & 41.80 & 65.76 & \textbf{95.73} \\
& XQDT-R3 & 53.69 & \textbf{69.49} & \textbf{41.85} & \textbf{65.81} & \textbf{95.73} \\
\midrule
\multirow{9}{*}{L3.1-8B}
& Baseline & 53.81 & 69.18 & 41.58 & 65.60 & 95.72 \\
\cmidrule(lr){2-7}
& SR-R1 & 48.51 & 66.47 & 39.86 & 62.93 & 95.19 \\
& SR-R3 & 47.26 & 65.98 & 39.72 & 62.65 & 95.09 \\
\cmidrule(lr){2-7}
& FS-R1 & 53.52 & 68.88 & 41.44 & 65.30 & 95.71 \\
& FS-R3 & 53.45 & 69.21 & 41.65 & 65.21 & 95.72 \\
\cmidrule(lr){2-7}
& NLI-R1 & 53.05 & 68.77 & 41.38 & 65.00 & 95.71 \\
& NLI-R3 & 53.08 & 69.02 & 41.60 & 65.17 & 95.72 \\
\cmidrule(lr){2-7}
& XQDT-R1 & \textbf{53.93} & 69.48 & 41.82 & \textbf{65.73} & \textbf{95.75} \\
& XQDT-R3 & 53.88 & \textbf{69.55} & \textbf{41.86} & 65.71 & \textbf{95.75} \\
\bottomrule
\end{tabular}
\end{adjustbox}
\caption{\textbf{Data-to-Text improvement scores on WebNLG} using the Control Prefix baseline. BERTS-F1 denotes BERTScore-F1. }
\label{tab:webnlg_controlprefix20_two_models}
\end{table}

\section{Conclusion}
\label{sec:conclusion}

We propose a novel evaluation metric for bidirectional data-text alignment that differs from existing work in that it provides both a fine-grained, qualitative (error types) and quantitative (R/P/F1) analysis of data-text alignment errors. By identifying the sources of errors, \metric\ enables detailed error analysis and suggests directions for model improvement.
Our experiments show that \metric\ is indeed effective both as an evaluation tool
and as a feedback signal for data extraction and text generation:
it yields substantial gains for text-to-data extraction and smaller but
measurable improvements for data-to-text generation. Results across multiple revision models suggest that alignment-oriented verification is not only useful for assessment, but can also serve as a practical component in downstream correction and refinement pipelines.

\section*{Limitations}

Our framework is designed around triples as the basic semantic units of alignment, and both the perturbation strategy and the error labels rely on the relative atomicity of subject--predicate--object structures. This makes the formulation well suited to graph-to-text and text-to-graph settings, but also limits its direct transfer to richer structured representations such as AMR, dependency structures, or discourse graphs, where meaning is often distributed across multiple components and longer contexts. In such settings, local substitutions or merge operations do not necessarily yield equally clear and controlled alignment deviations.

Our precision, recall, and F1 are computed with data-item-level counting, whereas human judgements of data--text alignment quality may operate at a finer semantic granularity and allow some degree of partial credit. In particular, when only one or two elements of a triple are incorrect, annotators may penalize partially, while our formulation counts the entire triple as incorrect. %
As a result, our scores and human scores may differ systematically in absolute scale.

Because our model produces structured unit-level error reports rather than a single global score, inference cost can increase with the number of error units that must be reported in the worst case, especially when the text contains a large amount of content unsupported by the input data. This makes the approach more computationally demanding than regression- or classification-based metrics that return only one scalar score.

The downstream utility of verifier-based feedback is also task-dependent. Although the same alignment labels are used in both text-to-data and data-to-text improvement, the corresponding correction spaces differ substantially: in text-to-data, feedback maps naturally to structured edit operations over triples, whereas in data-to-text it must be realised through free-form text revision. As a result, gains in downstream improvement should not be interpreted as directly comparable across tasks, even when they are driven by the same verifier outputs.

Because our verifier outputs are consumed as revision feedback, local misjudgments can induce unnecessary edits rather than merely affecting evaluation scores. In iterative settings, this increases the sensitivity of downstream revisions to verifier errors, even though some earlier mistakes may be corrected in later rounds.

\section*{Acknowledgments}
We thank the anonymous reviewers for their feedback. This work received government funding managed by the French National Research Agency under France 2030, reference number ``ANR-23-IACL-0004'' (AI Chair Gardent: ``Semantically Consistent LLM Based Text Generation''). Experiments presented in this paper were carried out using the Grid'5000 testbed, supported by a scientific interest group hosted by Inria and including CNRS, RENATER, and several universities as well as other organizations (see \url{https://www.grid5000.fr}). This work was also granted access to the HPC resources of IDRIS under the allocation AD011016561 made by GENCI.

\bibliography{custom}

\begin{thebibliography}{45}
\providecommand{\natexlab}[1]{#1}

\bibitem[{Agarwal et~al.(2021)Agarwal, Ge, Shakeri, and
  Al-Rfou}]{agarwal-etal-2021-knowledge}
Oshin Agarwal, Heming Ge, Siamak Shakeri, and Rami Al-Rfou. 2021.
\newblock \href {https://doi.org/10.18653/v1/2021.naacl-main.278} {Knowledge
  graph based synthetic corpus generation for knowledge-enhanced language model
  pre-training}.
\newblock In \emph{Proceedings of the 2021 Conference of the North American
  Chapter of the Association for Computational Linguistics: Human Language
  Technologies}, pages 3554--3565, Online. Association for Computational
  Linguistics.

\bibitem[{Agarwal et~al.(2020)Agarwal, Kale, Ge, Shakeri, and
  Al-Rfou}]{agarwal-etal-2020-machine}
Oshin Agarwal, Mihir Kale, Heming Ge, Siamak Shakeri, and Rami Al-Rfou. 2020.
\newblock \href {https://aclanthology.org/2020.webnlg-1.13/} {Machine
  translation aided bilingual data-to-text generation and semantic parsing}.
\newblock In \emph{Proceedings of the 3rd International Workshop on Natural
  Language Generation from the Semantic Web (WebNLG+)}, pages 125--130, Dublin,
  Ireland (Virtual). Association for Computational Linguistics.

\bibitem[{Banerjee and Lavie(2005)}]{banerjee-lavie-2005-meteor}
Satanjeev Banerjee and Alon Lavie. 2005.
\newblock \href {https://aclanthology.org/W05-0909/} {{METEOR}: An automatic
  metric for {MT} evaluation with improved correlation with human judgments}.
\newblock In \emph{Proceedings of the {ACL} Workshop on Intrinsic and Extrinsic
  Evaluation Measures for Machine Translation and/or Summarization}, pages
  65--72, Ann Arbor, Michigan. Association for Computational Linguistics.

\bibitem[{Castro~Ferreira et~al.(2020)Castro~Ferreira, Gardent, Ilinykh,
  van~der Lee, Mille, Moussallem, and Shimorina}]{webnlg2020}
Thiago Castro~Ferreira, Claire Gardent, Nikolai Ilinykh, Chris van~der Lee,
  Simon Mille, Diego Moussallem, and Anastasia Shimorina. 2020.
\newblock \href {https://aclanthology.org/2020.webnlg-1.7/} {The 2020
  bilingual, bi-directional {W}eb{NLG}+ shared task: Overview and evaluation
  results ({W}eb{NLG}+ 2020)}.
\newblock In \emph{Proceedings of the 3rd International Workshop on Natural
  Language Generation from the Semantic Web (WebNLG+)}, pages 55--76, Dublin,
  Ireland (Virtual). Association for Computational Linguistics.

\bibitem[{Clive et~al.(2022)Clive, Cao, and Rei}]{clive-etal-2022-control}
Jordan Clive, Kris Cao, and Marek Rei. 2022.
\newblock \href {https://doi.org/10.18653/v1/2022.gem-1.31} {Control prefixes
  for parameter-efficient text generation}.
\newblock In \emph{Proceedings of the Second Workshop on Natural Language
  Generation, Evaluation, and Metrics (GEM)}, pages 363--382, Abu Dhabi, United
  Arab Emirates (Hybrid). Association for Computational Linguistics.

\bibitem[{Cripwell et~al.(2023)Cripwell, Belz, Gardent, Gatt, Borg, Borg,
  Judge, Lorandi, Nikiforovskaya, Soto-Martinez, and
  Thomson}]{cripwell-etal-2023-2023}
Liam Cripwell, Anya Belz, Claire Gardent, Albert Gatt, Claudia Borg, Marthese
  Borg, John Judge, Michela Lorandi, Anna Nikiforovskaya, William
  Soto-Martinez, and Craig Thomson. 2023.
\newblock \href {https://aclanthology.org/2023.mmnlg-1.6/} {The 2023 {W}eb{NLG}
  shared task on low resource languages. overview and evaluation results
  ({W}eb{NLG} 2023)}.
\newblock In \emph{Proceedings of the Workshop on Multimodal, Multilingual
  Natural Language Generation and Multilingual WebNLG Challenge (MM-NLG 2023)},
  pages 55--66, Prague, Czech Republic. Association for Computational
  Linguistics.

\bibitem[{Dhingra et~al.(2019)Dhingra, Faruqui, Parikh, Chang, Das, and
  Cohen}]{dhingra-etal-2019-handling}
Bhuwan Dhingra, Manaal Faruqui, Ankur Parikh, Ming-Wei Chang, Dipanjan Das, and
  William Cohen. 2019.
\newblock \href {https://doi.org/10.18653/v1/P19-1483} {Handling divergent
  reference texts when evaluating table-to-text generation}.
\newblock In \emph{Proceedings of the 57th Annual Meeting of the Association
  for Computational Linguistics}, pages 4884--4895, Florence, Italy.
  Association for Computational Linguistics.

\bibitem[{Dognin et~al.(2021)Dognin, Padhi, Melnyk, and
  Das}]{dognin-etal-2021-regen}
Pierre Dognin, Inkit Padhi, Igor Melnyk, and Payel Das. 2021.
\newblock \href {https://doi.org/10.18653/v1/2021.emnlp-main.83} {{R}e{G}en:
  {R}einforcement learning for text and knowledge base generation using
  pretrained language models}.
\newblock In \emph{Proceedings of the 2021 Conference on Empirical Methods in
  Natural Language Processing}, pages 1084--1099, Online and Punta Cana,
  Dominican Republic. Association for Computational Linguistics.

\bibitem[{Du{\v{s}}ek et~al.(2019)Du{\v{s}}ek, Howcroft, and
  Rieser}]{dusek-etal-2019-semantic}
Ond{\v{r}}ej Du{\v{s}}ek, David~M. Howcroft, and Verena Rieser. 2019.
\newblock \href {https://doi.org/10.18653/v1/W19-8652} {Semantic noise matters
  for neural natural language generation}.
\newblock In \emph{Proceedings of the 12th International Conference on Natural
  Language Generation}, pages 421--426, Tokyo, Japan. Association for
  Computational Linguistics.

\bibitem[{Du{\v{s}}ek and Kasner(2020)}]{dusek-kasner-2020-evaluating}
Ond{\v{r}}ej Du{\v{s}}ek and Zden{\v{e}}k Kasner. 2020.
\newblock \href {https://doi.org/10.18653/v1/2020.inlg-1.19} {Evaluating
  semantic accuracy of data-to-text generation with natural language
  inference}.
\newblock In \emph{Proceedings of the 13th International Conference on Natural
  Language Generation}, pages 131--137, Dublin, Ireland. Association for
  Computational Linguistics.

\bibitem[{Du{\v{s}}ek et~al.(2020)Du{\v{s}}ek, Novikova, and
  Rieser}]{DUSEK2020123}
Ond{\v{r}}ej Du{\v{s}}ek, Jekaterina Novikova, and Verena Rieser. 2020.
\newblock \href {https://doi.org/10.1016/j.csl.2019.06.009} {Evaluating the
  state-of-the-art of end-to-end natural language generation: The e2e nlg
  challenge}.
\newblock \emph{Computer Speech and Language}, 59:123--156.

\bibitem[{Feller(1957)}]{feller1957introduction}
W.~Feller. 1957.
\newblock \href {https://books.google.fr/books?id=rZIPzwEACAAJ} {\emph{An
  Introduction to Probability Theory and Its Applications}}.
\newblock A Wiley publication in mathematical statistics. Wiley.

\bibitem[{Gardent et~al.(2017)Gardent, Shimorina, Narayan, and
  Perez-Beltrachini}]{gardent-etal-2017-webnlg}
Claire Gardent, Anastasia Shimorina, Shashi Narayan, and Laura
  Perez-Beltrachini. 2017.
\newblock \href {https://doi.org/10.18653/v1/W17-3518} {The {W}eb{NLG}
  challenge: Generating text from {RDF} data}.
\newblock In \emph{Proceedings of the 10th International Conference on Natural
  Language Generation}, pages 124--133, Santiago de Compostela, Spain.
  Association for Computational Linguistics.

\bibitem[{Gero et~al.(2023)Gero, Singh, Cheng, Naumann, Galley, Gao, and
  Poon}]{gero2023selfverification}
Zelalem Gero, Chandan Singh, Hao Cheng, Tristan Naumann, Michel Galley,
  Jianfeng Gao, and Hoifung Poon. 2023.
\newblock \href {https://openreview.net/forum?id=SBbJICrglS} {Self-verification
  improves few-shot clinical information extraction}.
\newblock In \emph{ICML 3rd Workshop on Interpretable Machine Learning in
  Healthcare (IMLH)}.

\bibitem[{Grattafiori et~al.(2024)Grattafiori, Dubey, Jauhri, Pandey, Kadian,
  Al-Dahle, Letman, Mathur, Schelten, Vaughan, Yang, Fan, Goyal, Hartshorn,
  Yang, Mitra, Sravankumar, Korenev, Hinsvark, Rao, Zhang, Rodriguez,
  Gregerson, Spataru, Roziere, Biron, Tang, Chern, Caucheteux, Nayak, Bi,
  Marra, McConnell, Keller, Touret, Wu, Wong, Ferrer, Nikolaidis, Allonsius,
  Song, Pintz, Livshits, Wyatt, Esiobu, Choudhary, Mahajan, Garcia-Olano,
  Perino, Hupkes, Lakomkin, AlBadawy, Lobanova, Dinan, Smith, Radenovic,
  Guzmán, Zhang, Synnaeve, Lee, Anderson, Thattai, Nail, Mialon, Pang,
  Cucurell, Nguyen, Korevaar, Xu, Touvron, Zarov, Ibarra, Kloumann, Misra,
  Evtimov, Zhang, Copet, Lee, Geffert, Vranes, Park, Mahadeokar, Shah, van~der
  Linde, Billock, Hong, Lee, Fu, Chi, Huang, Liu, Wang, Yu, Bitton, Spisak,
  Park, Rocca, Johnstun, Saxe, Jia, Alwala, Prasad, Upasani, Plawiak, Li,
  Heafield, Stone, El-Arini, Iyer, Malik, Chiu, Bhalla, Lakhotia,
  Rantala-Yeary, van~der Maaten, Chen, Tan, Jenkins, Martin, Madaan, Malo,
  Blecher, Landzaat, de~Oliveira, Muzzi, Pasupuleti, Singh, Paluri, Kardas,
  Tsimpoukelli, Oldham, Rita, Pavlova, Kambadur, Lewis, Si, Singh, Hassan,
  Goyal, Torabi, Bashlykov, Bogoychev, Chatterji, Zhang, Duchenne, Çelebi,
  Alrassy, Zhang, Li, Vasic, Weng, Bhargava, Dubal, Krishnan, Koura, Xu, He,
  Dong, Srinivasan, Ganapathy, Calderer, Cabral, Stojnic, Raileanu, Maheswari,
  Girdhar, Patel, Sauvestre, Polidoro, Sumbaly, Taylor, Silva, Hou, Wang,
  Hosseini, Chennabasappa, Singh, Bell, Kim, Edunov, Nie, Narang, Raparthy,
  Shen, Wan, Bhosale, Zhang, Vandenhende, Batra, Whitman, Sootla, Collot,
  Gururangan, Borodinsky, Herman, Fowler, Sheasha, Georgiou, Scialom,
  Speckbacher, Mihaylov, Xiao, Karn, Goswami, Gupta, Ramanathan, Kerkez,
  Gonguet, Do, Vogeti, Albiero, Petrovic, Chu, Xiong, Fu, Meers, Martinet,
  Wang, Wang, Tan, Xia, Xie, Jia, Wang, Goldschlag, Gaur, Babaei, Wen, Song,
  Zhang, Li, Mao, Coudert, Yan, Chen, Papakipos, Singh, Srivastava, Jain,
  Kelsey, Shajnfeld, Gangidi, Victoria, Goldstand, Menon, Sharma, Boesenberg,
  Baevski, Feinstein, Kallet, Sangani, Teo, Yunus, Lupu, Alvarado, Caples, Gu,
  Ho, Poulton, Ryan, Ramchandani, Dong, Franco, Goyal, Saraf, Chowdhury,
  Gabriel, Bharambe, Eisenman, Yazdan, James, Maurer, Leonhardi, Huang, Loyd,
  Paola, Paranjape, Liu, Wu, Ni, Hancock, Wasti, Spence, Stojkovic, Gamido,
  Montalvo, Parker, Burton, Mejia, Liu, Wang, Kim, Zhou, Hu, Chu, Cai, Tindal,
  Feichtenhofer, Gao, Civin, Beaty, Kreymer, Li, Adkins, Xu, Testuggine, David,
  Parikh, Liskovich, Foss, Wang, Le, Holland, Dowling, Jamil, Montgomery,
  Presani, Hahn, Wood, Le, Brinkman, Arcaute, Dunbar, Smothers, Sun, Kreuk,
  Tian, Kokkinos, Ozgenel, Caggioni, Kanayet, Seide, Florez, Schwarz, Badeer,
  Swee, Halpern, Herman, Sizov, Guangyi, Zhang, Lakshminarayanan, Inan,
  Shojanazeri, Zou, Wang, Zha, Habeeb, Rudolph, Suk, Aspegren, Goldman, Zhan,
  Damlaj, Molybog, Tufanov, Leontiadis, Veliche, Gat, Weissman, Geboski, Kohli,
  Lam, Asher, Gaya, Marcus, Tang, Chan, Zhen, Reizenstein, Teboul, Zhong, Jin,
  Yang, Cummings, Carvill, Shepard, McPhie, Torres, Ginsburg, Wang, Wu, U,
  Saxena, Khandelwal, Zand, Matosich, Veeraraghavan, Michelena, Li, Jagadeesh,
  Huang, Chawla, Huang, Chen, Garg, A, Silva, Bell, Zhang, Guo, Yu, Moshkovich,
  Wehrstedt, Khabsa, Avalani, Bhatt, Mankus, Hasson, Lennie, Reso, Groshev,
  Naumov, Lathi, Keneally, Liu, Seltzer, Valko, Restrepo, Patel, Vyatskov,
  Samvelyan, Clark, Macey, Wang, Hermoso, Metanat, Rastegari, Bansal,
  Santhanam, Parks, White, Bawa, Singhal, Egebo, Usunier, Mehta, Laptev, Dong,
  Cheng, Chernoguz, Hart, Salpekar, Kalinli, Kent, Parekh, Saab, Balaji,
  Rittner, Bontrager, Roux, Dollar, Zvyagina, Ratanchandani, Yuvraj, Liang,
  Alao, Rodriguez, Ayub, Murthy, Nayani, Mitra, Parthasarathy, Li, Hogan,
  Battey, Wang, Howes, Rinott, Mehta, Siby, Bondu, Datta, Chugh, Hunt, Dhillon,
  Sidorov, Pan, Mahajan, Verma, Yamamoto, Ramaswamy, Lindsay, Lindsay, Feng,
  Lin, Zha, Patil, Shankar, Zhang, Zhang, Wang, Agarwal, Sajuyigbe, Chintala,
  Max, Chen, Kehoe, Satterfield, Govindaprasad, Gupta, Deng, Cho, Virk,
  Subramanian, Choudhury, Goldman, Remez, Glaser, Best, Koehler, Robinson, Li,
  Zhang, Matthews, Chou, Shaked, Vontimitta, Ajayi, Montanez, Mohan, Kumar,
  Mangla, Ionescu, Poenaru, Mihailescu, Ivanov, Li, Wang, Jiang, Bouaziz,
  Constable, Tang, Wu, Wang, Wu, Gao, Kleinman, Chen, Hu, Jia, Qi, Li, Zhang,
  Zhang, Adi, Nam, Yu, Wang, Zhao, Hao, Qian, Li, He, Rait, DeVito, Rosnbrick,
  Wen, Yang, Zhao, and Ma}]{grattafiori2024llama3herdmodels}
Aaron Grattafiori, Abhimanyu Dubey, Abhinav Jauhri, Abhinav Pandey, Abhishek
  Kadian, Ahmad Al-Dahle, Aiesha Letman, Akhil Mathur, Alan Schelten, Alex
  Vaughan, Amy Yang, Angela Fan, Anirudh Goyal, Anthony Hartshorn, Aobo Yang,
  Archi Mitra, Archie Sravankumar, Artem Korenev, Arthur Hinsvark, and 542
  others. 2024.
\newblock \href {https://arxiv.org/abs/2407.21783} {The llama 3 herd of
  models}.
\newblock \emph{Preprint}, arXiv:2407.21783.

\bibitem[{Han et~al.(2024)Han, Collier, Buntine, and
  Shareghi}]{han-etal-2024-pive}
Jiuzhou Han, Nigel Collier, Wray Buntine, and Ehsan Shareghi. 2024.
\newblock \href {https://doi.org/10.18653/v1/2024.findings-acl.400} {{P}i{V}e:
  Prompting with iterative verification improving graph-based generative
  capability of {LLM}s}.
\newblock In \emph{Findings of the Association for Computational Linguistics:
  ACL 2024}, pages 6702--6718, Bangkok, Thailand. Association for Computational
  Linguistics.

\bibitem[{Hu et~al.(2022)Hu, Shen, Wallis, Allen-Zhu, Li, Wang, Wang, and
  Chen}]{hu2022lora}
Edward~J. Hu, Yelong Shen, Phillip Wallis, Zeyuan Allen-Zhu, Yuanzhi Li, Shean
  Wang, Lu~Wang, and Weizhu Chen. 2022.
\newblock \href
  {https://www.microsoft.com/en-us/research/publication/lora-low-rank-adaptation-of-large-language-models/}
  {Lora: Low-rank adaptation of large language models}.
\newblock In \emph{ICLR 2022}.

\bibitem[{Huidrom et~al.(2025)Huidrom, Lorandi, Mille, Thomson, and
  Belz}]{huidrom-etal-2025-assessing}
Rudali Huidrom, Michela Lorandi, Simon Mille, Craig Thomson, and Anya Belz.
  2025.
\newblock \href {https://aclanthology.org/2025.inlg-main.6/} {Assessing
  semantic consistency in {D}ata{-}to{-}{T}ext generation: A meta-evaluation of
  textual, semantic and model-based metrics}.
\newblock In \emph{Proceedings of the 18th International Natural Language
  Generation Conference}, pages 98--107, Hanoi, Vietnam. Association for
  Computational Linguistics.

\bibitem[{Krippendorff(1970)}]{krippendorff1970}
Klaus Krippendorff. 1970.
\newblock \href {https://doi.org/10.1177/001316447003000105} {Estimating the
  reliability, systematic error and random error of interval data}.
\newblock \emph{Educational and Psychological Measurement}, 30(1):61--70.

\bibitem[{Li et~al.(2025)Li, Dai, and Li}]{li-etal-2025-generating}
Zexuan Li, Hongliang Dai, and Piji Li. 2025.
\newblock \href {https://doi.org/10.18653/v1/2025.acl-long.35} {Generating
  diverse training samples for relation extraction with large language models}.
\newblock In \emph{Proceedings of the 63rd Annual Meeting of the Association
  for Computational Linguistics (Volume 1: Long Papers)}, pages 713--726,
  Vienna, Austria. Association for Computational Linguistics.

\bibitem[{Lin(2004)}]{lin-2004-rouge}
Chin-Yew Lin. 2004.
\newblock \href {https://aclanthology.org/W04-1013/} {{ROUGE}: A package for
  automatic evaluation of summaries}.
\newblock In \emph{Text Summarization Branches Out}, pages 74--81, Barcelona,
  Spain. Association for Computational Linguistics.

\bibitem[{Madaan et~al.(2023)Madaan, Tandon, Gupta, Hallinan, Gao, Wiegreffe,
  Alon, Dziri, Prabhumoye, Yang, Gupta, Majumder, Hermann, Welleck,
  Yazdanbakhsh, and Clark}]{madaan2023selfrefine}
Aman Madaan, Niket Tandon, Prakhar Gupta, Skyler Hallinan, Luyu Gao, Sarah
  Wiegreffe, Uri Alon, Nouha Dziri, Shrimai Prabhumoye, Yiming Yang, Shashank
  Gupta, Bodhisattwa~Prasad Majumder, Katherine Hermann, Sean Welleck, Amir
  Yazdanbakhsh, and Peter Clark. 2023.
\newblock \href {https://openreview.net/forum?id=S37hOerQLB} {Self-refine:
  Iterative refinement with self-feedback}.
\newblock In \emph{Thirty-seventh Conference on Neural Information Processing
  Systems}.

\bibitem[{Martinez et~al.(2025)Martinez, Parmentier, and
  Gardent}]{martinez-etal-2025-semantic}
William~Soto Martinez, Yannick Parmentier, and Claire Gardent. 2025.
\newblock \href {https://doi.org/10.18653/v1/2025.findings-acl.542} {Semantic
  evaluation of multilingual data-to-text generation via {NLI} fine-tuning:
  Precision, recall and f1 scores}.
\newblock In \emph{Findings of the Association for Computational Linguistics:
  ACL 2025}, pages 10407--10427, Vienna, Austria. Association for Computational
  Linguistics.

\bibitem[{Min et~al.(2023)Min, Krishna, Lyu, Lewis, Yih, Koh, Iyyer,
  Zettlemoyer, and Hajishirzi}]{min-etal-2023-factscore}
Sewon Min, Kalpesh Krishna, Xinxi Lyu, Mike Lewis, Wen-tau Yih, Pang Koh, Mohit
  Iyyer, Luke Zettlemoyer, and Hannaneh Hajishirzi. 2023.
\newblock \href {https://doi.org/10.18653/v1/2023.emnlp-main.741}
  {{FA}ct{S}core: Fine-grained atomic evaluation of factual precision in long
  form text generation}.
\newblock In \emph{Proceedings of the 2023 Conference on Empirical Methods in
  Natural Language Processing}, pages 12076--12100, Singapore. Association for
  Computational Linguistics.

\bibitem[{Novikova et~al.(2017)Novikova, Du{\v{s}}ek, and
  Rieser}]{novikova-etal-2017-e2e}
Jekaterina Novikova, Ond{\v{r}}ej Du{\v{s}}ek, and Verena Rieser. 2017.
\newblock \href {https://doi.org/10.18653/v1/W17-5525} {The {E}2{E} dataset:
  New challenges for end-to-end generation}.
\newblock In \emph{Proceedings of the 18th Annual {SIG}dial Meeting on
  Discourse and Dialogue}, pages 201--206, Saarbr{\"u}cken, Germany.
  Association for Computational Linguistics.

\bibitem[{OpenAI et~al.(2024)OpenAI, Achiam, Adler, Agarwal, Ahmad, Akkaya,
  Aleman, Almeida, Altenschmidt, Altman, Anadkat, Avila, Babuschkin, Balaji,
  Balcom, Baltescu, Bao, Bavarian, Belgum, Bello, Berdine, Bernadett-Shapiro,
  Berner, Bogdonoff, Boiko, Boyd, Brakman, Brockman, Brooks, Brundage, Button,
  Cai, Campbell, Cann, Carey, Carlson, Carmichael, Chan, Chang, Chantzis, Chen,
  Chen, Chen, Chen, Chen, Chess, Cho, Chu, Chung, Cummings, Currier, Dai,
  Decareaux, Degry, Deutsch, Deville, Dhar, Dohan, Dowling, Dunning, Ecoffet,
  Eleti, Eloundou, Farhi, Fedus, Felix, Fishman, Forte, Fulford, Gao, Georges,
  Gibson, Goel, Gogineni, Goh, Gontijo-Lopes, Gordon, Grafstein, Gray, Greene,
  Gross, Gu, Guo, Hallacy, Han, Harris, He, Heaton, Heidecke, Hesse, Hickey,
  Hickey, Hoeschele, Houghton, Hsu, Hu, Hu, Huizinga, Jain, Jain, Jang, Jiang,
  Jiang, Jin, Jin, Jomoto, Jonn, Jun, Kaftan, Łukasz Kaiser, Kamali,
  Kanitscheider, Keskar, Khan, Kilpatrick, Kim, Kim, Kim, Kirchner, Kiros,
  Knight, Kokotajlo, Łukasz Kondraciuk, Kondrich, Konstantinidis, Kosic,
  Krueger, Kuo, Lampe, Lan, Lee, Leike, Leung, Levy, Li, Lim, Lin, Lin, Litwin,
  Lopez, Lowe, Lue, Makanju, Malfacini, Manning, Markov, Markovski, Martin,
  Mayer, Mayne, McGrew, McKinney, McLeavey, McMillan, McNeil, Medina, Mehta,
  Menick, Metz, Mishchenko, Mishkin, Monaco, Morikawa, Mossing, Mu, Murati,
  Murk, Mély, Nair, Nakano, Nayak, Neelakantan, Ngo, Noh, Ouyang, O'Keefe,
  Pachocki, Paino, Palermo, Pantuliano, Parascandolo, Parish, Parparita,
  Passos, Pavlov, Peng, Perelman, de~Avila Belbute~Peres, Petrov,
  de~Oliveira~Pinto, Michael, Pokorny, Pokrass, Pong, Powell, Power, Power,
  Proehl, Puri, Radford, Rae, Ramesh, Raymond, Real, Rimbach, Ross, Rotsted,
  Roussez, Ryder, Saltarelli, Sanders, Santurkar, Sastry, Schmidt, Schnurr,
  Schulman, Selsam, Sheppard, Sherbakov, Shieh, Shoker, Shyam, Sidor, Sigler,
  Simens, Sitkin, Slama, Sohl, Sokolowsky, Song, Staudacher, Such, Summers,
  Sutskever, Tang, Tezak, Thompson, Tillet, Tootoonchian, Tseng, Tuggle,
  Turley, Tworek, Uribe, Vallone, Vijayvergiya, Voss, Wainwright, Wang, Wang,
  Wang, Ward, Wei, Weinmann, Welihinda, Welinder, Weng, Weng, Wiethoff,
  Willner, Winter, Wolrich, Wong, Workman, Wu, Wu, Wu, Xiao, Xu, Yoo, Yu, Yuan,
  Zaremba, Zellers, Zhang, Zhang, Zhao, Zheng, Zhuang, Zhuk, and
  Zoph}]{openai2024gpt4technicalreport}
OpenAI, Josh Achiam, Steven Adler, Sandhini Agarwal, Lama Ahmad, Ilge Akkaya,
  Florencia~Leoni Aleman, Diogo Almeida, Janko Altenschmidt, Sam Altman,
  Shyamal Anadkat, Red Avila, Igor Babuschkin, Suchir Balaji, Valerie Balcom,
  Paul Baltescu, Haiming Bao, Mohammad Bavarian, Jeff Belgum, and 262 others.
  2024.
\newblock \href {https://arxiv.org/abs/2303.08774} {Gpt-4 technical report}.
\newblock \emph{Preprint}, arXiv:2303.08774.

\bibitem[{Papineni et~al.(2002)Papineni, Roukos, Ward, and
  Zhu}]{papineni-etal-2002-bleu}
Kishore Papineni, Salim Roukos, Todd Ward, and Wei-Jing Zhu. 2002.
\newblock \href {https://doi.org/10.3115/1073083.1073135} {{B}leu: a method for
  automatic evaluation of machine translation}.
\newblock In \emph{Proceedings of the 40th Annual Meeting of the Association
  for Computational Linguistics}, pages 311--318, Philadelphia, Pennsylvania,
  USA. Association for Computational Linguistics.

\bibitem[{Popovi{\'c}(2015)}]{popovic-2015-chrf}
Maja Popovi{\'c}. 2015.
\newblock \href {https://doi.org/10.18653/v1/W15-3049} {chr{F}: character
  n-gram {F}-score for automatic {MT} evaluation}.
\newblock In \emph{Proceedings of the Tenth Workshop on Statistical Machine
  Translation}, pages 392--395, Lisbon, Portugal. Association for Computational
  Linguistics.

\bibitem[{Rebuffel et~al.(2021)Rebuffel, Scialom, Soulier, Piwowarski,
  Lamprier, Staiano, Scoutheeten, and Gallinari}]{rebuffel-etal-2021-data}
Clement Rebuffel, Thomas Scialom, Laure Soulier, Benjamin Piwowarski, Sylvain
  Lamprier, Jacopo Staiano, Geoffrey Scoutheeten, and Patrick Gallinari. 2021.
\newblock \href {https://doi.org/10.18653/v1/2021.emnlp-main.633}
  {Data-{Q}uest{E}val: A referenceless metric for data-to-text semantic
  evaluation}.
\newblock In \emph{Proceedings of the 2021 Conference on Empirical Methods in
  Natural Language Processing}, pages 8029--8036, Online and Punta Cana,
  Dominican Republic. Association for Computational Linguistics.

\bibitem[{Reimers and Gurevych(2019)}]{reimers-2019-sentence-bert}
Nils Reimers and Iryna Gurevych. 2019.
\newblock \href {http://arxiv.org/abs/1908.10084} {Sentence-bert: Sentence
  embeddings using siamese bert-networks}.
\newblock In \emph{Proceedings of the 2019 Conference on Empirical Methods in
  Natural Language Processing}. Association for Computational Linguistics.

\bibitem[{Singh et~al.(2026)Singh, Fry, Perelman, Tart, Ganesh, El-Kishky,
  McLaughlin, Low, Ostrow, Ananthram, Nathan, Luo, Helyar, Madry, Efremov,
  Spyra, Baker-Whitcomb, Beutel, Karpenko, Makelov, Neitz, Wei, Barr,
  Kirchmeyer, Ivanov, Christakis, Gillespie, Tam, Bennett, Wan, Huang,
  Sandjideh, Yang, Kumar, Saraiva, Vallone, Gheorghe, Garcia, Braunstein, Liu,
  Schmidt, Mereskin, Mishchenko, Applebaum, Rogerson, Rajan, Wei, Kotha,
  Srivastava, Agrawal, Vijayvergiya, Tyra, Nair, Nayak, Eggers, Ji, Hoover,
  Chen, Chen, Barak, Minaiev, Hao, Baker, Lightcap, McKinzie, Wang, Quinn,
  Fioca, Hsu, Yang, Yu, Zhang, Brenner, Zetino, Raymond, Lugaresi, Paz, Hudson,
  Whitney, Li, Chen, Cole, Voss, Ding, Shen, Huang, Colby, Hallacy, Koch, Lu,
  Kaplan, Kim, Minott-Henriques, Frey, Yu, Czarnecki, Reid, Wei, Decareaux,
  Scheau, Zhang, Forbes, Tang, Goldberg, Roberts, Palmie, Kappler, Levine,
  Wright, Leo, Lin, Robinson, Grabb, Chen, Lim, Salama, Bhattacharjee, Tsipras,
  Li, Yu, Strouse, Williams, Hunn, Bayes, Arbus, Akyurek, Le, Widmann, Yani,
  Proehl, Sert, Cheung, Schwartz, Han, Jiang, Mitchell, Sigler, Wallace,
  Ritter, Kavanaugh, Mays, Nikishin, Li, Such, de~Avila Belbute~Peres, Raso,
  Bekerman, Tsimpourlas, Chantzis, Song, Zhang, Raila, McGrath, Briggs, Yang,
  Parascandolo, Chabot, Kim, Zhao, Valiant, Leclerc, Salman, Wang, Sheng,
  Jiang, Wang, Jin, Sikchi, Schmidt, Aspegren, Chen, Qiu, Lightman, Covert,
  Kivlichan, Silber, Sohl, Hammoud, Clavera, Lan, Akkaya, Kostrikov, Kofman,
  Etinger, Singal, Hehir, Huh, Pan, Wilczynski, Pachocki, Lee, Quinn, Kiros,
  Kalra, Samaroo, Wang, Wolfe, Chen, Wang, Harb, Han, Wang, Zhao, Chen, Yang,
  Tworek, Chand, Landon, Liang, Lin, Liu, Wang, Tang, Yin, Jang, Morris, Flynn,
  Ferstad, Heidecke, Fishbein, Hallman, Grant, Chien, Gordon, Park, Liss,
  Kraaijeveld, Guay, Mo, Lawson, McGrath, Vendrow, Jiao, Lee, Steele, Wang,
  Mao, Chen, Hayashi, Xiao, Salahi, Wu, Sekhri, Sharma, Singhal, Li, Nguyen,
  Gu-Lemberg, King, Liu, Stone, Yu, Ying, Georgiev, Lim, Tirumala, Miller,
  Ahmad, Lv, Clare, Fauconnet, Itow, Yang, Romaniuk, Anise, Byron, Pathak,
  Maksin, Lo, Ho, Jing, Wu, Xiong, Mamitsuka, Yang, McCallum, Held, Bourgeois,
  Engstrom, Kuhn, Feuvrier, Zhang, Switzer, Kondraciuk, Kaiser, Joglekar,
  Singh, Shah, Stratta, Williams, Chen, Sun, Cayton, Li, Zhang, Aljubeh,
  Nichols, Haines, Schwarzer, Gupta, Shah, Guan, Huang, Dong, Wang, Glaese,
  Carroll, Lampe, Malek, Sharman, Zhang, Wang, Pokrass, Florian, Pavlov, Wang,
  Chen, Wang, Feng, Bavarian, Lin, Abdool, Rohaninejad, Soto, Staudacher,
  LaFontaine, Marwell, Liu, Preston, Turley, Ansman, Blades, Pancha, Mikhaylin,
  Felix, Handa, Rai, Keskar, Brown, Nachum, Boiko, Murk, Watkins, Gleeson,
  Mishkin, Lesiewicz, Baltescu, Belov, Zhokhov, Pronin, Guo, Thacker, Liu,
  Yuan, Liu, Dias, Puckett, Arora, Mullapudi, Gaon, Miyara, Song, Aggarwal,
  Marsan, Yemiru, Xiong, Kshirsagar, Nuttall, Tsiupa, Eldan, Wang, James, Ziv,
  Shu, Nigmatullin, Jain, Talaie, Altman, Arnesen, Toizer, Toyer, Miserendino,
  Agarwal, Yoo, Heon, Ethersmith, Grove, Taylor, Bubeck, Banesiu, Amdo, Zhao,
  Wu, Santurkar, Zhao, Chaudhuri, Krishnaswamy, Shuaiqi, Xia, Cheng, Anadkat,
  Fishman, Tobin, Fu, Jain, Mei, Egoian, Kim, Golden, Mah, Lin, Imm, Sharpe,
  Yadlowsky, Choudhry, Eum, Sanjeev, Khan, Stramer, Wang, Xin, Gogineni,
  Christianson, Sanders, Patwardhan, Degry, Shadwell, Fu, Gao, Garipov,
  Sriskandarajah, Sherbakov, Korbak, Kaftan, Hiratsuka, Wang, Song, Zhao,
  Peterson, Kharitonov, Chernova, Kosaraju, Kuo, Pong, Verma, Petrov, Jiang,
  Zhang, Zhou, Xie, Zhan, McCabe, DePue, Ellsworth, Bain, Thompson, Chen, Qi,
  Xiang, Shi, Dubois, Yu, Khakbaz, Wu, Qian, Lee, Chen, Zhang, Xiong, Tian,
  Cha, Bai, Yang, Yuan, Li, Zhang, Yang, Jin, Jiang, Wang, Wang, Liu,
  Stubenvoll, Dou, Wu, and Wang}]{singh2026openaigpt5card}
Aaditya Singh, Adam Fry, Adam Perelman, Adam Tart, Adi Ganesh, Ahmed El-Kishky,
  Aidan McLaughlin, Aiden Low, AJ~Ostrow, Akhila Ananthram, Akshay Nathan, Alan
  Luo, Alec Helyar, Aleksander Madry, Aleksandr Efremov, Aleksandra Spyra, Alex
  Baker-Whitcomb, Alex Beutel, Alex Karpenko, and 467 others. 2026.
\newblock \href {https://arxiv.org/abs/2601.03267} {Openai gpt-5 system card}.
\newblock \emph{Preprint}, arXiv:2601.03267.

\bibitem[{Song and Gardent(2025)}]{song-gardent-2025-mucal}
Yifei Song and Claire Gardent. 2025.
\newblock \href {https://doi.org/10.18653/v1/2025.emnlp-main.720} {{M}u{CAL}:
  Contrastive alignment for preference-driven {KG}-to-text generation}.
\newblock In \emph{Proceedings of the 2025 Conference on Empirical Methods in
  Natural Language Processing}, pages 14227--14270, Suzhou, China. Association
  for Computational Linguistics.

\bibitem[{Song et~al.(2025)Song, Martinez, Nikiforovskaya, Chapple, and
  Gardent}]{song-etal-2025-multilingual}
Yifei Song, William~Soto Martinez, Anna Nikiforovskaya, Evan Chapple, and
  Claire Gardent. 2025.
\newblock \href {https://doi.org/10.18653/v1/2025.findings-emnlp.60}
  {Multilingual verbalisation of knowledge graphs}.
\newblock In \emph{Findings of the Association for Computational Linguistics:
  EMNLP 2025}, pages 1111--1162, Suzhou, China. Association for Computational
  Linguistics.

\bibitem[{Team et~al.(2025)Team, Kamath, Ferret, Pathak, Vieillard, Merhej,
  Perrin, Matejovicova, Ramé, Rivière, Rouillard, Mesnard, Cideron, bastien
  Grill, Ramos, Yvinec, Casbon, Pot, Penchev, Liu, Visin, Kenealy, Beyer, Zhai,
  Tsitsulin, Busa-Fekete, Feng, Sachdeva, Coleman, Gao, Mustafa, Barr,
  Parisotto, Tian, Eyal, Cherry, Peter, Sinopalnikov, Bhupatiraju, Agarwal,
  Kazemi, Malkin, Kumar, Vilar, Brusilovsky, Luo, Steiner, Friesen, Sharma,
  Sharma, Gilady, Goedeckemeyer, Saade, Feng, Kolesnikov, Bendebury, Abdagic,
  Vadi, György, Pinto, Das, Bapna, Miech, Yang, Paterson, Shenoy, Chakrabarti,
  Piot, Wu, Shahriari, Petrini, Chen, Lan, Choquette-Choo, Carey, Brick,
  Deutsch, Eisenbud, Cattle, Cheng, Paparas, Sreepathihalli, Reid, Tran, Zelle,
  Noland, Huizenga, Kharitonov, Liu, Amirkhanyan, Cameron, Hashemi,
  Klimczak-Plucińska, Singh, Mehta, Lehri, Hazimeh, Ballantyne, Szpektor,
  Nardini, Pouget-Abadie, Chan, Stanton, Wieting, Lai, Orbay, Fernandez,
  Newlan, yeong Ji, Singh, Black, Yu, Hui, Vodrahalli, Greff, Qiu, Valentine,
  Coelho, Ritter, Hoffman, Watson, Chaturvedi, Moynihan, Ma, Babar, Noy, Byrd,
  Roy, Momchev, Chauhan, Sachdeva, Bunyan, Botarda, Caron, Rubenstein,
  Culliton, Schmid, Sessa, Xu, Stanczyk, Tafti, Shivanna, Wu, Pan, Rokni,
  Willoughby, Vallu, Mullins, Jerome, Smoot, Girgin, Iqbal, Reddy, Sheth,
  Põder, Bhatnagar, Panyam, Eiger, Zhang, Liu, Yacovone, Liechty, Kalra, Evci,
  Misra, Roseberry, Feinberg, Kolesnikov, Han, Kwon, Chen, Chow, Zhu, Wei,
  Egyed, Cotruta, Giang, Kirk, Rao, Black, Babar, Lo, Moreira, Martins,
  Sanseviero, Gonzalez, Gleicher, Warkentin, Mirrokni, Senter, Collins, Barral,
  Ghahramani, Hadsell, Matias, Sculley, Petrov, Fiedel, Shazeer, Vinyals, Dean,
  Hassabis, Kavukcuoglu, Farabet, Buchatskaya, Alayrac, Anil, Dmitry, Lepikhin,
  Borgeaud, Bachem, Joulin, Andreev, Hardin, Dadashi, and
  Hussenot}]{gemmateam2025gemma3technicalreport}
Gemma Team, Aishwarya Kamath, Johan Ferret, Shreya Pathak, Nino Vieillard,
  Ramona Merhej, Sarah Perrin, Tatiana Matejovicova, Alexandre Ramé, Morgane
  Rivière, Louis Rouillard, Thomas Mesnard, Geoffrey Cideron, Jean bastien
  Grill, Sabela Ramos, Edouard Yvinec, Michelle Casbon, Etienne Pot, Ivo
  Penchev, and 197 others. 2025.
\newblock \href {https://arxiv.org/abs/2503.19786} {Gemma 3 technical report}.
\newblock \emph{Preprint}, arXiv:2503.19786.

\bibitem[{Wang et~al.(2023)Wang, Liang, Meng, Sun, Shi, Li, Xu, Qu, and
  Zhou}]{wang-etal-2023-chatgpt}
Jiaan Wang, Yunlong Liang, Fandong Meng, Zengkui Sun, Haoxiang Shi, Zhixu Li,
  Jinan Xu, Jianfeng Qu, and Jie Zhou. 2023.
\newblock \href {https://doi.org/10.18653/v1/2023.newsum-1.1} {Is {C}hat{GPT} a
  good {NLG} evaluator? a preliminary study}.
\newblock In \emph{Proceedings of the 4th New Frontiers in Summarization
  Workshop}, pages 1--11, Singapore. Association for Computational Linguistics.

\bibitem[{Wei et~al.(2024)Wei, Yang, Song, Lu, Hu, Huang, Tran, Peng, Liu,
  Huang, Du, and Le}]{wei2024longform}
Jerry Wei, Chengrun Yang, Xinying Song, Yifeng Lu, Nathan~Zixia Hu, Jie Huang,
  Dustin Tran, Daiyi Peng, Ruibo Liu, Da~Huang, Cosmo Du, and Quoc~V Le. 2024.
\newblock \href {https://openreview.net/forum?id=4M9f8VMt2C} {Long-form
  factuality in large language models}.
\newblock In \emph{The Thirty-eighth Annual Conference on Neural Information
  Processing Systems}.

\bibitem[{Xiao et~al.(2024)Xiao, Liu, Zhang, Muennighoff, Lian, and
  Nie}]{xiao23bge}
Shitao Xiao, Zheng Liu, Peitian Zhang, Niklas Muennighoff, Defu Lian, and
  Jian-Yun Nie. 2024.
\newblock \href {https://doi.org/10.1145/3626772.3657878} {C-pack: Packed
  resources for general chinese embeddings}.
\newblock In \emph{Proceedings of the 47th International ACM SIGIR Conference
  on Research and Development in Information Retrieval}, SIGIR '24, page
  641–649, New York, NY, USA. Association for Computing Machinery.

\bibitem[{Xu et~al.(2023)Xu, Wang, Pan, Song, Freitag, Wang, and
  Li}]{xu-etal-2023-instructscore}
Wenda Xu, Danqing Wang, Liangming Pan, Zhenqiao Song, Markus Freitag, William
  Wang, and Lei Li. 2023.
\newblock \href {https://doi.org/10.18653/v1/2023.emnlp-main.365}
  {{INSTRUCTSCORE}: Towards explainable text generation evaluation with
  automatic feedback}.
\newblock In \emph{Proceedings of the 2023 Conference on Empirical Methods in
  Natural Language Processing}, pages 5967--5994, Singapore. Association for
  Computational Linguistics.

\bibitem[{Yang et~al.(2025)Yang, Li, Yang, Zhang, Hui, Zheng, Yu, Gao, Huang,
  Lv, Zheng, Liu, Zhou, Huang, Hu, Ge, Wei, Lin, Tang, Yang, Tu, Zhang, Yang,
  Yang, Zhou, Zhou, Lin, Dang, Bao, Yang, Yu, Deng, Li, Xue, Li, Zhang, Wang,
  Zhu, Men, Gao, Liu, Luo, Li, Tang, Yin, Ren, Wang, Zhang, Ren, Fan, Su,
  Zhang, Zhang, Wan, Liu, Wang, Cui, Zhang, Zhou, and
  Qiu}]{yang2025qwen3technicalreport}
An~Yang, Anfeng Li, Baosong Yang, Beichen Zhang, Binyuan Hui, Bo~Zheng, Bowen
  Yu, Chang Gao, Chengen Huang, Chenxu Lv, Chujie Zheng, Dayiheng Liu, Fan
  Zhou, Fei Huang, Feng Hu, Hao Ge, Haoran Wei, Huan Lin, Jialong Tang, and 41
  others. 2025.
\newblock \href {https://arxiv.org/abs/2505.09388} {Qwen3 technical report}.
\newblock \emph{Preprint}, arXiv:2505.09388.

\bibitem[{Yang et~al.(2023)Yang, Xu, Yu, and Xia}]{yang-etal-2023-unicoqe}
Zinong Yang, Feng Xu, Jianfei Yu, and Rui Xia. 2023.
\newblock \href {https://doi.org/10.18653/v1/2023.findings-acl.775}
  {{U}ni{COQE}: Unified comparative opinion quintuple extraction as a set}.
\newblock In \emph{Findings of the Association for Computational Linguistics:
  ACL 2023}, pages 12229--12240, Toronto, Canada. Association for Computational
  Linguistics.

\bibitem[{Zhang et~al.(2023)Zhang, Balalau, and
  Manolescu}]{zhang-etal-2023-factspotter}
Kun Zhang, Oana Balalau, and Ioana Manolescu. 2023.
\newblock \href {https://doi.org/10.18653/v1/2023.findings-emnlp.672}
  {{F}act{S}potter: Evaluating the factual faithfulness of graph-to-text
  generation}.
\newblock In \emph{Findings of the Association for Computational Linguistics:
  EMNLP 2023}, pages 10025--10042, Singapore. Association for Computational
  Linguistics.

\bibitem[{Zhang et~al.(2025)Zhang, Balalau, and
  Manolescu}]{zhang-etal-2025-structured}
Kun Zhang, Oana Balalau, and Ioana Manolescu. 2025.
\newblock \href {https://doi.org/10.18653/v1/2025.findings-acl.46} {Structured
  discourse representation for factual consistency verification}.
\newblock In \emph{Findings of the Association for Computational Linguistics:
  ACL 2025}, pages 820--838, Vienna, Austria. Association for Computational
  Linguistics.

\bibitem[{Zhang et~al.(2020)Zhang, Kishore, Wu, Weinberger, and
  Artzi}]{Zhang*2020BERTScore:}
Tianyi Zhang, Varsha Kishore, Felix Wu, Kilian~Q. Weinberger, and Yoav Artzi.
  2020.
\newblock \href {https://openreview.net/forum?id=SkeHuCVFDr} {Bertscore:
  Evaluating text generation with bert}.
\newblock In \emph{International Conference on Learning Representations}.

\bibitem[{Zhao et~al.(2026)Zhao, Min, Wu, Li, Sun, Cai, Wang, Chen, and
  Penn}]{zhao-etal-2026-step-pruning}
Jinman Zhao, Erxue Min, Hui Wu, Ziheng Li, Zexu Sun, Hengyi Cai, Shuaiqiang
  Wang, Xu~Chen, and Gerald Penn. 2026.
\newblock \href {https://doi.org/10.1609/aaai.v40i41.40798} {Beyond step
  pruning: information theory based step-level optimization for self-refining
  large language models}.
\newblock In \emph{Proceedings of the Fortieth AAAI Conference on Artificial
  Intelligence and Thirty-Eighth Conference on Innovative Applications of
  Artificial Intelligence and Sixteenth Symposium on Educational Advances in
  Artificial Intelligence}, AAAI'26/IAAI'26/EAAI'26. AAAI Press.

\bibitem[{Zhao and Zhang(2024)}]{zhao2024large}
Jinman Zhao and Xueyan Zhang. 2024.
\newblock \href {https://openreview.net/forum?id=wLQ3I0F1oj} {Large language
  model is not a (multilingual) compositional relation reasoner}.
\newblock In \emph{First Conference on Language Modeling}.

\end{thebibliography}

\appendix

\section{Label Definition and Examples}
\label{app:labelsdef}
\paragraph{Omitted} \( d_i \) exists in \(\mathbf D\) but is left unmentioned in \(\mathbf T\). Taking knowledge graph triples as examples, when \textbf{two or more} elements (subject, predicate, or object) are inconsistent with the content expressed in the text, the triple is considered omitted, as it has been severely distorted to the point that it cannot be recognized as a verbalisation of \( d_i \). For instance, if \(\mathbf D\) includes \( \langle \text{``Mars''}, \text{``has moons''}, \text{``Phobos and Deimos''} \rangle \) but \(\mathbf T\) only states ``Mars is a terrestrial planet,'' then the triple about Mars' moons is omitted, since two elements (the predicate and object) are inconsistent with the text content. Similarly, if a triple \( \langle \text{``Paris''}, \text{``capital of''}, \text{``France''} \rangle \) is verbalised as ``Berlin is the capital of Germany,'' it is also regarded as omitted because both the subject and object have been replaced.

\paragraph{Incorrect}: \( d_i \) is verbalised in \(\mathbf T\) but with a distortion that preserves most of the original information while introducing a factual error. In such cases, the data unit is clearly verbalised and can be traced back to \( d_i \). Incorrect verbalisation specifically includes two scenarios: (1) \textit{incorrect element}, where \textbf{only one} element (subject, predicate, or object) in the triple is replaced with an incorrect element. For example, verbalising \( \langle \text{``Confucius''}, \text{``birth place''}, \text{``Shandong''} \rangle \) as ``Confucius was born in Korea'' is incorrect, as one element in the triple (the object) has been replaced. (2) \textit{entity reversal}, where the subject and object are swapped. For example, if \( \langle \text{``United States''}, \text{``demonym''}, \text{``American''} \rangle \) is verbalised as ``The United States is the demonym of Americans,'' it is also incorrect, as the subject and object are interchanged.

\color{black}

\paragraph{Extra}: information present in \(\mathbf T\) that has no corresponding unit in \(\mathbf D\), representing content the text introduces beyond what the data specifies. For instance, if \(\mathbf D\) contains only triples about Albert Einstein’s private life, while \(\mathbf T\) states ``Einstein developed the theory of relativity,'' the scientific achievement is unsupported by the data and is therefore marked as extra content.

\paragraph{Correct}: the data item \( d_i \) is accurately and completely verbalised in the text \(\mathbf T\), meaning the text expresses the exact information in the data without omissions, inaccuracies, or extra content. For example, the triple \( \langle \text{``Galápagos Islands''}, \text{``located in''}, \text{``Ecuador''} \rangle \) is correctly verbalised in ``The Galápagos Islands lie within the territory of Ecuador.''

\section{Model Description}
\label{appendix:model_v}

The descriptions of base models used for our verification are as follows. 

\paragraph{Gemma3.}
Gemma3 \citep{gemmateam2025gemma3technicalreport}
comprises language models ranging from 270M to 27B parameters, designed for long-context and multilingual applications.
The smaller variants (270M and 1B) support a context length of 32K tokens, while the larger ones (4B and above) support up to 128K tokens.
To reduce memory overhead for long sequences, Gemma3 adopts a 5:1 ratio of local to global attention layers with a local attention span of 1,024 tokens, substantially reducing KV-cache growth.
Knowledge distillation from larger teacher models is applied during pre-training to improve performance across model scales.
In our experiments, we use variants from 270M to 12B.

\paragraph{Qwen3.}
{Qwen3} \citep{yang2025qwen3technicalreport}
is a model family providing both dense and Mixture-of-Experts (MoE) variants, with parameter scales ranging from 0.6B to 235B.
Its dense models adopt grouped-query attention (GQA) with RMSNorm and QK-Norm for training stability. The smaller variants (0.6B and 1.7B) support a context length of 32K tokens, while the larger ones (4B and above) support up to 128K tokens.
Compared with its predecessor Qwen2.5, Qwen3 expands multilingual coverage from 29 to 119 languages and dialects.
Its pre-training follows a three-stage curriculum, progressing from general language training over 30 trillion tokens to a reasoning-focused stage and finally a long-context adaptation stage.

\begin{figure*}[!h]   
\centering
\begin{minipage}{0.75\textwidth}
\begin{lstlisting}[
  basicstyle=\ttfamily\footnotesize,
  columns=fullflexible,
  breaklines=true,
  frame=single
]
Verify if the data has missing, extra, or incorrect contents, compared with the text
TEXT: {text}
DATA: {triples_formatted}
Find three error types:
1. EXTRA CONTENTS
The text mentions contents NOT captured by any triple.
Example
Text: "Abilene Regional Airport is in Texas and serves Abilene."
Given only: [S] Abilene Regional Airport [P] location [O] Texas
EXTRA: served city is mentioned but not captured in triples
Output: [S] Abilene Regional Airport [P] city served [O] Abilene

2. MISSING DATA
The given triple contains 2 or 3 elements NOT in the text.
Example (3 elements not in text)
Text: "Abilene Regional Airport is in Texas and serves Abilene."
MISSING: [S] Adirondack Regional Airport [P] operated by [O] Port Authority
All 3 elements not mentioned (Adirondack, operated by, Port Authority)

Example (2 elements not in text)
Text: "Abilene Regional Airport is in Texas and serves Abilene."
MISSING: [S] Abilene Regional Airport [P] opened in [O] 1960
Two elements not mentioned: "opened" relation and "1960"

3. INCORRECT TRIPLES
ONE element in the triple is wrong, or the triple is reversed.
Example (entity error):
Text: "Abilene Regional Airport serves the city of Abilene in Texas."
Wrong: [S] Abilene Regional Airport [P] city served [O] Austin
Right: [S] Abilene Regional Airport [P] city served [O] Abilene
Object entity is wrong

Example (reversed error):
Text: "Texas is part of the United States."
Wrong: [S] United States [P] is part of [O] Texas
Right: [S] Texas [P] is part of [O] United States
Subject and object are swapped

RULES:
- Each triple is ONE string with [S] [P] [O] markers
- Use "-" for empty cells in the markdown table
- Multiple errors? Add multiple rows
- No errors? Output "All correct"

The output should be only a markdown table:
| Type | Triple |
| ---- | ------ |
| Extra | [S] subject1 [P] predicate1 [O] object1 |
| Extra | [S] subject2 [P] predicate2 [O] object2 |
| Missing | [S] subject3 [P] predicate3 [O] object3 |
| Missing | [S] subject4 [P] predicate4 [O] object4 |
| Missing | [S] subject5 [P] predicate5 [O] object5 |
| Incorrect | [S] subject6 [P] predicate6 [O] object6 |
Don't output extra explanations.
\end{lstlisting}
\captionof{figure}{Prompt for Error Detection}
\label{fig:prompt}
\end{minipage}
\end{figure*}

\paragraph{Llama~3.}
{Llama~3} \citep{grattafiori2024llama3herdmodels} is a series of language models designed for multilingual generation, reasoning, and tool use.
Llama~3.1 expands the series with an extended context window of 128K tokens and multilingual support across eight languages, with dense variants ranging from 8B to 405B parameters.
Llama~3.2 further introduces lightweight 1B and 3B models derived via structured pruning and knowledge distillation from the Llama~3.1 8B, retaining the same 128K context window while targeting efficient inference.
In our experiments, we use the 1B and 3B variants from Llama~3.2, and the 8B variant from Llama~3.1.

\section{Prompt Baseline for Error Detection}
\label{appendix:prompt_baseline}
Similar to \citet{wang-etal-2023-chatgpt}, we prompt multiple LLMs
(Gemma3-27B,
Qwen3-32B,
Llama3.3-70B,
GPT-4.1 \citep{openai2024gpt4technicalreport},
GPT-5.1 \citep{singh2026openaigpt5card}) as baselines for data-text alignment error detection with the instructions in Figure \ref{fig:prompt}. 

\section{Data Source Description}
\label{app:data}

The descriptions of source data are as follows. 

\paragraph{WebNLG. }
The {WebNLG} shared tasks provide human evaluation annotations across multiple editions, enabling the comparison between automatic metrics and human judgements to evaluate data-text alignment.  
WebNLG 2017\footnote{\url{https://gitlab.com/webnlg/webnlg-human-evaluation}} \citep{gardent-etal-2017-webnlg} 
includes 223 samples from 9 systems, rated on \emph{semantic adequacy}, \emph{structure}, and \emph{fluency}. We compute the correlation of precision, recall, and F1 with human ratings of \emph{semantic adequacy}.  
In WebNLG 2020\footnote{\url{https://github.com/WebNLG/challenge-2020/tree/main/evaluation/human-evaluation}} \citep{webnlg2020}, each system was evaluated on 178 samples with five criteria: \emph{data correctness}, \emph{coverage}, \emph{relevance}, \emph{structure}, and \emph{fluency}. Our correlation analysis focuses on the three semantic dimensions, \emph{data correctness}, \emph{coverage}, and \emph{relevance}. %
The WebNLG 2023 \citep{cripwell-etal-2023-2023} shared task emphasised \emph{omissions}, \emph{additions}, \emph{repetitions}, and \emph{fluency} as human evaluation criteria, with explicit annotations on faithfulness errors on 100 samples. We compare our scores 
against the \textit{precision} and \textit{recall} annotations in the 4L-RP-Human subset of \citet{martinez-etal-2025-semantic}.  

\paragraph{E2E. }
The E2E dataset \citep{novikova-etal-2017-e2e} is a restaurant-domain data-to-text benchmark, where each meaning representation (MR) is a set of slot--value pairs (e.g., \texttt{name}, \texttt{food}, \texttt{area}, \texttt{priceRange}) along with multiple crowdsourced references.
Subsequent work by \citet{dusek-etal-2019-semantic} addresses semantic noise in the crowdsourced references and provides a cleaned E2E release with automatically corrected MRs.
In the original release, the train/dev/test splits contain 6039 distinct MRs paired with 51426 reference texts, while the cleaned release contains 10852 distinct MRs paired with 42517 reference texts.
All E2E experiments in our paper use the cleaned E2E release.
In addition to the parallel MR--text data, the E2E challenge provides crowdsourced human judgements of system outputs \citep{DUSEK2020123}.
Each (MR, output) pair is annotated with a coarse semantic label (\textit{ok}, \textit{missing}, or \textit{added}) as well as overall quality and naturalness scores on 0--100 scale.
The released annotation set covers 630 test MRs and 2979 items over 21 systems.

\paragraph{KELM}
The {KELM} corpus \citep{agarwal-etal-2021-knowledge} provides a large-scale synthetic resource designed for knowledge-enhanced language model pre-training. 
It is built from the English {Wikidata} knowledge graph by loosely aligning triples with texts from their corresponding {Wikipedia} pages, based on entity matching.  
A T5 model is then pretrained on the aligned data and fine-tuned with WebNLG to verbalise the data entries into texts.  
The resulting corpus contains approximately 18M synthetic sentences verbalising about 45M Wikidata triples spanning roughly 1500 relations, offering a broad-coverage benchmark for knowledge-grounded pre-training.

\section{Artifact and Data Statement}

We use publicly released datasets and open-source models under their original research-use conditions. The artifacts in this work are intended solely for research on alignment evaluation and feedback between structured inputs and text. We do not collect new personal data; our experiments rely on public benchmark datasets and structured knowledge verbalizations rather than newly curated personally identifiable information.

\section{Statistics of the Synthetic Data and Details for the Construction}
\label{app:synth-stats}

\begin{table}[ht]
\centering
\small
\setlength{\tabcolsep}{4pt}
\begin{adjustbox}{width=\columnwidth}
\begin{tabular}{lrrrrr}
\toprule
& \multicolumn{2}{c}{Samples} & \multicolumn{3}{c}{Negative error types} \\
\cmidrule(lr){2-3}\cmidrule(lr){4-6}
Data Split & Positive & Negative & Omitted & Incorrect & Extra \\
\midrule
\multicolumn{6}{l}{\textbf{WebNLG}} \\
Train & 7,177 & 28,249 & 13,199 & 17,662 & 12,500 \\
Dev   & 382   & 1,285  & 620    & 671    & 706    \\
Test  & 424   & 1,355  & 561    & 739    & 725    \\
\midrule
\multicolumn{6}{l}{\textbf{E2E}} \\
Train & 1,944 & 7,241 & 3,443 & 3,727 & 4,120 \\
Dev   & 230   & 888   & 432   & 480   & 511   \\
Test  & 270   & 959   & 456   & 507   & 524   \\
\bottomrule
\end{tabular}
\end{adjustbox}
\caption{Data distribution across splits. Error Types (Omitted, Extra, Incorrect) indicate the number of negative samples containing each error type. Different types of errors could co-occur on the same negative sample. }
\label{tab:dataset_stats}
\end{table}

Table~\ref{tab:dataset_stats} summarises the statistics of our synthetic error-detection train/dev/test data derived from WebNLG and E2E, reporting the numbers of positive and negative instances as well as how often each error type (\textit{omitted}, \textit{incorrect}, \textit{extra}) appears in the negative set. To further mitigate residual semantic noise in E2E, we apply an additional reference-quality filter on top of the cleaned release. Specifically, we retain only MR--text pairs whose automatic semantic consistency score exceeds a threshold (MonoLR $F1 \ge 0.7$ in our implementation) and use the resulting subset to construct our synthetic training data. This extra filtering explains why the number of E2E-derived synthetic instances is smaller than the total number of MRs reported for the cleaned E2E release.

\paragraph{Manual audit of synthetic examples.}
We manually inspected 180 synthetic examples,
including 90 from WebNLG and 90 from E2E,
with 60 examples for each target error type.
Perturbation validity measures whether the intended mismatch is valid
and assigned the correct error type.
Label completeness measures whether the data--text pair contains any
additional unlabelled alignment errors.
Table~\ref{tab:synthetic-audit} reports the results.

\begin{table}[t]
\centering
\small
\begin{tabular}{lcc}
\toprule
Target error & Validity & Completeness \\
\midrule
Extra     & 55/60 (91.7\%) & 49/60 (81.7\%) \\
Omitted   & 57/60 (95.0\%) & 52/60 (86.7\%) \\
Incorrect & 57/60 (95.0\%) & 53/60 (88.3\%) \\
\midrule
Overall   & 169/180 (93.9\%) & 154/180 (85.6\%) \\
\bottomrule
\end{tabular}
\caption{Manual audit of 180 synthetic examples.}
\label{tab:synthetic-audit}
\end{table}

\begin{table}[t]
\centering
\small
\setlength{\tabcolsep}{6pt}
\begin{tabular}{lrrr}
\toprule
Data Split & Single & Dual & Triple \\
\midrule
\multicolumn{4}{l}{\textbf{WebNLG}} \\
Train & 15,419 & 10,548 & 2,282 \\
Dev   & 683    & 492    & 110   \\
Test  & 780    & 480    & 95    \\
\midrule
\multicolumn{4}{l}{\textbf{E2E}} \\
Train & 3,863 & 2,707 & 671 \\
Dev   & 454   & 333   & 101 \\
Test  & 518   & 354   & 87  \\
\bottomrule
\end{tabular}
\caption{Error type co-occurrence in negative samples. Single, Dual, and Triple indicate samples containing one, two, or all three error types.}
\label{tab:error_cooccurrence}
\end{table}

\paragraph{Representative examples.}
In each example below,
the text is kept unchanged and only the structured data are modified.

\begin{itemize}
\item \textbf{Extra.}
\emph{Text:} Ajoblanco has almond (of the Rosales order) as one of its
ingredients.
\emph{Original data:}
Ajoblanco~|~ingredient~|~Almond;
Almond~|~order~|~Rosales.
\emph{Perturbation:}
the second triple is removed.
Its unchanged verbalisation is therefore labelled Extra.

\item \textbf{Omitted.}
\emph{Text:} Clowns is a low customer rating coffee shop that serves fast
food. It is in riverside near Clare Hall.
\emph{Original data:}
Clowns~|~area~|~riverside;
Clowns~|~customer rating~|~low;
Clowns~|~eat type~|~coffee shop;
Clowns~|~food~|~Fast food;
Clowns~|~near~|~Clare Hall.
\emph{Perturbation:}
Strada~|~family friendly~|~no is added.
Because it is not expressed in the unchanged text,
it is labelled Omitted.

\item \textbf{Incorrect.}
\emph{Text:} There is a coffee shop named Wildwood that serves French food
near Ranch. The average rating is 3 out of 5 and it has a high price range.
\emph{Original data:}
Wildwood~|~customer rating~|~3 out of 5;
Wildwood~|~eat type~|~coffee shop;
Wildwood~|~food~|~French;
Wildwood~|~near~|~Ranch;
Wildwood~|~price range~|~high.
\emph{Perturbation:}
the last triple is changed to
high~|~price range~|~Wildwood,
and is therefore labelled Incorrect.
\end{itemize}

\begin{table*}[htpb]
\centering
\small
\begin{tabular}{|l|ccc|ccc|ccc|}
\hline
& \multicolumn{3}{c|}{$P_m$ vs $P_h$} & \multicolumn{3}{c|}{{$R_m$ vs $R_h$}} & \multicolumn{3}{c|}{{$F1_m$ vs $F1_h$}} \\
Model& \textit{r}$\uparrow$ & $\rho$$\uparrow$ & $\tau$$\uparrow$ & \textit{r}$\uparrow$ & $\rho$$\uparrow$ & $\tau$$\uparrow$ & \textit{r}$\uparrow$ & $\rho$$\uparrow$ & $\tau$$\uparrow$ \\
\hline
Data-QuestEval & 53.3 & 61.2 & 44.9 & 59.5 & 58.3 & 44.5 & 60.3 & 62.5 & 49.0 \\
MonoLR & 70.7 & 71.0 & 55.9 & 63.5 & 63.7 & 48.7 & 72.3 & 70.6 & 53.1 \\
NLI-Based & 70.7 & 73.4 & 56.1 & 75.5 & 73.2 & 55.9 & 73.5 & 73.3 & 54.1 \\
FactSpotter & — & — & — & \underline{87.6} & 86.5 & 68.3 & — & — & — \\
\hline
Gemma3-27B-Prompt & 38.1 & 28.9 & 19.9 & 37.9 & 35.1 & 25.4 & 39.5 & 36.1 & 25.5 \\
Qwen3-32B-Prompt & 34.9 & 35.9 & 28.3 & 31.1 & 30.4 & 23.3 & 34.7 & 35.0 & 26.3 \\
Llama3.3-70B-Prompt & 39.4 & 32.4 & 24.6 & 41.0 & 35.5 & 26.4 & 41.0 & 32.8 & 24.1 \\
GPT-5.1-Prompt & 82.3 & 80.9 & 68.5 & 79.0 & 76.8 & 66.0 & 81.8 & 78.5 & 65.2 \\
\hline
Gemma3-270M%
& 81.2 & 75.7 & 66.2 & 85.2 & 83.5 & 73.2 & 84.3 & 82.8 & 71.0 \\
Gemma3-1B & 80.6 & 74.3 & 65.8 & 82.7 & 81.5 & 72.2 & 82.8 & 81.3 & 71.1 \\
Gemma3-4B & 84.0 & 79.0 & 67.8 & \underline{87.9} & \underline{87.6} & \underline{75.8} & 87.4 & \underline{87.4} & \underline{73.9} \\
Gemma3-12B & 66.5 & 61.5 & 51.8 & 61.1 & 55.7 & 48.2 & 66.8 & 62.7 & 51.0 \\
\hline
Qwen3-0.6B & 77.9 & 71.8 & 61.0 & 80.5 & 79.3 & 68.1 & 80.1 & 78.4 & 66.1 \\
Qwen3-1.7B & 72.2 & 66.3 & 55.6 & 74.9 & 74.0 & 63.2 & 74.4 & 72.5 & 59.9 \\
Qwen3-4B & \underline{89.6} & \underline{88.5} & \textbf{76.5} & 87.0 & \underline{86.9} & \underline{74.2} & \underline{88.5} & \underline{88.1} & \underline{74.4} \\
Qwen3-8B & \textbf{91.4} & \textbf{88.6} & \underline{76.3} & \textbf{90.1} & \textbf{88.9} & \textbf{77.2} & \textbf{91.1} & \textbf{89.6} & \textbf{76.1} \\
Qwen3-14B & 68.9 & 60.2 & 50.3 & 70.3 & 64.8 & 54.4 & 75.2 & 69.5 & 55.0 \\
\hline
Llama3.2-1B & 76.6 & 71.0 & 61.9 & 78.0 & 77.0 & 67.0 & 78.3 & 76.7 & 65.7 \\
Llama3.2-3B & \underline{88.9} & \underline{85.0} & \underline{72.1} & 82.8 & 80.4 & 70.0 & \underline{88.5} & 85.9 & 72.2 \\
Llama3.1-8B & 87.7 & 82.4 & 68.8 & 81.3 & 78.3 & 68.0 & 87.2 & 84.3 & 69.5 \\
\hline
\end{tabular}
\caption{Correlations (Pearson $r$, Spearman $\rho$, Kendall $\tau$) between Precision / Recall / F1 scores from verification models ($P_m$, $R_m$, $F1_m$) and {4L-RP-Human} Precision / Recall / F1 annotations ($P_h$, $R_h$, $F1_h$).%
}
\label{tab:4lrphuman_correlation_full}
\end{table*}

\begin{table*}[!t]
\centering
\small
\begin{tabular}{|l|ccc|ccc|ccc|}
\hline
& \multicolumn{3}{c|}{$P_m$ vs $CR\times RV$} & \multicolumn{3}{c|}{$R_m$ vs $CV$} & \multicolumn{3}{c|}{$F1_m$ vs $F1_{a}$} \\
Model & \textit{r}$\uparrow$ & $\rho$$\uparrow$ & $\tau$$\uparrow$ &
\textit{r}$\uparrow$ & $\rho$$\uparrow$ & $\tau$$\uparrow$ &
\textit{r}$\uparrow$ & $\rho$$\uparrow$ & $\tau$$\uparrow$ \\
\hline
Data-QuestEval & 88.0 & 66.3 & 51.8 & 82.9 & 72.3 & 56.1 & 88.6 & 74.3 & 58.6 \\
MonoLR & 93.8 & 77.1 & 61.3 & \textbf{94.6} & 90.8 & 78.0 & 95.3 & 85.7 & 71.7 \\
NLI-Based & 91.4 & 66.0 & 52.1 & 89.0 & 85.3 & 69.5 & 91.7 & 80.5 & 64.3 \\
FactSpotter & --- & --- & --- & 91.7 & 89.5 & 74.5 & --- & --- & --- \\
\hline
G3-27B-Prompt
& 81.7 & 60.8 & 46.3 & 72.7 & 56.2 & 42.5 & 78.3 & 58.2 & 44.5 \\
Q3-32B-Prompt
& 86.5 & 59.0 & 44.8 & 86.5 & 66.0 & 50.0 & 87.6 & 62.6 & 47.2 \\
L3.3-70B-Prompt
& 93.3 & 65.8 & 51.0 & 92.6 & 73.5 & 58.0 & 93.8 & 70.4 & 55.0 \\
GPT-5.1-Prompt
& 93.7 & 71.2 & 55.9 & 93.6 & 80.4 & 67.0 & 94.7 & 76.9 & 62.5 \\
\hline
Gemma3-270M%
 & 93.7 & \underline{82.9} & \underline{67.0} & 92.0 & \underline{93.4} & \underline{80.8} & 94.3 & \underline{89.9} & \underline{75.7} \\
Gemma3-1B%
 & \textbf{95.4} & 78.8 & 63.8 & {94.3} & 91.1 & 78.5 & \textbf{95.9} & 86.4 & 72.2 \\
Gemma3-4B%
& 92.1 & 73.3 & 57.5 & 91.5 & 88.7 & 74.3 & 93.1 & 83.1 & 67.7 \\
Gemma3-12B%
& 91.9 & 72.5 & 57.2 & 91.2 & 87.6 & 74.0 & 92.9 & 81.3 & 66.9 \\
\hline
Qwen3-0.6B%
& 93.8 & \underline{80.9} & \underline{65.4} & {91.8} & \underline{91.8} & 78.7 & 94.4 & 87.5 & 73.1 \\
Qwen3-1.7B%
& \underline{95.1} & 80.0 & 65.1 & \underline{94.4} & 91.3 & \underline{79.0} & \underline{95.6} & \underline{87.6} & \underline{74.0} \\
Qwen3-4B%
& 94.0 & 65.9 & 51.2 & 93.3 & 83.8 & 69.5 & 94.9 & 76.2 & 61.6 \\
Qwen3-8B%
& 93.0 & 76.0 & 60.9 & 92.9 & 91.2 & 78.3 & 94.0 & 85.6 & 71.4 \\
Qwen3-14B%
& 94.3 & 71.1 & 56.3 & \textbf{94.6} & 90.0 & 77.7 & \underline{95.5} & 84.3 & 70.3 \\
\hline
Llama3.2-1B%
& \underline{94.7} & \textbf{84.8} & \textbf{69.8} & 93.6 & \textbf{94.2} & \textbf{83.0} & {95.4} & \textbf{90.8} & \textbf{77.7} \\
Llama3.2-3B%
& 91.8 & 69.2 & 53.7 & 90.9 & 85.4 & 70.8 & 93.1 & 78.9 & 63.5 \\
Llama3.1-8B%
& 89.7 & 73.7 & 58.1 & 89.9 & 87.8 & 72.6 & 91.2 & 82.2 & 66.1 \\
\hline
\end{tabular}
\caption{System-level correlation with precision/recall/F1 targets derived from WebNLG 2020 human judgements. We approximate human precision as the product of \textit{Correctness} and \textit{Relevance} judgements ($P_a=CR \times RV$) and treat the \textit{Data Coverage} score $CV$ as human recall $R_a$, and $F1_a$ as the harmonic mean of $P_a$ and $R_a$.}
\label{tab:webnlg20-composed-system_level}
\end{table*}

\begin{table*}[t]
\centering
\small
\begin{tabular}{|l|ccc|ccc|ccc|}
\hline
& \multicolumn{3}{c|}{$P_m$ vs $CR\times RV$} & \multicolumn{3}{c|}{$R_m$ vs $CV$} & \multicolumn{3}{c|}{$F1_m$ vs $F1_a$} \\
Model & \textit{r}$\uparrow$ & $\rho$$\uparrow$ & $\tau$$\uparrow$ &
\textit{r}$\uparrow$ & $\rho$$\uparrow$ & $\tau$$\uparrow$ &
\textit{r}$\uparrow$ & $\rho$$\uparrow$ & $\tau$$\uparrow$ \\
\hline
Data-QuestEval & 67.6 & 56.2 & 44.0 & 62.3 & 56.6 & 42.9 & 69.9 & 62.8 & 48.4 \\
MonoLR & 74.3 & 58.5 & 46.6 & 73.5 & 61.0 & 47.2 & \underline{75.3} & \underline{65.1} & 49.7 \\
NLI-Based & 67.7 & 62.4 & 48.9 & 70.9 & 62.1 & 47.9 & 71.2 & 63.2 & 49.8 \\
FactSpotter & --- & --- & --- & 72.8 & 63.7 & 50.3 & --- & --- & --- \\
\hline
G3-27B-Prompt
& 52.8 & 48.0 & 39.1 & 56.8 & 52.1 & 43.5 & 58.0 & 53.2 & 43.2 \\
Q3-32B-Prompt
& 68.5 & 56.6 & 48.5 & 63.5 & 58.9 & 50.3 & 65.3 & 53.6 & 45.1 \\
L3.3-70B-Prompt
& 61.1 & 57.2 & 48.6 & 58.1 & 54.5 & 46.9 & 56.7 & 54.4 & 45.9 \\
GPT-5.1-Prompt
& 74.1 & 61.9 & 51.8 & 71.9 & 62.3 & 53.1 & 73.5 & 62.4 & 51.4 \\
\hline
Gemma3-270M%
& 73.6 & 60.8 & 51.3 & 72.8 & 62.7 & 52.4 & \underline{74.9} & 64.1 & 53.1 \\
Gemma3-1B%
& 71.7 & 62.6 & 52.6 & 74.5 & 63.7 & 54.2 & 75.0 & 64.0 & 52.7 \\
Gemma3-4B%
& \underline{74.5} & \underline{63.2} & \underline{53.2} & \underline{75.2} & 62.3 & 53.3 & 73.6 & 64.3 & \underline{53.2} \\
Gemma3-12B%
& \textbf{75.4} & 62.8 & 52.6 & \textbf{75.6} & \textbf{65.5} & \textbf{55.8} & 74.5 & \underline{64.9} & \underline{53.3} \\
\hline
Qwen3-0.6B%
& 72.5 & 63.0 & 52.1 & 74.1 & 63.7 & \underline{54.4} & 72.7 & 63.9 & 53.0 \\
Qwen3-1.7B%
& \underline{74.4} & \underline{63.2} & 52.4 & 72.7 & \underline{64.3} & \underline{54.6} & 73.8 & \textbf{65.3} & \textbf{53.5} \\
Qwen3-4B%
& 73.5 & 62.5 & 52.7 & {74.9} & 62.8 & 53.4 & 74.6 & 64.5 & \underline{53.3} \\
Qwen3-8B%
& 72.2 & \underline{63.2} & \underline{52.8} & 74.4 & \underline{64.1} & \underline{54.4} & 72.0 & 63.2 & 51.8 \\
Qwen3-14B%
& 70.1 & 60.5 & 52.6 & 72.7 & 62.7 & 53.7 & 72.4 & 63.7 & 52.4 \\
\hline
Llama3.2-1B%
& 74.9 & 62.4 & 53.0 & \underline{75.0} & 63.5 & 54.1 & 74.8 & 63.9 & 53.1 \\
Llama3.2-3B%
& 74.0 & \textbf{63.7} & \textbf{53.4} & 74.5 & 62.7 & 53.0 & \textbf{76.7} & 63.8 & 52.8 \\
Llama3.1-8B%
& 74.5 & 62.4 & 52.4 & 74.4 & 62.4 & 52.8 & 74.9 & 63.6 & 53.0 \\
\hline
\end{tabular}
\caption{Text-level correlation with precision/recall/F1 targets derived from WebNLG 2020 human judgements.}
\label{tab:webnlg20-composed-text_level2}
\end{table*}

\begin{table*}[t]
\centering
\small
\setlength{\tabcolsep}{4pt}
\begin{tabular}{|l|c|p{11.5cm}|}
\hline
Type & Workers & Description \\
\hline
Keyboard error & 4 & Scores made up of repeated keystrokes, such as ``111'' or ``111111,'' likely from accidental key presses. \\
Out-of-range & 4 & Scores occasionally exceeding 100. \\
Label mismatch & 6 & A clear mismatch between the fine score and the worker's own coarse label, e.g., marking the output as \textit{ok} but giving a score below 20, or the reverse. \\
Constant score & 12 & Always the same score for every annotation (3, 4, 8, 21, or 100 depending on the worker), providing no useful signal. \\
Wrong scale & 1 & Uses a 1--4 scale instead of 0--100. \\
\hline
\end{tabular}
\caption{Categories of annotation quality issues in the E2E dataset, with the number of affected workers per type.}
\label{tab:e2e-workers}
\end{table*}

Because WebNLG covers substantially broader domains and relation types than the restaurant-domain E2E benchmark, we intentionally upweight WebNLG during training to reduce overfitting to E2E.
Concretely, for WebNLG we construct synthetic positive/negative pairs at the \emph{reference-text} level, while for E2E we construct them at the \emph{MR} level by sampling a single text per MR.
This design yields roughly a $4{:}1$ ratio of WebNLG-derived to E2E-derived training instances. 
Table~\ref{tab:error_cooccurrence} further characterises the overlap between different error types by grouping negative instances into those containing exactly one, two, or three error types, indicating that the synthetic corruption process produces diverse and realistically mixed error patterns rather than only single-error cases.

\section{Verifier Finetuning Settings}
\label{sec:lora}

Our models are fine-tuned with LoRA \citep{hu2022lora} applied to all linear
layers of the models, with a dropout of 0.05, a cosine-with-restarts schedule (15 cycles after one-cycle warmup). Training runs for up to 16 epochs with early
stopping after 2 epochs without improvement. For models up to 1.7B parameters,
we use LoRA rank $r=8$ and scaling factor $\alpha=16$; for models between 3B
and 14B parameters, we use $r=16$ and $\alpha=32$. The learning rate is set by
model size: $1\times10^{-4}$ for models below 1B, $5\times10^{-5}$ for 1B--4B
models, and $3\times10^{-5}$ for 8B--14B models. 
The finetuning experiments were run on NVIDIA H100 GPUs in bf16 precision. Our implementation is built on PyTorch 2.10, Transformers 4.57.6, Accelerate 1.13.0, Datasets 3.6.0, vLLM 0.18.1, and ModelScope 1.36.3.

\begin{table}[t!]
\centering
\scriptsize
\begin{tabular}{|l|ccc|ccc|}
\hline
& \multicolumn{3}{c|}{{System Level}} & \multicolumn{3}{c|}{{Text Level}} \\
Model & \textit{r}$\uparrow$ & $\rho$$\uparrow$ & $\tau$$\uparrow$ & \textit{r}$\uparrow$ & $\rho$$\uparrow$ & $\tau$$\uparrow$ \\
\hline
BLEU & 77.0 & 71.6 & 56.8 & 77.6 & 79.6 & 66.9 \\
METEOR & 86.8 & 83.5 & 67.8 & 79.9 & 80.2 & 69.3 \\
PARENT & 88.2 & 82.6 & 66.7 & 79.2 & 79.6 & 66.9 \\
BERTScore & 70.1 & 78.3 & 63.7 & 80.0 & 81.1 & 69.5 \\
BARTScore & 91.0 & 87.5 & 78.2 & 81.4 & 81.8 & 70.9 \\
BLEURT & 90.5 & 88.1 & 72.1 & 82.1 & 82.6 & 71.4 \\
Data-QuestEval-F1 & \underline{94.9} & \underline{93.4} & \underline{85.9} & 78.1 & 79.3 & 67.6 \\
FactSpotter & \textbf{97.2} & \underline{93.9} & \underline{85.1} & \textbf{84.7} & 81.8 & 70.3 \\
NLI-F1 & {94.5} & \textbf{95.0} & \textbf{87.0} & 82.6 & 78.5 & 68.3 \\
MonoLR-F1 & \underline{95.1} & 89.7 & 77.9 & \underline{84.1} & 81.4 & 70.4 \\
\hline
Gemma3-270M-F1 & 94.7 & 88.5 & {79.4} & \underline{83.9} & 82.3 & 75.1 \\
Gemma3-1B-F1 & 93.0 & 86.5 & 76.4 & 83.5 & 82.2 & 74.8 \\
Gemma3-4B-F1 & 93.2 & 87.7 & 78.1 & 82.5 & 82.1 & 74.7 \\
Gemma3-12B-F1 & 91.6 & 87.7 & 78.0 & 83.5 & 82.6 & 75.2 \\
\hline
Qwen3-0.6B-F1 & 92.9 & 87.8 & 78.2 & 83.2 & \underline{82.9} & \textbf{75.8} \\
Qwen3-1.7B-F1 & 92.4 & 86.9 & 77.1 & 83.6 & \textbf{83.1} & \underline{75.5} \\
Qwen3-4B-F1 & 91.3 & 85.2 & 74.2 & 82.2 & 82.6 & 75.3 \\
Qwen3-8B-F1 & 91.1 & 85.2 & 73.4 & 82.4 & 82.5 & 75.3 \\
Qwen3-14B-F1 & -- & -- & 58.3 & 80.9 & 76.7 & 69.2 \\
\hline 
Llama3.2-1B-F1 & 91.8 & 86.9 & 76.3 & 83.6 & 82.4 & 75.3 \\
Llama3.2-3B-F1 & 92.0 & 86.9 & 76.8 & 82.8 & \underline{83.0} & \underline{75.7} \\
Llama3.1-8B-F1 & 92.2 & 86.5 & 76.1 & 82.2 & 81.7 & 74.9 \\
\hline
\end{tabular}
\caption{Correlations between model-predicted F1 and WebNLG 2017 annotation on semantic adequacy. 
}
\label{tab:semantic_adequacy17}
\end{table}

\section{Detailed Correlations with Human Judgements on WebNLG}
\label{sec:webnlg_further}
\label{sec:webnlg2017}
We report detailed Pearson ($r$), Spearman ($\rho$), and Kendall ($\tau$) correlations with the 4L-RP-Human annotations \citep{martinez-etal-2025-semantic} in Table~\ref{tab:4lrphuman_correlation_full}, complementing the results in Table~\ref{tab:4lrphuman_correlation}. Concretely, we compare the precision, recall, and F1 scores produced by the verification models ($P_m$, $R_m$, $F1_m$) against the corresponding human-annotated targets ($P_h$, $R_h$, $F1_h$). Overall, our models achieve consistently stronger agreement with human precision/recall/F1 than the baseline metrics, indicating that our fine-grained unit-level judgments yield reliable scores that align well with human assessments.

We further report system-level and sample-level correlations with Precision/Recall/F1 targets derived from WebNLG 2020 \citep{webnlg2020} human judgements in Tables~\ref{tab:webnlg20-composed-system_level}--\ref{tab:webnlg20-composed-text_level2}. Following \citet{martinez-etal-2025-semantic}, we approximate human precision $P_a$ as the product of \textit{Correctness} and \textit{Relevance} ($P_a = CR \times RV$) and treat \textit{Data Coverage} $CV$ as human recall $R_a$. 
The intuition is that the precision score should penalize both incorrect realizations and irrelevant additions: \textit{Relevance} captures how much of the \emph{output} content stays grounded in the input (discouraging hallucinated extra content), while \textit{Correctness} captures how accurate the grounded content is; their product therefore serves as a proxy for precision.
$F1_a$ is computed as the harmonic mean of $P_a$ and $R_a$.
Across both system-level and sample-level analyses among three correlations, our models achieve consistently stronger correlations with these derived targets than the baseline methods.

We also report correlations on WebNLG 2017 \citep{gardent-etal-2017-webnlg}, which contains 223 outputs from 9 systems with human ratings. We evaluate at both system and text levels in Table~\ref{tab:semantic_adequacy17}, correlating model-predicted $F1_m$ with human scores on \emph{semantic adequacy}. At the text level, our models achieve stronger rank-based agreement with human judgements (Spearman and Kendall) than the baselines, while Pearson correlations remain competitive with the strongest baselines (e.g., FactSpotter and MonoLR-F1). 
At the system level, several reference-less baselines achieve higher correlations, while the correlations are computed over only 9 systems and can be sensitive to small variations in system scores.
Model size shows minimal impact, with small models performing comparably to larger variants. Notably, Qwen3-14B exhibits lower system-level correlations than smaller Qwen3 models, with some metrics failing to reach statistical significance ($p > 0.05$).

\begin{table}[ht]
\centering
\scriptsize
\begin{tabular}{|l|ccc|ccc|}
\hline
& \multicolumn{3}{c|}{{System Level}} & \multicolumn{3}{c|}{{Text Level}} \\
Model & \textit{r}$\uparrow$ & $\rho$$\uparrow$ & $\tau$$\uparrow$ & \textit{r}$\uparrow$ & $\rho$$\uparrow$ & $\tau$$\uparrow$ \\
\hline
BLEU & 58.3 & 48.6 & 34.3 & 24.8 & 24.2 & 19.6 \\
METEOR & 62.9 & 70.1 & 49.5 & 30.6 & 28.6 & 22.9 \\
PARENT-F1 & 10.4 & 6.0 & 2.9 & 14.0 & 13.2 & 11.0 \\
BERTScore-F1 & 70.3 & 66.1 & 54.3 & 30.5 & 28.8 & 23.4 \\
BARTScore & \textbf{90.3} & \underline{88.2} & \underline{72.4} & 38.6 & 34.9 & 28.1 \\
BLEURT & \underline{87.7} & \textbf{90.4} & \textbf{76.2} & 41.1 & 38.1 & 31.4 \\
NLI-P & 35.6 & 27.1 & 23.8 & 11.1 & 8.5 & 7.2 \\
NLI-R & 85.6 & 79.9 & 63.8 & 53.2 & 42.9 & 34.7 \\
NLI-F1 & 82.9 & 86.2 & 69.5 & 49.6 & 44.1 & 35.9 \\
MonoLR-P & 20.3 & 27.0 & 20.0 & 8.9 & 12.7 & 10.3 \\
MonoLR-R & \underline{86.8} & 86.8 & \underline{72.4} & 41.5 & 33.6 & 27.0 \\
MonoLR-F1 & 76.9 & 80.3 & 61.9 & 34.5 & 29.1 & 23.4 \\
FactSpotter & 85.5 & \underline{89.6} & \underline{75.2} & 39.5 & 32.9 & 26.4 \\
Data-QuestEval-P & 36.9 & 18.1 & 15.2 & 5.6 & 4.3 & 3.8 \\
Data-QuestEval-R & 70.5 & 59.2 & 44.8 & 20.3 & 16.5 & 14.0 \\
Data-QuestEval-F1 & 58.1 & 48.4 & 39.1 & 18.4 & 14.6 & 12.3 \\
\hline
Gemma3-270M-P & 24.1 & 48.0 & 32.9 & 25.2 & 22.6 & 20.5 \\
Gemma3-270M-R & 85.9 & 71.7 & 59.1 & 52.9 & 50.9 & 45.8 \\
Gemma3-270M-F1 & 83.0 & 71.7 & 57.1 & 51.5 & 50.8 & 45.6 \\
\hline
Gemma3-1B-P & 12.7 & 36.9 & 26.1 & 22.7 & 20.9 & 18.9 \\
Gemma3-1B-R & 84.7 & 71.2 & 57.1 & \textbf{53.8} & 51.4 & \underline{46.5} \\
Gemma3-1B-F1 & 80.9 & 69.7 & 55.2 & 52.8 & 51.4 & 46.4 \\
\hline
Gemma3-4B-P & 24.1 & 33.7 & 25.4 & 24.1 & 21.5 & 19.5 \\
Gemma3-4B-R & 83.6 & 66.9 & 55.2 & 53.2 & 51.2 & 46.3 \\
Gemma3-4B-F1 & 79.4 & 69.0 & 58.1 & 51.7 & 50.7 & 45.7 \\
\hline
Gemma3-12B-P & 18.3 & 7.1 & 7.5 & 21.7 & 19.1 & 17.3 \\
Gemma3-12B-R & 82.4 & 66.1 & 53.9 & 53.6 & \underline{51.5} & \textbf{46.6} \\
Gemma3-12B-F1 & 76.5 & 63.5 & 50.1 & 52.4 & 51.3 & 46.3 \\
\hline
Qwen3-0.6B-P & 18.5 & 42.5 & 30.0 & 29.3 & 26.8 & 24.4 \\
Qwen3-0.6B-R & 83.8 & 68.1 & 54.3 & 53.4 & 51.2 & 46.2 \\
Qwen3-0.6B-F1 & 79.1 & 69.0 & 53.3 & 52.5 & 51.2 & 46.2 \\
\hline
Qwen3-1.7B-P & 11.0 & 36.4 & 26.1 & 21.5 & 18.7 & 17.1 \\
Qwen3-1.7B-R & 83.3 & 67.7 & 55.2 & 53.5 & 51.1 & 46.2 \\
Qwen3-1.7B-F1 & 78.2 & 67.0 & 52.4 & 52.5 & 51.0 & 46.0 \\
\hline
Qwen3-4B-P & 19.5 & 27.2 & 21.5 & 26.7 & 23.1 & 20.9 \\
Qwen3-4B-R & 83.9 & 67.3 & 54.3 & \underline{53.7} & 51.4 & 46.4 \\
Qwen3-4B-F1 & 78.6 & 69.5 & 56.2 & 52.8 & \underline{51.5} & 46.4 \\
\hline
Qwen3-8B-P & 52.0 & 63.5 & 46.4 & 33.0 & 30.6 & 27.5 \\
Qwen3-8B-R & 83.6 & 69.0 & 56.2 & 53.1 & \underline{51.5} & \underline{46.5} \\
Qwen3-8B-F1 & 79.8 & 70.1 & 57.1 & 51.8 & 51.4 & 46.3 \\
\hline
Qwen3-14B-P & 13.2 & 5.6 & 4.5 & 20.5 & 18.1 & 16.4 \\
Qwen3-14B-R & 83.3 & 72.0 & 56.2 & 53.2 & 51.3 & 46.4 \\
Qwen3-14B-F1 & 77.0 & 67.1 & 54.3 & 51.9 & 51.1 & 46.1 \\
\hline
Llama3.2-1B-P & 10.4 & 21.3 & 16.4 & 22.8 & 20.9 & 19.1 \\
Llama3.2-1B-R & 83.0 & 66.5 & 54.9 & 53.1 & 51.1 & 46.1 \\
Llama3.2-1B-F1 & 78.5 & 66.9 & 52.0 & 52.2 & 51.2 & 46.2 \\
\hline
Llama3.2-3B-P & 21.7 & 36.7 & 25.1 & 26.6 & 23.9 & 21.7 \\
Llama3.2-3B-R & 83.9 & 70.4 & 57.1 & 53.2 & 51.3 & 46.3 \\
Llama3.2-3B-F1 & 79.3 & 68.7 & 53.3 & 52.1 & 51.1 & 46.1 \\
\hline
Llama3.1-8B-P & 14.4 & 21.5 & 15.5 & 23.9 & 20.1 & 18.3 \\
Llama3.1-8B-R & 83.2 & 67.3 & 55.2 & \textbf{53.8} & \textbf{51.6} & \textbf{46.6} \\
Llama3.1-8B-F1 & 77.8 & 66.9 & 51.4 & 52.7 & 51.3 & 46.3 \\
\hline\end{tabular}
\caption{Detailed correlations with E2E fine-grained human judgements on quality assessments. 
}
\label{tab:e2e-corr-full}
\end{table}

\section{Further Analysis on E2E Human Annotations}
\label{app:e2e_details}
\label{sec:e2e-annotation-quality}

Upon inspecting the E2E annotations~\citep{DUSEK2020123}, we find five categories of quality issues at the worker level, summarised in Table~\ref{tab:e2e-workers}. %
The outlier annotations from these workers are removed before computing the correlations\footnote{The filtered annotation data will be publicly released.}.
After removing these outlier annotations, 5{,}515 valid ratings remain from the original 6{,}012 (8.3\% removed). 

When comparing baseline metrics (NLI-Based, Data-QuestEval, FactSpotter) against the E2E coarse labels in Table \ref{tab:e2e-coarse} in Section \ref{sec:corr_e2e}, we convert each metric output into two binary indicators, \textit{has-missing} and \textit{has-added}, and define \textit{is-ok} only when neither indicator is triggered. For the NLI-based metric, \textit{has-missing} is triggered if any source triple receives an entailment probability below the binary threshold (0.5) when checking whether the text entails that triple, while \textit{has-added} is triggered when the precision-side NLI score falls below the binary threshold, indicating that the text is not fully supported by the input data. For Data-QuestEval, we use its two directional answerability scores analogously: if questions generated from the data cannot be answered by the text, we predict \textit{has-missing}; if questions generated from
the text cannot be answered from the data, we predict
\textit{has-added}. We then define \textit{is-ok} as the case where
neither of these two error indicators is triggered. 
FactSpotter is recall-only in this setting: a sample is marked as \textit{has-missing} if any source triple receives a score below the binary threshold against the text, while the samples are otherwise treated as \textit{is-ok}. 

The classification performance on \textit{ok} and \textit{missing} is solid in Section~\ref{sec:corr_e2e}, consistent with the fair agreement observed for these two tasks. 
We compute Krippendorff $\alpha$ for each coarse-label task to assess annotation reliability, as shown in Table~\ref{tab:e2e-iaa}. 
The \textit{added} task yields only slight agreement ($\alpha{=}0.104$), reflecting the difficulty of establishing a consistent standard for this category in a crowdsourced setting, which accounts for its considerably weaker classification results.

Table~\ref{tab:e2e-corr-full} reports Pearson, Spearman, and Kendall correlations with E2E  human quality judgments at both system and text levels. While prior baselines remain strong at the system level, our verifiers are consistently stronger at the text level across all three correlation measures. Across model families, recall-based scores also tend to align better with the human ratings. This likely reflects the fact that E2E annotators agree more on missing content than on additions.

\section{KELM Human Annotation Data Selection for Human Judgement}
\label{sec:data_selection}

To evaluate our verification models across different scales, we select samples from KELM to maximise predicate coverage and diversity in precision, recall, and F1 scores, ensuring annotators evaluate both high-quality and low-quality (data, text) pairs. The construction steps are as follows.

\paragraph{Predicate Selection}
We first remove samples appearing in train and dev sets in Section \ref{subsec:synthetic-data} to ensure the fairness of human judgements. To maximize predicate diversity, we calculate predicate frequency within subsets of different graph sizes (1-6) and then for each size we select 100 least frequent predicates and keep one sample for each of them. 
If the initial selection does not reach the target sample size, we randomly sample additional entries. To ensure maximum predicate coverage, we repeated this random complementation process 100 times and retained the iteration with the highest number of unique predicates.

\paragraph{Precision, Recall and F1 Diversity}
Apart from predicate diversity, we also aim to select samples with diverse precision, recall, and F1 scores so that annotators would evaluate model predictions on both high-quality and low-quality (data, text) pairs. 
For the 100 samples selected by the diversity of predicates, we use MonoLR \citep{martinez-etal-2025-semantic} to estimate the precision, recall, and F1 scores for all candidate samples. We then apply a greedy algorithm to select 10 samples for each graph size that maximise the variance in (precision, recall, F1) space. The algorithm starts by selecting the sample with the median F1 score, and then it iteratively selects samples that maximize the Euclidean distance to already-selected samples in (precision, recall, F1) space. This strategy ensures broad coverage of the score distribution rather than clustering around specific values.

To examine model performance across scales, we generate error detection predictions for each of the 60 selected samples (10 per graph size from 1--6 triples) using four Qwen3 models of different scales: Qwen3-0.6B-8864, Qwen3-4B-4432, Qwen3-8B-7479, and Qwen3-14B-6648. This produces 240 error detection outputs (60 samples $\times$ 4 models) for human annotation.

\begin{figure*}[!t]
    \centering
    \includegraphics[width=\textwidth]{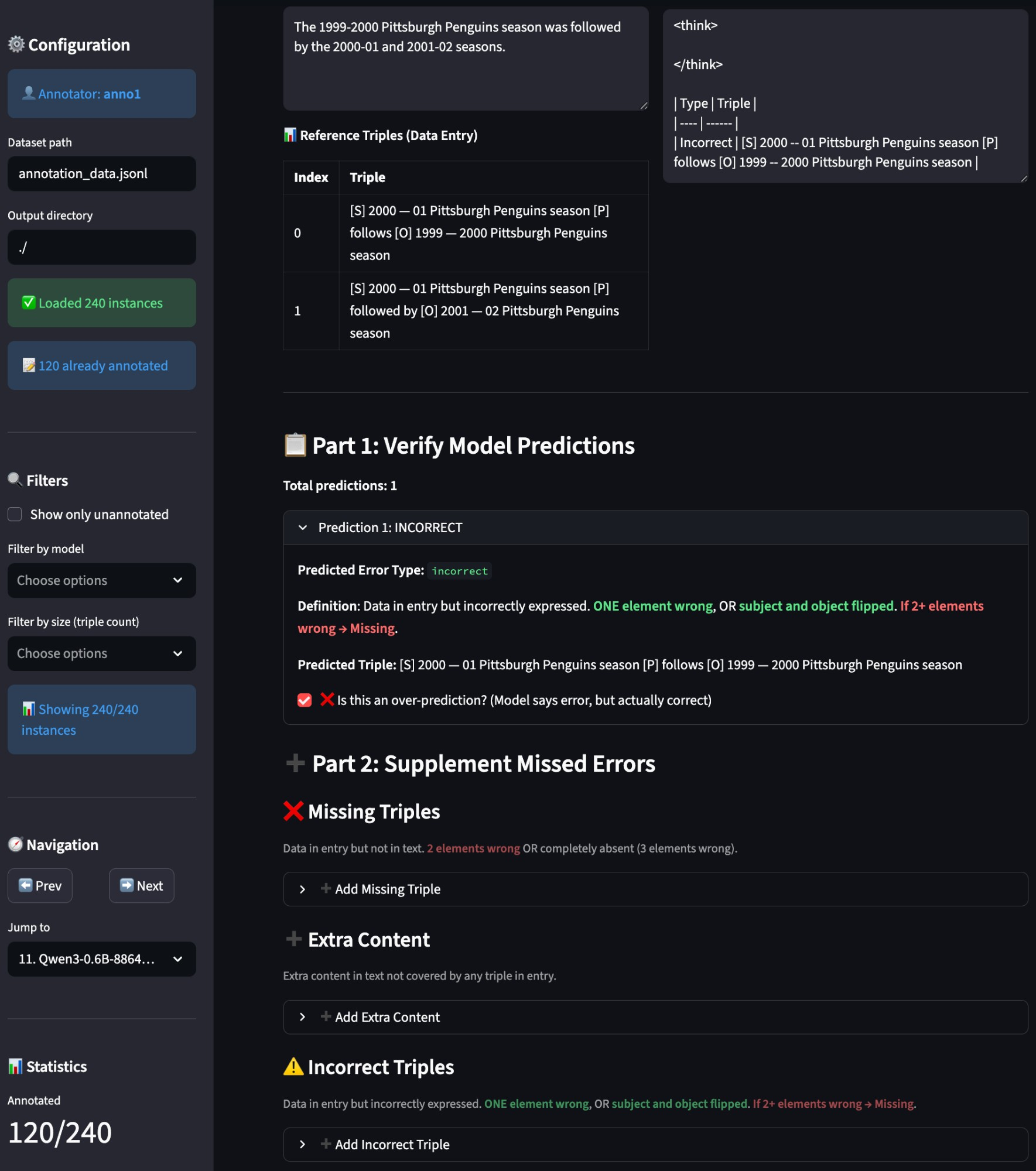}
    \caption{The annotation interface used for human verification on KELM. The left panel shows configuration and navigation controls. The upper area displays the reference data and model response, and the lower area contains Part 1 (verify model predictions) and Part 2 (supplement missed errors) annotation tasks.}
    \label{fig:anno_interface}
\end{figure*}

\section{Human Annotation Protocol}
\label{sec:annotation_protocol}

We compare Qwen3 models of four sizes (0.6B, 4B, 8B, and 14B) across instances grouped by graph size (1--6 triples), sampling 10 instances per group per model, yielding 240 annotated samples (see Appendix~\ref{sec:data_selection} for details). Three annotators each labeled 120 samples: a shared overlap set of 60 annotated by all three, plus 60 additional samples independently assigned to each. 
All annotations were conducted by the authors; no external participants were recruited.
On the overlap set, Krippendorff's $\alpha$ is 0.573 for verifying error predictions and 0.646 for supplementing missed errors. %

\subsection{Part 1: Verify Model Predictions}

For each model prediction containing (triple, error type), annotators answer the following questions:

\paragraph{Question 1: Is this item an over-prediction?}
\begin{itemize}
    \item \textbf{YES}: Model says there is an error, but actually it isn't an error. For example, if the Model predicts \texttt{"MISSING, [S] John [P] age [O] 30"}, while the reference has this triple and the text correctly mentions ``John is 30 years old''. 
    \item \textbf{NO}: Answer question 2. 
\end{itemize}

\paragraph{Question 2: Is the prediction correct?}
\begin{itemize}
    \item \textbf{YES}: The triple and error types of this item are completely correct. 
    \item \textbf{NO}: Prediction is wrong, answer Question 3.
\end{itemize}

\paragraph{Question 3: The reason of wrong judgement?}
\begin{itemize}
    
    \item \textbf{Wrong Error Triple}: The error type output is correct, but the corresponding triple identification is wrong. {Example:}
    \begin{itemize}
        \item Model prediction: \texttt{INCORRECT, [S] John [P] age [O] 35}. 
        \item Actual situation: The data has \texttt{[S] John [P] age [O] 30}, and the text says ``John is 25''.
        \item Analysis: Error type Incorrect is right, but the triple is wrong. The object "35" should be "30". 
    \end{itemize}

    \item \textbf{Wrong Error Type}: The identified error triple is valid, but the corresponding error type is wrong. {For example:}
    \begin{itemize}
        \item Model prediction: \texttt{MISSING, [S] John [P] born in [O] 1990}
        \item Actual situation: The input data has this triple, but the text says ``John was born in~1991.''
        \item Analysis: Only the object "1990" is wrong. Then the error type should not be missing, but incorrect.
    \end{itemize}
\end{itemize}

\subsection{Part 2: Supplement Missed Errors}

After checking error prediction items, the annotators supplement errors that models failed to detect:

\begin{itemize}
    \item \textbf{Add incorrect data}: Copy-paste triples from the input data that are incorrectly expressed in the text.

    \item \textbf{Add omitted data}: Copy-paste triples from the input data that are not mentioned in text.

    \item \textbf{Add extra content}: Copy natural language fragments from the text absent from the data. For example, if the text mentions ``the runway length is 2200 meters'' but the input data does not, the annotator should copy and paste it.
\end{itemize}

\subsection{Annotation Workflow}

The annotation process is divided into two phases to ensure quality and consistency. In the first phase, the goal is to familiarize the annotation process and to achieve sufficient inter-annotator agreement. All three annotators independently annotate the same 60 samples (1/4 of all model outputs). After their annotation, we calculate Krippendorff's $\alpha$ \citep{krippendorff1970} to measure their agreement. Annotators then discuss disagreements to resolve ambiguities and refine their understanding of the annotation guidelines. 
Once adequate agreement is achieved, each annotator labels 60 samples from the remaining samples in the second phase, with no overlap between annotators. This distributes the annotation workload while maintaining the quality standards established in Phase 1.

\section{Further Consistency Analysis with Human Annotations on Noisy KELM}
\label{sec:kelm_appendix}

Table~\ref{tab:kelm_binary} reports model performance on binary consistency judgment, i.e., predicting whether a data--text pair is fully consistent as a whole, measured by F1 and accuracy. %
Performance improves substantially with scale up to 8B, which achieves the best F1 of {75.5\%} and accuracy of {78.3\%}. The 14B model, however, underperforms 4B despite its larger scale, suggesting that scaling alone does not guarantee better calibration on this task.

\begin{table}[pbt]
\centering
\small
\begin{tabular}{lcccc}
\toprule
 & 0.6B & 4B & 8B & 14B \\
\midrule
Bin-F1 (\%)  & 48.5 & 65.2 & \textbf{75.5} & 61.9 \\
Bin-Acc (\%) & 71.7 & 73.3 & \textbf{78.3} & 73.3 \\
\bottomrule
\end{tabular}
\caption{Binary consistency judgment on KELM. \textit{How well does \metricw\ match human judgements of data-text semantic consistency ?}}
\label{tab:kelm_binary}
\end{table}

\begin{table}[pbt]
\centering
\small
\begin{adjustbox}{width=\columnwidth}
\begin{tabular}{lcccc}
\toprule
Model & Over-pred. & Wrong-type & Wrong-triple & Missed \\
\midrule
Q3-0.6B & 75.0 & 9.9 & 0.4 & 14.7 \\
Q3-4B   & 58.5 & 2.4 & 0.5 & 38.5 \\
Q3-8B   & 74.1 & 1.7 & 0.0 & 24.1 \\
Q3-14B  & 56.2 & 6.8 & 0.0 & 37.0 \\
\bottomrule
\end{tabular}
\end{adjustbox}
\caption{Mistake composition of alignment error detection on KELM (all values in \%).}
\label{tab:kelm_mistake}
\end{table}

\begin{table}[pbt]
\centering
\small
\setlength{\tabcolsep}{4pt}
\begin{adjustbox}{width=\columnwidth}
\begin{tabular}{l|ccc |ccc|ccc}
\hline
& \multicolumn{3}{c|}{Missing} & \multicolumn{3}{c|}{Extra} & \multicolumn{3}{c}{Incorrect} \\
Model & P & R & F1 & P & R & F1 & P & R & F1 \\
\hline
Q3-0.6B & 19.5 & \textbf{73.9} & 30.9 & 44.4 & 64.5 & 52.6 & 39.1 & 63.3 & 48.4 \\
Q3-4B   & \textbf{70.0} & 46.7 & 56.0 & 57.6 & 46.3 & 51.4 & 43.9 & 66.4 & 52.9 \\
Q3-8B   & 54.5 & 66.7 & \textbf{60.0} & 58.3 & \textbf{87.5} & \textbf{70.0} & \textbf{48.5} & \textbf{74.4} & \textbf{58.7} \\
Q3-14B  & 63.6 & 35.0 & 45.2 & \textbf{100.0} & 30.8 & 47.1 & 30.0 & 64.3 & 40.9 \\
\hline
\end{tabular}
\end{adjustbox}
\caption{P/R/F1 by error type on KELM: \textit{How well does \metric\ predict the different types of errors?}}
\label{tab:kelm_typewise}
\end{table}

\begin{table}[pbt]
\centering
\small
\begin{tabular}{lcccccc}
\toprule
Model & 1 & 2 & 3 & 4 & 5 & 6 \\
\midrule
Q3-0.6B & 60.0 & 43.3 & 46.2 & 47.8 & 40.1 & 35.3 \\
Q3-4B   & \textbf{100.0} & 58.3 & 34.4 & 52.8 & 40.2 & \textbf{59.3} \\
Q3-8B   & 80.0 & \textbf{88.0} & \textbf{71.7} & \textbf{63.0} & \textbf{52.4} & 56.9 \\
Q3-14B  & 80.0 & 63.3 & 35.0 & 48.7 & 29.7 & 29.2 \\
\bottomrule
\end{tabular}
\caption{F1 by triple count on KELM (in \%).}
\label{tab:kelm_size}
\end{table}

Table~\ref{tab:kelm_mistake} categorises incorrect predictions into 4 types: false alarms (errors reported on correct items), wrong-type (correct detection but wrong error category), wrong-triple (correct category but wrong target triple), and missed errors (unreported errors). False alarms account for 56--75\% of mistakes across models 
pointing to an {error-reporting calibration} bottleneck: the primary challenge lies in deciding {when} to flag an error. %

A key driver of over-prediction is: \metric\ is trained on corpora whose verbalisations tend to closely mirror the surface form of the source triples, so the model may flag semantically faithful but more loosely phrased expressions as errors. For instance, given the predicate ``located in the administrative territorial entity,''  and the text ``Rosewood is located in the Division of Farrer, New South Wales'', %
the model flags this as Incorrect, while annotators judge it a false alarm. Missed errors, by contrast, tend to involve subtle relational distinctions: for instance, the text ``He was also nominated for the Heinrich Heine Prize'' is paired with \texttt{(Max Frisch, award received, Heinrich Heine Prize)}, but the model reports no error, confusing \emph{nominated} with \emph{received}.

Table~\ref{tab:kelm_typewise} reports precision, recall, and F1 by error type. The contrast is most pronounced on Missing errors: 0.6B achieves high recall but very low precision, reflecting indiscriminate over-reporting. The 14B model shows the opposite tendency on Extra errors, with perfect precision but low recall.

\begin{table*}[t]
\centering
\small
\begin{tabular}{llcccccc}
\toprule
& & \multicolumn{3}{c}{System level}
& \multicolumn{3}{c}{Text level} \\
\cmidrule(lr){3-5}\cmidrule(lr){6-8}
Model & Training data
& $r$ & $\rho$ & $\tau$
& $r$ & $\rho$ & $\tau$ \\
\midrule
\multirow{2}{*}{Qwen3-0.6B}
& WebNLG+E2E & 79.1 & 69.0 & 53.3 & 52.5 & 51.2 & 46.2 \\
& WebNLG only & 80.4 & 84.2 & 68.6 & 34.4 & 33.7 & 29.6 \\
\addlinespace[2pt]
\multirow{2}{*}{Qwen3-1.7B}
& WebNLG+E2E & 78.2 & 67.0 & 52.4 & 52.5 & 51.0 & 46.0 \\
& WebNLG only & 84.1 & 83.8 & 67.6 & 36.4 & 34.7 & 30.2 \\
\addlinespace[2pt]
\multirow{2}{*}{Qwen3-4B}
& WebNLG+E2E & 78.6 & 69.5 & 56.2 & 52.8 & 51.5 & 46.4 \\
& WebNLG only & 86.7 & 71.0 & 59.0 & 53.2 & 51.4 & 46.4 \\
\addlinespace[2pt]
\multirow{2}{*}{Qwen3-8B}
& WebNLG+E2E & 79.8 & 70.1 & 57.1 & 51.8 & 51.4 & 46.3 \\
& WebNLG only & 86.3 & 69.9 & 57.1 & 52.8 & 51.2 & 46.1 \\
\addlinespace[2pt]
\multirow{2}{*}{Qwen3-14B}
& WebNLG+E2E & 77.0 & 67.1 & 54.3 & 51.9 & 51.1 & 46.0 \\
& WebNLG only & 85.8 & 78.4 & 58.1 & 46.8 & 45.4 & 40.7 \\
\bottomrule
\end{tabular}
\caption{Pearson $r$, Spearman $\rho$, and Kendall $\tau$ correlations
(\%) between verifier F1 scores and E2E global human quality scores.
System-level results cover 21 systems, and text-level results cover 625
meaning representations.}
\label{tab:e2e-transfer-corr}
\end{table*}

\begin{table}[t]
\centering
\small
\begin{adjustbox}{max width=\columnwidth}
\begin{tabular}{llccc}
\toprule
Model & Training data
& \textit{is-ok}
& \textit{has-missing}
& \textit{has-added} \\
\midrule
\multirow{2}{*}{Qwen3-0.6B}
& WebNLG+E2E & 89.3 & 75.7 & 20.8 \\
& WebNLG only & 66.8 & 54.0 & 20.9 \\
\addlinespace[2pt]
\multirow{2}{*}{Qwen3-1.7B}
& WebNLG+E2E & 89.3 & 75.7 & 23.4 \\
& WebNLG only & 61.4 & 51.6 & 19.8 \\
\addlinespace[2pt]
\multirow{2}{*}{Qwen3-4B}
& WebNLG+E2E & 89.4 & 75.8 & 21.1 \\
& WebNLG only & 89.8 & 76.4 & 23.9 \\
\addlinespace[2pt]
\multirow{2}{*}{Qwen3-8B}
& WebNLG+E2E & 89.4 & 75.7 & 23.3 \\
& WebNLG only & 89.2 & 75.2 & 20.7 \\
\addlinespace[2pt]
\multirow{2}{*}{Qwen3-14B}
& WebNLG+E2E & 89.1 & 75.5 & 18.7 \\
& WebNLG only & 86.2 & 70.1 & 19.8 \\
\bottomrule
\end{tabular}
\end{adjustbox}
\caption{F1 scores (\%) on the E2E coarse-label tasks.
All models are evaluated on the same E2E examples.
``WebNLG+E2E'' denotes joint training on the two synthetic training sets,
whereas ``WebNLG only'' denotes direct transfer without E2E training.}
\label{tab:e2e-transfer-coarse}
\end{table}

\begin{table}[t]
\centering
\small
\begin{tabular}{|l|ccc|}
\hline
& \textit{ok} & \textit{missing} & \textit{added} \\
\hline
Krippendorff $\alpha$ & 0.378 & 0.369 & 0.104 \\
\hline
\end{tabular}
\caption{IAA for the E2E Human annotation labels.}
\label{tab:e2e-iaa}
\end{table}

Table~\ref{tab:kelm_size} reports instance-averaged F1 grouped by triple count. The 8B model performs best for instances with 2--5 triples, and remains close to 4B at size 6. Both 0.6B and 14B degrade more noticeably as input complexity grows, indicating limited robustness to larger graphs.

\section{WebNLG-to-E2E Cross-Dataset Transfer}
\label{app:cross_dataset}

To examine whether the verifiers transfer beyond the dataset used for
training, we take the Qwen3 verifiers trained only on WebNLG 3.0 synthetic
data and evaluate them on E2E without further training.
We compare these models with the verifiers used in our main E2E experiments,
which are trained on the combined WebNLG and E2E synthetic training sets.
Both settings are evaluated on the same E2E examples.

\begin{table*}[t]
\centering
\small
\begin{tabular}{lcccccc}
\toprule
& \multicolumn{2}{c}{Omitted}
& \multicolumn{2}{c}{Incorrect}
& \multicolumn{2}{c}{Extra} \\
\cmidrule(lr){2-3}\cmidrule(lr){4-5}\cmidrule(lr){6-7}
Model
& Strict & Similarity
& Strict & Similarity
& Strict & Similarity \\
\midrule
Qwen3-0.6B & 85.8 & 90.2 & 81.8 & 84.1 & 2.8  & 68.3 \\
Qwen3-1.7B & 90.1 & 90.1 & 72.1 & 74.8 & 9.8  & 75.1 \\
Qwen3-4B   & 94.2 & 94.3 & 91.2 & 91.8 & 18.3 & 80.1 \\
Qwen3-8B   & 94.5 & 94.6 & 90.3 & 91.4 & 27.7 & 80.7 \\
Qwen3-14B  & 94.2 & 94.2 & 90.9 & 91.6 & 44.1 & 84.0 \\
\bottomrule
\end{tabular}
\caption{Strict and similarity-based F1 scores (\%) for fine-grained
cross-dataset error localisation.
The Qwen3 verifiers are trained only on WebNLG 3.0 synthetic data and
evaluated on the E2E synthetic error-detection test set.}
\label{tab:e2e-transfer-localisation}
\end{table*}

\paragraph{Correlation with human judgements.}
Table~\ref{tab:e2e-transfer-corr} reports correlations between verifier F1
scores and E2E global human quality scores.
For the 4B and 8B models,
text-level correlations differ by at most 1.0 point between the two choices
of training data.
Some system-level point estimates are higher with WebNLG-only training,
but all corresponding system-level paired bootstrap intervals include zero.
By contrast, 
the 0.6B and 1.7B WebNLG-only models show a marked reduction at text level. 

\paragraph{Coarse-label classification.}
Table~\ref{tab:e2e-transfer-coarse} compares the two choices of training data
on the three E2E coarse-label tasks.
The 4B and 8B models show the smallest differences:
their \textit{is-ok} and \textit{has-missing} F1 scores differ by no more
than 0.6 points between WebNLG-only and WebNLG+E2E training. 
The smaller models degrade substantially, while the 14B model shows a more
moderate decrease.
Performance on \textit{has-added} remains relatively low in both settings,
consistent with the low inter-annotator agreement for this label
(Table~\ref{tab:e2e-iaa}).

\paragraph{Paired bootstrap analysis.}
We quantify uncertainty in the close 4B and 8B comparisons using 10,000
paired bootstrap resamples with seed 42.
For coarse-label F1 and text-level correlations,
we resample meaning representations as clusters and retain all system outputs
associated with each sampled meaning representation.
For system-level correlations,
we resample systems.
The same sampled units are used for the WebNLG+E2E and WebNLG-only models
in every replicate.
We report
$\Delta=\text{WebNLG only}-\text{WebNLG+E2E}$
with percentile 95\% confidence intervals.

\begin{table}[t]
\centering
\scriptsize
\begin{adjustbox}{max width=\columnwidth}
\begin{tabular}{lcc}
\toprule
Measure & Qwen3-4B & Qwen3-8B \\
\midrule
\textit{is-ok} F1
  & $+0.5\,[+0.2,+0.7]$
  & $-0.2\,[-0.6,+0.2]$ \\
\textit{has-missing} F1
  & $+0.6\,[+0.1,+1.1]$
  & $-0.5\,[-1.5,+0.3]$ \\
\textit{has-added} F1
  & $+2.9\,[-4.2,+9.6]$
  & $-2.6\,[-9.6,+4.3]$ \\
\midrule
Text $r$
  & $+0.3\,[-0.9,+1.5]$
  & $+1.0\,[-0.2,+2.1]$ \\
Text $\rho$
  & $-0.1\,[-1.1,+0.8]$
  & $-0.2\,[-1.0,+0.6]$ \\
Text $\tau$
  & $0.0\,[-0.9,+0.8]$
  & $-0.2\,[-1.0,+0.6]$ \\
\midrule
System $r$
  & $+8.1\,[-0.8,+18.5]$
  & $+6.5\,[-0.3,+15.9]$ \\
System $\rho$
  & $+1.6\,[-9.1,+12.5]$
  & $-0.3\,[-12.4,+10.3]$ \\
System $\tau$
  & $+2.9\,[-7.8,+13.2]$
  & $0.0\,[-11.9,+10.3]$ \\
\bottomrule
\end{tabular}
\end{adjustbox}
\caption{Paired bootstrap differences between WebNLG-only and WebNLG+E2E
training on E2E.
Values are percentage points;
brackets give percentile 95\% confidence intervals and
$\Delta=\text{WebNLG only}-\text{WebNLG+E2E}$.}
\label{tab:e2e-transfer-bootstrap}
\end{table}

As shown in Table~\ref{tab:e2e-transfer-bootstrap}, for Qwen3-4B, the intervals for \textit{is-ok} and \textit{has-missing} exclude zero,
although the differences are small.
All remaining intervals include zero.
For Qwen3-8B,
the two choices of training data are therefore not reliably distinguishable
under this resampling procedure.

\paragraph{Fine-grained error localisation.}
We further evaluate in Table~\ref{tab:e2e-transfer-localisation} whether the WebNLG-trained verifiers can identify
individual errors in the E2E synthetic test set.
Strict F1 requires an exact match with the E2E slot--value representation,
whereas similarity F1 gives partial credit to semantically corresponding
units, as defined in Appendix~\ref{appendix:pr_scores}.

The transferred 4B--14B models obtain strict F1 scores above 94.0 for
Omitted errors and above 90.0 for Incorrect errors. 
Extra errors show a different pattern: strict F1 remains substantially lower,
while similarity F1 reaches 80.1--84.0.
Inspection of the predictions shows that this gap often arises when the
transferred verifier identifies the relevant unsupported content but expresses
it as a WebNLG-style triple rather than an E2E slot--value unit.
Strict matching rejects this representation, whereas similarity matching
recognises the semantic correspondence.

Overall, transfer is most consistent for the 4B and 8B models.
Their coarse-label performance and text-level correlations remain close to
those obtained with joint training, while their fine-grained Omitted and
Incorrect localisation remains strong.

\section{Detailed Precision, Recall, and F1 for Error Detection}
\label{appendix:pr_scores}

We evaluate the ability of the models to detect omitted, extra, and incorrect content by comparing their predictions against the test data labels constructed in Section \ref{subsec:synthetic-data}. Our evaluation employs three complementary levels: triple-level, element-level, and similarity-based metrics to capture different aspects of error prediction quality. 

\paragraph{Triple-level metrics.} Triple-level evaluation measures whether predicted triples match gold-standard triples exactly as complete units. A triple is counted as correct only when the predicted subject, predicate, and object match the gold triple exactly. This strict criterion assesses the model's ability to identify errors with complete accuracy but does not credit partial matches.

\paragraph{Element-level metrics.} Element-level evaluation accounts for partially correct predictions by measuring alignment at the granularity of individual elements. For each matched pair, we count how many of the three elements are identical, and then compute precision, recall, and F1. This metric captures cases where a model partially mis-specifies error elements.

\paragraph{Similarity-based metrics.}
We encode predicted and gold triples using BGE-1.5-Large
\citep{xiao23bge} and compute all pairwise cosine similarities.
We then use the Hungarian algorithm to find the one-to-one matching that
maximises the total cosine similarity between predicted and gold triples
\citep{yang-etal-2023-unicoqe}. The sum of matched similarities is divided by the number of predicted triples
for precision and by the number of gold triples for recall; F1 is their
harmonic mean.
This gives partial credit to semantically close matches while preventing one
gold triple from matching multiple predictions.

We report precision, recall, and F1 separately for each error type (omission, extra, incorrect) and compute the macro-average 
by taking the arithmetic mean of the three per-class scores. %
Since our test set includes positive data--text pairs with no errors, we assign perfect scores (P=R=F1=1) if the model correctly predicts no errors, and we assign P=R=F1=0 if the model incorrectly predicts errors to penalise false alarms. 

Tables~\ref{tab:webnlg-precision}--\ref{tab:e2e-recall} report precision and
recall (over Extra, Omitted, and Incorrect errors) corresponding to the full
F1 results in Tables~\ref{tab:all-models-F1} and~\ref{tab:e2e-all-models-F1}. Across both WebNLG and E2E data, the performance generally increases with model size up to 8B, while further scaling typically yields only marginal improvements. Our fine-tuned verification models consistently outperform prompt-based LLM-as-judge baselines on both precision and recall, even though the baselines are based on much larger LLMs.

\section{Details for Text-to-Data and Data-to-Text Improvement}
\label{sec:appendix_repair_results}

This appendix reports data-to-text and text-to-data improvement prompts in experiments. Figure~\ref{fig:d2t-repair-prompt} shows the data-to-text repair prompt, which revises a previously generated text using source triples and verifier feedback. Figure~\ref{fig:t2d-repair-prompt} shows the text-to-data repair prompt, which revises extracted triples using the source text and verifier feedback. For the revision experiments, we follow \citet{song-etal-2025-multilingual}, using temperature = 0.7 and seed = 42.

We report full model-wise results for text-to-data repair on WebNLG and GenWiki in Tables~\ref{tab:appendix_webnlg_bt5_r123}--\ref{tab:appendix_genwiki_regen_r123}. These tables expand the compact main-text comparison by covering all evaluated repair models, three repair rounds, and three metrics (Exact, Strict, and Partial F1) under both BT5 and ReGen base text-to-data models. Across settings, XQDT remains the strongest method on the stricter Exact and Strict criteria in most cases, while Partial F1 occasionally narrows the gap for NLI and PiVe on GenWiki. These Partial improvements do not generally carry over to Exact or Strict F1, suggesting that they mostly reflect increased local overlap rather than consistently better factual correction in outputs.

We also report full results for data-to-text repair on WebNLG with the Control Prefix baseline in Table~\ref{tab:appendix_webnlg_control_prefix_r13}. This table expands the compact main-text comparison by covering all evaluated repair models, three repair rounds, and the full set of automatic metrics. We report BLEU \citep{papineni-etal-2002-bleu}, ROUGE \citep{lin-2004-rouge}, METEOR \citep{banerjee-lavie-2005-meteor}, chrF++ \citep{popovic-2015-chrf}, PARENT-F1 \citep{dhingra-etal-2019-handling}, BERTScore-F1 \citep{Zhang*2020BERTScore:}, and SBERT \citep{reimers-2019-sentence-bert}. As in the main text, the gains are modest overall because Control Prefix is already a strong baseline on WebNLG, but XQDT still yields measurable improvements on several metrics.

We provide representative data-to-text and text-to-data improvement examples to illustrate how verifier feedback changes model outputs across rounds (Figures \ref{fig:appendix_d2t_case_701},~\ref{fig:appendix_d2t_case_400},~\ref{fig:appendix_t2d_case_id1287}, \ref{fig:appendix_t2d_case_id1296}). We focus on cases where the repairs are visually interpretable and reflect common error types observed in our analysis.

\begin{table*}[t]
\centering
\small
\setlength{\tabcolsep}{3pt}
\begin{tabular}{|l|ccc| ccc| ccc| ccc|}
\hline
& \multicolumn{3}{c|}{Extra} & \multicolumn{3}{c|}{Omitted} & \multicolumn{3}{c|}{Incorrect} & \multicolumn{3}{c|}{Macro} \\
Model 
& $F1_{\text{tri}}$ & $F1_{\text{ele}}$ & $F1_{\text{sim}}$
& $F1_{\text{tri}}$ & $F1_{\text{ele}}$ & $F1_{\text{sim}}$
& $F1_{\text{tri}}$ & $F1_{\text{ele}}$ & $F1_{\text{sim}}$
& $F1_{\text{tri}}$ & $F1_{\text{ele}}$ & $F1_{\text{sim}}$ \\
\hline
Gemma3-27B-Prompt & 4.6 & 18.4 & 39.5 & 14.4 & 18.0 & 27.4 & 43.2 & 45.1 & 49.8 & 20.7 & 27.2 & 38.9 \\
Qwen3-32B-Prompt & 2.8 & 10.9 & 25.3 & 1.2 & 5.7 & 17.4 & 47.6 & 48.8 & 53.4 & 17.2 & 21.8 & 32.0 \\
Llama3.3-70B-Prompt & 13.2 & 36.9 & 61.5 & 51.8 & 56.1 & 61.8 & 52.3 & 56.7 & 61.2 & 39.1 & 49.9 & 61.5 \\
GPT-4.1-Prompt & 16.8 & 43.6 & 70.6 & 30.8 & 34.0 & 41.3 & 64.1 & 64.3 & 65.6 & 37.2 & 47.3 & 59.2 \\
GPT-5.1-Prompt & 18.0 & 45.1 & 72.8 & 81.7 & 83.3 & 84.8 & 78.1 & 78.5 & 79.5 & 59.2 & 69.0 & 79.0 \\
\hline
Gemma3-270M & 57.1 & 69.6 & 81.3 & 91.9 & 92.3 & 93.2 & 87.7 & 88.1 & 89.2 & 78.9 & 83.3 & 87.9 \\
Gemma3-1B & 66.8 & 77.8 & 86.4 & 94.5 & 94.8 & 95.0 & 92.3 & 92.5 & 92.8 & 84.5 & 88.3 & 91.4 \\
Gemma3-4B & 78.3 & 87.1 & 93.0 & \textbf{97.6} & \textbf{97.6} & \textbf{97.7} & 96.5 & 96.5 & 96.5 & 90.8 & 93.7 & 95.7 \\
Gemma3-12B & \underline{81.3} & \underline{88.9} & \underline{94.0} & 96.5 & 96.6 & 96.6 & \underline{97.0} & \underline{97.0} & \underline{97.0} & \underline{91.6} & \underline{94.1} & \underline{95.9} \\
\hline
Qwen3-0.6B & 69.7 & 81.2 & 89.4 & 96.2 & 96.4 & 96.5 & 95.2 & 95.3 & 95.4 & 87.1 & 91.0 & 93.8 \\
Qwen3-1.7B & 73.8 & 83.7 & 90.4 & \underline{96.9} & \underline{97.0} & 97.1 & 96.1 & 96.1 & 96.1 & 88.9 & 92.3 & 94.6 \\
Qwen3-4B & 77.5 & 86.6 & 92.5 & \underline{97.2} & \underline{97.3} & \underline{97.4} & 96.6 & 96.6 & 96.7 & 90.4 & 93.5 & 95.5 \\
Qwen3-8B & 79.2 & 87.5 & 92.9 & 96.8 & 96.8 & 96.9 & \underline{96.9} & \underline{97.0} & \underline{97.0} & \underline{90.9} & 93.8 & 95.6 \\
Qwen3-14B & \underline{81.3} & \underline{89.3} & \underline{94.0} & 96.8 & 96.9 & 97.1 & \textbf{97.1} & \textbf{97.1} & \underline{97.1} & \textbf{91.7} & \textbf{94.4} & \underline{96.1} \\
\hline
Llama3.2-1B & 71.6 & 81.9 & 89.3 & 92.4 & 94.7 & 96.0 & 93.1 & 94.4 & 95.1 & 85.7 & 90.3 & 93.4 \\
Llama3.2-3B & 78.6 & 86.7 & 91.7 & 93.7 & 96.0 & \underline{97.2} & 95.0 & 96.2 & 96.9 & 89.1 & 93.0 & 95.3 \\
Llama3.1-8B & \textbf{82.2} & \textbf{89.8} & \textbf{94.4} & 93.8 & 96.0 & \underline{97.2} & 95.6 & 96.8 & \textbf{97.4} & 90.5 & \underline{94.2} & \textbf{96.3} \\
\hline
\end{tabular}
\caption{F1 scores for prompt-based and our fine-tuned verifier models on error detection data constructed with WebNLG. 
$F1_{\text{tri}}$, $F1_{\text{ele}}$, and $F1_{\text{sim}}$ denote triple-level, element-level, and similarity-based F1, respectively. 
The "Macro" column reports macro-averaged scores across the three error types. %
}
\label{tab:all-models-F1}
\end{table*}

\begin{table*}[!t]
\centering
\footnotesize
\setlength{\tabcolsep}{3pt}
\begin{tabular}{|l|ccc|ccc|ccc|ccc|}
\hline
& \multicolumn{3}{c|}{Extra} & \multicolumn{3}{c|}{Omitted} & \multicolumn{3}{c|}{Incorrect} & \multicolumn{3}{c|}{Macro} \\
Model & $P_{\text{tri}}$ & $P_{\text{ele}}$ & $P_{\text{sim}}$ & $P_{\text{tri}}$ & $P_{\text{ele}}$ & $P_{\text{sim}}$ & $P_{\text{tri}}$ & $P_{\text{ele}}$ & $P_{\text{sim}}$ & $P_{\text{tri}}$ & $P_{\text{ele}}$ & $P_{\text{sim}}$ \\
\hline
Gemma3-27B-Prompt & 3.6 & 14.5 & 31.2 & 9.3 & 11.6 & 17.7 & 35.7 & 37.3 & 41.3 & 16.2 & 21.1 & 30.0 \\
Qwen3-32B-Prompt & 2.8 & 10.9 & 25.4 & 1.0 & 4.5 & 13.9 & 48.5 & 49.8 & 54.4 & 17.4 & 21.7 & 31.2 \\
Llama3.3-70B-Prompt & 12.1 & 33.9 & 56.5 & 40.8 & 44.2 & 48.7 & 42.0 & 45.6 & 49.2 & 31.6 & 41.2 & 51.5 \\
GPT-4.1-Prompt & 14.5 & 37.7 & 61.0 & 34.0 & 37.5 & 45.6 & 52.4 & 52.5 & 53.6 & 33.6 & 42.6 & 53.4 \\
GPT-5.1-Prompt & 15.7 & 39.3 & 63.5 & 75.8 & 77.4 & 78.7 & 75.2 & 75.7 & 76.6 & 55.6 & 64.1 & 72.9 \\
\hline
Gemma3-270M & 54.2 & 66.0 & 77.1 & 90.2 & 90.5 & 91.4 & 84.2 & 84.6 & 85.6 & 76.2 & 80.4 & 84.7 \\
Gemma3-1B & 64.5 & 75.1 & 83.5 & 93.4 & 93.7 & 94.0 & 89.4 & 89.7 & 90.0 & 82.4 & 86.1 & 89.1 \\
Gemma3-4B & 78.3 & 87.0 & 93.0 & \textbf{98.2} & \textbf{98.2} & \textbf{98.3} & 94.5 & 94.5 & 94.5 & 90.3 & 93.3 & \underline{95.3} \\
Gemma3-12B & \underline{81.1} & \underline{88.6} & \underline{93.8} & 96.0 & 96.1 & 96.1 & \textbf{97.4} & \textbf{97.4} & \textbf{97.4} & \textbf{91.5} & \underline{94.0} & \underline{95.8} \\
\hline
Qwen3-0.6B & 68.9 & 80.3 & 88.4 & 95.8 & 95.9 & 96.0 & 93.6 & 93.6 & 93.7 & 86.1 & 89.9 & 92.7 \\
Qwen3-1.7B & 72.4 & 82.2 & 88.8 & \underline{96.5} & 96.6 & 96.7 & 94.3 & 94.3 & 94.3 & 87.7 & 91.0 & 93.3 \\
Qwen3-4B & 77.1 & 86.2 & 92.0 & \underline{97.6} & \underline{97.7} & \underline{97.7} & 94.9 & 94.9 & 94.9 & 89.8 & 92.9 & 94.9 \\
Qwen3-8B & 78.7 & 87.0 & 92.4 & 96.4 & 96.5 & 96.5 & \underline{95.5} & 95.5 & 95.5 & 90.2 & 93.0 & 94.8 \\
Qwen3-14B & \underline{80.7} & \underline{88.5} & \underline{93.2} & 96.2 & 96.3 & 96.5 & \underline{96.3} & \underline{96.3} & \underline{96.3} & \underline{91.0} & \underline{93.7} & \underline{95.3} \\
\hline
Llama3.2-1B & 70.3 & 80.3 & 87.6 & 92.4 & 94.7 & 96.0 & 91.4 & 92.6 & 93.3 & 84.7 & 89.2 & 92.3 \\
Llama3.2-3B & 77.6 & 85.6 & 90.5 & 94.3 & \underline{96.7} & \underline{97.9} & 94.0 & 95.3 & 95.9 & 88.6 & 92.5 & 94.8 \\
Llama3.1-8B & \textbf{82.6} & \textbf{90.3} & \textbf{94.9} & 94.3 & 96.6 & \underline{97.7} & 94.5 & \underline{95.7} & \underline{96.4} & \underline{90.5} & \textbf{94.2} & \textbf{96.3} \\
\hline
\end{tabular}
\caption{Precision (\%) on WebNLG error detection across Extra, Omission, and Incorrect errors.}
\label{tab:webnlg-precision}
\end{table*}

\begin{table*}[!t]
\centering
\footnotesize
\setlength{\tabcolsep}{3pt}
\begin{tabular}{|l|ccc|ccc|ccc|ccc|}
\hline
& \multicolumn{3}{c|}{Extra} & \multicolumn{3}{c|}{Omitted} & \multicolumn{3}{c|}{Incorrect} & \multicolumn{3}{c|}{Macro} \\
Model & $R_{\text{tri}}$ & $R_{\text{ele}}$ & $R_{\text{sim}}$ & $R_{\text{tri}}$ & $R_{\text{ele}}$ & $R_{\text{sim}}$ & $R_{\text{tri}}$ & $R_{\text{ele}}$ & $R_{\text{sim}}$ & $R_{\text{tri}}$ & $R_{\text{ele}}$ & $R_{\text{sim}}$ \\
\hline
Gemma3-27B-Prompt & 6.3 & 25.1 & 54.0 & 32.1 & 40.3 & 61.3 & 54.5 & 56.9 & 62.9 & 31.0 & 40.8 & 59.4 \\
Qwen3-32B-Prompt & 2.8 & 10.9 & 25.3 & 1.6 & 7.6 & 23.3 & 46.7 & 47.9 & 52.4 & 17.0 & 22.2 & 33.7 \\
Llama3.3-70B-Prompt & 14.5 & 40.6 & 67.6 & 70.8 & 76.8 & 84.6 & 69.1 & 74.9 & 80.9 & 51.5 & 64.1 & 77.7 \\
GPT-4.1-Prompt & 20.0 & 51.8 & 83.8 & 28.2 & 31.1 & 37.8 & 82.6 & 82.9 & 84.6 & 43.6 & 55.3 & 68.7 \\
GPT-5.1-Prompt & 21.1 & 52.9 & 85.4 & 88.5 & 90.3 & 91.9 & 81.1 & 81.6 & 82.6 & 63.6 & 74.9 & 86.6 \\
\hline
Gemma3-270M & 60.5 & 73.7 & 86.1 & 93.7 & 94.1 & 95.0 & 91.5 & 91.9 & 93.0 & 81.9 & 86.6 & 91.4 \\
Gemma3-1B & 69.3 & 80.6 & 89.6 & 95.6 & 95.9 & 96.1 & 95.3 & 95.6 & 95.9 & 86.7 & 90.7 & 93.9 \\
Gemma3-4B & 78.4 & 87.2 & 93.1 & 96.9 & 97.0 & 97.0 & \textbf{98.5} & \textbf{98.5} & \textbf{98.5} & 91.3 & 94.2 & 96.2 \\
Gemma3-12B & \underline{81.5} & \underline{89.1} & \underline{94.3} & 97.0 & 97.1 & 97.1 & 96.6 & 96.6 & 96.6 & \underline{91.7} & \underline{94.3} & 96.0 \\
\hline
Qwen3-0.6B & 70.5 & 82.2 & 90.4 & 96.7 & 96.9 & 97.0 & 96.9 & 97.0 & 97.1 & 88.1 & 92.0 & 94.8 \\
Qwen3-1.7B & 75.2 & 85.4 & 92.2 & \textbf{97.3} & \underline{97.3} & \underline{97.5} & 98.0 & 98.0 & 98.0 & 90.1 & 93.6 & 95.9 \\
Qwen3-4B & 78.0 & 87.1 & 93.1 & 96.9 & 97.0 & 97.0 & \underline{98.4} & \underline{98.4} & 98.4 & 91.1 & 94.2 & 96.2 \\
Qwen3-8B & 79.6 & 88.0 & 93.4 & \underline{97.1} & \underline{97.2} & \underline{97.2} & \underline{98.4} & \underline{98.4} & \textbf{98.5} & \underline{91.7} & \underline{94.5} & \underline{96.4} \\
Qwen3-14B & \textbf{82.0} & \textbf{90.0} & \textbf{94.8} & \textbf{97.3} & \textbf{97.5} & \textbf{97.7} & 97.9 & 97.9 & 97.9 & \textbf{92.4} & \textbf{95.1} & \textbf{96.8} \\
\hline
Llama3.2-1B & 73.1 & 83.5 & 91.0 & 92.4 & 94.7 & 96.0 & 94.9 & 96.2 & 96.9 & 86.8 & 91.5 & 94.6 \\
Llama3.2-3B & 79.7 & 87.9 & 93.0 & 93.0 & 95.4 & 96.6 & 95.9 & 97.2 & 97.9 & 89.5 & 93.5 & 95.8 \\
Llama3.1-8B & \underline{81.8} & \underline{89.4} & \underline{93.9} & 93.3 & 95.5 & 96.6 & 96.6 & 97.9 & \textbf{98.5} & 90.6 & \underline{94.3} & \underline{96.3} \\
\hline
\end{tabular}
\caption{Recall (\%) on WebNLG error detection across Extra, Omission, and Incorrect errors.}
\label{tab:webnlg-recall}
\end{table*}

\begin{table*}[!t]
\centering
\small
\setlength{\tabcolsep}{3pt}
\begin{tabular}{|l|ccc|ccc|ccc|ccc|}
\hline
& \multicolumn{3}{c|}{Extra} & \multicolumn{3}{c|}{Omitted} & \multicolumn{3}{c|}{Incorrect} & \multicolumn{3}{c|}{Macro} \\
Model
& $F1_{\text{tri}}$ & $F1_{\text{ele}}$ & $F1_{\text{sim}}$
& $F1_{\text{tri}}$ & $F1_{\text{ele}}$ & $F1_{\text{sim}}$
& $F1_{\text{tri}}$ & $F1_{\text{ele}}$ & $F1_{\text{sim}}$
& $F1_{\text{tri}}$ & $F1_{\text{ele}}$ & $F1_{\text{sim}}$ \\
\hline
Gemma3-27B-Prompt & 4.8 & 19.1 & 37.4 & 16.4 & 20.1 & 32.9 & 36.6 & 38.8 & 45.2 & 19.3 & 26.0 & 38.5 \\
Qwen3-32B-Prompt & 6.6 & 16.4 & 30.2 & 1.7 & 8.9 & 23.5 & 43.3 & 44.9 & 48.9 & 17.2 & 23.4 & 34.2 \\
Llama3.3-70B-Prompt & 19.8 & 39.7 & 57.1 & 57.1 & 59.9 & 66.0 & 54.3 & 56.4 & 59.6 & 43.7 & 52.0 & 60.9 \\
GPT-4.1-Prompt & 33.1 & 53.3 & 69.5 & 41.2 & 44.2 & 52.1 & 62.1 & 62.5 & 64.6 & 45.5 & 53.3 & 62.0 \\
GPT-5.1-Prompt & 35.7 & 57.3 & 72.8 & 82.4 & 82.8 & 83.5 & 78.7 & 78.9 & 80.0 & 65.6 & 73.0 & 78.8 \\
\hline
Gemma3-270M & 92.9 & 94.7 & 95.6 & 94.6 & 94.9 & 95.9 & 95.6 & 95.8 & 96.3 & 94.4 & 95.2 & 95.9 \\
Gemma3-1B & 93.5 & 95.0 & 95.7 & 97.3 & 97.4 & 97.6 & 96.7 & 96.7 & 96.9 & 95.8 & 96.4 & 96.7 \\
Gemma3-4B & 97.1 & 98.1 & 98.5 & 99.0 & 99.0 & 99.0 & \textbf{97.9} & \textbf{97.9} & \textbf{97.9} & 98.0 & 98.3 & 98.4 \\
Gemma3-12B & \underline{97.7} & \underline{98.4} & \underline{98.7} & \textbf{99.3} & \textbf{99.3} & \textbf{99.3} & 97.8 & \textbf{97.9} & \textbf{97.9} & \textbf{98.3} & \textbf{98.5} & \textbf{98.6} \\
\hline
Qwen3-0.6B & 96.3 & 97.3 & 97.7 & 98.7 & 98.7 & 98.7 & 97.7 & 97.7 & 97.7 & 97.6 & 97.9 & 98.1 \\
Qwen3-1.7B & 96.5 & 97.5 & 97.9 & 99.0 & 99.0 & \underline{99.1} & 97.7 & 97.7 & 97.7 & 97.7 & 98.1 & 98.2 \\
Qwen3-4B & \underline{97.7} & \underline{98.4} & \underline{98.7} & \underline{99.1} & \underline{99.1} & \underline{99.1} & \textbf{97.9} & \textbf{97.9} & \textbf{97.9} & \underline{98.2} & \underline{98.4} & \underline{98.5} \\
Qwen3-8B & 97.5 & 98.2 & 98.5 & 99.0 & 99.0 & 99.0 & 97.7 & 97.7 & 97.7 & 98.1 & 98.3 & 98.4 \\
Qwen3-14B & \textbf{97.9} & \textbf{98.6} & \textbf{98.8} & \underline{99.2} & \underline{99.2} & \underline{99.2} & \textbf{97.9} & \textbf{97.9} & \textbf{97.9} & \textbf{98.3} & \textbf{98.5} & \textbf{98.6} \\
\hline
Llama3.2-1B & 93.9 & 95.4 & 96.0 & 97.0 & 97.6 & 97.9 & 96.3 & 96.7 & 96.9 & 95.7 & 96.6 & 96.9 \\
Llama3.2-3B & 97.1 & 97.9 & 98.2 & 98.1 & 98.7 & 99.0 & 97.1 & 97.5 & 97.6 & 97.4 & 98.0 & 98.3 \\
Llama3.1-8B & \underline{97.8} & \underline{98.5} & \textbf{98.8} & 98.0 & 98.6 & 98.9 & 97.3 & 97.6 & 97.8 & 97.7 & 98.2 & \underline{98.5} \\
\hline
\end{tabular}
\caption{F1 scores of verification models on error detection data constructed with E2E.}
\label{tab:e2e-all-models-F1}
\end{table*}

\begin{table*}[!t]
\centering
\footnotesize
\setlength{\tabcolsep}{3pt}
\begin{tabular}{|l|ccc|ccc|ccc|ccc|}
\hline
& \multicolumn{3}{c|}{Extra} & \multicolumn{3}{c|}{Omitted} & \multicolumn{3}{c|}{Incorrect} & \multicolumn{3}{c|}{Macro} \\
Model & $P_{\text{tri}}$ & $P_{\text{ele}}$ & $P_{\text{sim}}$ & $P_{\text{tri}}$ & $P_{\text{ele}}$ & $P_{\text{sim}}$ & $P_{\text{tri}}$ & $P_{\text{ele}}$ & $P_{\text{sim}}$ & $P_{\text{tri}}$ & $P_{\text{ele}}$ & $P_{\text{sim}}$ \\
\hline
Gemma3-27B-Prompt & 3.8 & 14.9 & 29.1 & 11.7 & 14.3 & 23.4 & 32.2 & 34.1 & 39.7 & 15.9 & 21.1 & 30.7 \\
Qwen3-32B-Prompt & 6.0 & 15.1 & 27.7 & 1.5 & 7.6 & 19.9 & 53.0 & 55.0 & 59.8 & 20.2 & 25.9 & 35.8 \\
Llama3.3-70B-Prompt & 18.5 & 37.1 & 53.4 & 48.0 & 50.3 & 55.5 & 42.9 & 44.5 & 47.1 & 36.4 & 44.0 & 52.0 \\
GPT-4.1-Prompt & 28.2 & 45.4 & 59.2 & 40.3 & 43.3 & 51.0 & 52.6 & 53.0 & 54.7 & 40.4 & 47.2 & 55.0 \\
GPT-5.1-Prompt & 30.3 & 48.5 & 61.7 & 74.9 & 75.3 & 75.9 & 77.0 & 77.2 & 78.3 & 60.7 & 67.0 & 72.0 \\
\hline
Gemma3-270M & 92.7 & 94.5 & 95.4 & 93.1 & 93.4 & 94.3 & 94.1 & 94.4 & 94.9 & 93.3 & 94.1 & 94.9 \\
Gemma3-1B & 94.6 & 96.1 & 96.8 & 97.8 & 97.9 & 98.1 & 95.3 & 95.4 & 95.5 & 95.9 & 96.5 & 96.8 \\
Gemma3-4B & 97.4 & 98.5 & 98.8 & 99.4 & 99.4 & 99.4 & 96.0 & 96.0 & 96.0 & 97.6 & 98.0 & 98.1 \\
Gemma3-12B & 98.0 & \underline{98.7} & \underline{99.0} & \underline{99.5} & \underline{99.5} & 99.5 & \textbf{96.4} & \textbf{96.4} & \textbf{96.4} & \underline{97.9} & \underline{98.2} & \underline{98.3} \\
\hline
Qwen3-0.6B & 97.1 & 98.1 & 98.5 & 99.1 & 99.1 & 99.1 & 95.9 & 95.9 & 95.9 & 97.4 & 97.7 & 97.8 \\
Qwen3-1.7B & 96.4 & 97.4 & 97.8 & \underline{99.5} & \underline{99.6} & \underline{99.6} & 95.9 & 95.9 & 95.9 & 97.3 & 97.6 & 97.8 \\
Qwen3-4B & \underline{98.3} & \textbf{99.0} & \textbf{99.3} & \textbf{99.8} & \textbf{99.8} & \textbf{99.8} & \underline{96.2} & \underline{96.2} & \underline{96.2} & \textbf{98.1} & \textbf{98.3} & \textbf{98.4} \\
Qwen3-8B & \underline{98.1} & \underline{98.7} & \underline{99.0} & 99.4 & 99.4 & 99.4 & 95.9 & 95.9 & 95.9 & 97.8 & 98.0 & 98.1 \\
Qwen3-14B & \textbf{98.4} & \textbf{99.0} & \textbf{99.3} & \underline{99.5} & \underline{99.5} & 99.5 & \underline{96.2} & \underline{96.2} & \underline{96.2} & \underline{98.0} & \textbf{98.3} & \underline{98.3} \\
\hline
Llama3.2-1B & 93.9 & 95.4 & 96.0 & 96.6 & 97.2 & 97.4 & 95.0 & 95.4 & 95.6 & 95.2 & 96.0 & 96.3 \\
Llama3.2-3B & 97.6 & 98.5 & 98.8 & 98.7 & 99.4 & \underline{99.7} & 95.2 & 95.6 & 95.7 & 97.2 & 97.8 & 98.1 \\
Llama3.1-8B & 97.9 & 98.6 & \underline{99.0} & 98.6 & 99.2 & 99.5 & 95.6 & 96.0 & 96.1 & 97.4 & 97.9 & 98.2 \\
\hline
\end{tabular}
\caption{Precision (\%) on E2E error detection across Extra, Omission, and Incorrect errors.}
\label{tab:e2e-precision}
\end{table*}

\begin{table*}[!t]
\centering
\footnotesize
\setlength{\tabcolsep}{3pt}
\begin{tabular}{|l|ccc|ccc|ccc|ccc|}
\hline
& \multicolumn{3}{c|}{Extra} & \multicolumn{3}{c|}{Omitted} & \multicolumn{3}{c|}{Incorrect} & \multicolumn{3}{c|}{Macro} \\
Model & $R_{\text{tri}}$ & $R_{\text{ele}}$ & $R_{\text{sim}}$ & $R_{\text{tri}}$ & $R_{\text{ele}}$ & $R_{\text{sim}}$ & $R_{\text{tri}}$ & $R_{\text{ele}}$ & $R_{\text{sim}}$ & $R_{\text{tri}}$ & $R_{\text{ele}}$ & $R_{\text{sim}}$ \\
\hline
Gemma3-27B-Prompt & 6.8 & 26.7 & 52.4 & 27.6 & 33.9 & 55.4 & 42.5 & 45.0 & 52.5 & 25.6 & 35.2 & 53.4 \\
Qwen3-32B-Prompt & 7.2 & 18.1 & 33.2 & 2.1 & 11.0 & 28.8 & 36.6 & 37.9 & 41.3 & 15.3 & 22.3 & 34.5 \\
Llama3.3-70B-Prompt & 21.2 & 42.6 & 61.3 & 70.6 & 74.0 & 81.6 & 73.8 & 76.7 & 81.1 & 55.2 & 64.4 & 74.7 \\
GPT-4.1-Prompt & 40.0 & 64.5 & 84.1 & 42.1 & 45.1 & 53.2 & 75.8 & 76.4 & 78.9 & 52.7 & 62.0 & 72.0 \\
GPT-5.1-Prompt & 43.6 & 70.0 & 88.9 & 91.5 & 91.9 & 92.7 & 80.4 & 80.7 & 81.9 & 71.9 & 80.9 & 87.8 \\
\hline
Gemma3-270M & 93.1 & 95.0 & 95.9 & 96.2 & 96.4 & 97.4 & 97.1 & 97.3 & 97.9 & 95.5 & 96.2 & 97.0 \\
Gemma3-1B & 92.4 & 93.9 & 94.6 & 96.8 & 96.9 & 97.1 & 98.1 & 98.2 & 98.3 & 95.8 & 96.3 & 96.6 \\
Gemma3-4B & 96.7 & 97.8 & 98.1 & \underline{98.5} & \underline{98.5} & \underline{98.5} & \textbf{99.8} & \textbf{99.8} & \textbf{99.8} & 98.3 & \underline{98.7} & 98.8 \\
Gemma3-12B & \underline{97.4} & \underline{98.1} & \underline{98.4} & \textbf{99.2} & \textbf{99.2} & \textbf{99.2} & 99.4 & 99.4 & 99.5 & \textbf{98.7} & \textbf{98.9} & \textbf{99.0} \\
\hline
Qwen3-0.6B & 95.6 & 96.6 & 96.9 & 98.4 & 98.4 & 98.4 & 99.5 & 99.5 & 99.5 & 97.8 & 98.2 & 98.3 \\
Qwen3-1.7B & 96.6 & 97.6 & 98.0 & 98.4 & \underline{98.5} & \underline{98.5} & \underline{99.6} & \underline{99.6} & \underline{99.6} & 98.2 & 98.6 & 98.7 \\
Qwen3-4B & 97.1 & 97.8 & 98.1 & 98.3 & 98.3 & 98.3 & \underline{99.7} & \underline{99.7} & \underline{99.7} & \underline{98.4} & 98.6 & 98.7 \\
Qwen3-8B & 97.0 & 97.7 & 98.0 & \underline{98.5} & \underline{98.5} & \underline{98.5} & \underline{99.6} & \underline{99.6} & \underline{99.6} & \underline{98.4} & 98.6 & 98.7 \\
Qwen3-14B & \underline{97.4} & \underline{98.1} & \underline{98.3} & \underline{98.8} & \underline{98.8} & \underline{98.8} & \underline{99.6} & \underline{99.6} & \underline{99.6} & \underline{98.6} & \underline{98.8} & \underline{98.9} \\
\hline
Llama3.2-1B & 94.0 & 95.5 & 96.1 & 97.4 & 98.0 & 98.3 & 97.7 & 98.1 & 98.3 & 96.4 & 97.2 & 97.6 \\
Llama3.2-3B & 96.6 & 97.4 & 97.7 & 97.4 & 98.0 & 98.3 & 99.1 & 99.4 & \underline{99.6} & 97.7 & 98.3 & 98.6 \\
Llama3.1-8B & \textbf{97.6} & \textbf{98.3} & \textbf{98.7} & 97.5 & 98.1 & 98.4 & 99.0 & 99.4 & 99.5 & 98.0 & 98.6 & \underline{98.9} \\
\hline
\end{tabular}
\caption{Recall (\%) on E2E error detection across Extra, Omission, and Incorrect errors.}
\label{tab:e2e-recall}
\end{table*}

\begin{figure*}[!t]
\centering
\begin{minipage}{0.9\textwidth}
\begin{lstlisting}[
  basicstyle=\ttfamily\footnotesize,
  columns=fullflexible,
  breaklines=true,
  frame=single
]
Repair the extracted triples using the source text and verifier feedback.

# Source text
{source_text}

# Current triples
| subject | predicate | object |
|---|---|---|
| ... | ... | ... |
| ... | ... | ... |

# Verifier feedback

Unsupported triples to remove:
| subject | predicate | object |
|---|---|---|
| ... | ... | ... |
Note: Review whether each flagged row is truly unsupported by the text. Remove it only when the text does not support that fact, not merely because the wording differs.

Potentially incorrect triples to check:
| subject | predicate | object |
|---|---|---|
| ... | ... | ... |
Note: If the text makes the intended fact clear, revise the row into a supported form rather than simply deleting it. Prefer a local correction that preserves the row's intended fact.

Potentially missing facts to consider:
| subject | predicate | object |
|---|---|---|
| ... | ... | ... |
Note: Add a triple only when the fact is clearly and explicitly stated in the text, and do not duplicate a fact that is already captured by a supported row.

# Task
- Use the source text as the final authority.
- Treat the review notes as suggestions rather than facts.
- When a flagged incorrect row can be repaired into a supported fact, prefer correction over deletion.
- When you revise a row, align the subject and object with the wording used in the source text whenever possible.
- Do not remove a row only because its wording differs from the text if the same supported fact is still present.
- If no unsupported rows are flagged, avoid removing rows unless the text clearly contradicts them.
- Keep unaffected triples unchanged.
- If you are uncertain, keep the current triple instead of guessing.

# Output requirements
- Return only a markdown table with exactly three columns: subject, predicate, object.
- Keep one row per final triple.
- Keep unaffected triples unchanged.
- Prefer supported local correction and source-aligned wording; delete a row only when it is unsupported, contradicted, or clearly replaced by a better supported row.
- Do not explain your reasoning.
- If no triples remain, return only the header row and separator row.
\end{lstlisting}
\captionof{figure}{Prompt for improving text-to-data extraction.}
\label{fig:t2d-repair-prompt}
\end{minipage}
\end{figure*}

\begin{figure*}[!t]
\centering
\begin{minipage}{0.9\textwidth}
\begin{lstlisting}[
  basicstyle=\ttfamily\footnotesize,
  columns=fullflexible,
  breaklines=true,
  frame=single
]
You are revising a previously generated data-to-text output.

# Source triples
[
  {"subject": "...", "predicate": "...", "object": "..."},
  {"subject": "...", "predicate": "...", "object": "..."}
]

# Task
Produce a complete revised text that is a faithful lexicalisation of the source triples.
The source triples are the only facts allowed in the output.
Use the verifier feedback as a signal to inspect the previous text, but decide whether to revise by checking the source triples and the previous text together.
The verifier feedback points to possible issues, not to all source-supported content that should remain in the text.
If the previous text already faithfully expresses a source-supported fact, keep that wording unchanged.
If revision is needed, prefer the smallest revision scope that fully fixes the issue: keep local fixes local, and rewrite only the affected region when the surrounding wording is broadly broken.

# Verifier feedback
Missing facts to express:
[
  {"subject": "...", "predicate": "...", "object": "..."}
]
Guidance:
If a listed missing fact is already faithfully expressed in the previous text, do not add it again.
If it is not expressed, add it in the affected region without expanding unrelated content.
If a local insertion is not enough, revise only the affected region needed to express it naturally.

Unsupported facts to remove:
[
  {"subject": "...", "predicate": "...", "object": "..."}
]
Guidance:
If a listed extra fact is truly unsupported by the source triples, remove it.
If the wording is different but the meaning is source-supported, keep it.
Remove only the unsupported meaning that must change, and keep unaffected surrounding content unchanged unless a broader local rewrite is clearly necessary.
Potentially inconsistent facts to check:
[
  {"subject": "...", "predicate": "...", "object": "..."}
]
Guidance:
If the previous text already expresses the source fact faithfully, do not rewrite it only to mirror the predicate label more literally.
If a correction is needed, make the smallest local change that restores factual correctness and natural wording.
If the wording around the issue is broadly broken, rewrite only the affected region needed to express the fact correctly, without rewriting unrelated content.
Revision requirements:
- Output the complete revised text, not just the edits.
- If the previous text already faithfully expresses the source facts, output it unchanged.
- Keep all correct source-supported information from the previous text.
- Do not rewrite a full sentence or neighbouring sentence unless the affected region itself is broadly broken.
- If the previous text contains a broken wrapper such as ERROR or malformed JSON fragments, first recover the intended natural-language content from it, then revise the facts. Still return the final answer in the required JSON format.
- Use only facts supported by the source triples.
- Do not mention the verifier, feedback, or triples in the final text.
- Output only one valid JSON object with exactly this schema: {"full-text": "..."}
\end{lstlisting}
\captionof{figure}{Prompt for improving data-to-text generation.}
\label{fig:d2t-repair-prompt}
\end{minipage}
\end{figure*}

\begin{figure*}[!t]
\small
\textbf{Source data.}

\softbox{%
\ttfamily\footnotesize
Bootleg Series Volume 1: The Quine Tapes | recorded in | United States\\
Bootleg Series Volume 1: The Quine Tapes | recorded in | St. Louis, Missouri\\
Bootleg Series Volume 1: The Quine Tapes | genre | Rock music\\
Bootleg Series Volume 1: The Quine Tapes | preceded by | Squeeze (The Velvet Underground album)\\
Bootleg Series Volume 1: The Quine Tapes | release date | 2001-10-16\\
Bootleg Series Volume 1: The Quine Tapes | runtime | 230.05
}

\vspace{0.4em}
\textbf{Baseline text output.}

\softbox{%
\footnotesize
\bad{Squeeze The Velvet Underground was preceded by Bootleg Series Volume 1: The Quine Tapes} which was recorded in St. Louis, Missouri, United States. It was released on 16th October 2001 and has a runtime of 230.05.
}

\vspace{0.6em}
\textbf{Round 1}

\noindent
\begin{tabular}{@{}p{0.495\linewidth}@{\hspace{0.01\linewidth}}p{0.495\linewidth}@{}}
\textbf{Feedback} &
\textbf{Output} \\[0.15em]
\softbox{%
\ttfamily\footnotesize
omitted:\\{}
\fix{[S] Bootleg Series Volume 1: The Quine Tapes [P] genre [O] Rock music}
}
&
\softbox{%
\footnotesize
Bootleg Series Volume 1: The Quine Tapes was recorded in St. Louis, Missouri, United States and \fix{is categorized under Rock music}. It was released on 16th October 2001 and has a runtime of 230.05. \bad{Squeeze (The Velvet Underground album) was preceded by Bootleg Series Volume 1: The Quine Tapes.}
}
\end{tabular}

\textbf{Round 2}

\noindent
\begin{tabular}{@{}p{0.495\linewidth}@{\hspace{0.01\linewidth}}p{0.495\linewidth}@{}}
\textbf{Feedback} &
\textbf{Output} \\[0.15em]
\softbox{%
\ttfamily\footnotesize
incorrect:\\{}
\bad{[S] Bootleg Series Volume 1: The Quine Tapes [P] preceded by [O] Squeeze (The Velvet Underground album)}
}
&
\softbox{%
\footnotesize
Bootleg Series Volume 1: The Quine Tapes was recorded in St. Louis, Missouri, United States and is categorized under Rock music. It was released on 16th October 2001 and has a runtime of 230.05. \fix{Bootleg Series Volume 1: The Quine Tapes was preceded by Squeeze (The Velvet Underground album).}
}
\end{tabular}

\noindent\textbf{(No further repair)}

\caption{Qualitative data-to-text improvement example (1). }
\label{fig:appendix_d2t_case_701}
\end{figure*}

\begin{figure*}[!t]
\small

\textbf{Source data.}

\softbox{%
\ttfamily\footnotesize
Bootleg Series Volume 1: The Quine Tapes | record label | Polydor Records\\
Polydor Records | location | London\\
Bootleg Series Volume 1: The Quine Tapes | record label | Universal Music Group\\
Universal Music Group | location | Santa Monica, California
}

\vspace{0.4em}
\textbf{Baseline text output.}

\softbox{%
\footnotesize
The record label of Bootleg Series Volume 1: The Quine Tapes is Polydor Records which is located in London. Universal Music Group is located in Santa Monica, California.
}

\vspace{0.6em}
\textbf{Round 1}

\noindent
\begin{tabular}{@{}p{0.495\linewidth}@{\hspace{0.01\linewidth}}p{0.495\linewidth}@{}}
\textbf{Feedback} &
\textbf{Output} \\[0.15em]
\softbox{%
\ttfamily\footnotesize
incorrect:\\{}
\bad{[S] Bootleg Series Volume 1: The Quine Tapes [P] record label [O] Universal Music Group}
}
&
\softbox{%
\footnotesize
The record label of Bootleg Series Volume 1: The Quine Tapes is Polydor Records, which is located in London. \fix{Additionally, the record label of Bootleg Series Volume 1: The Quine Tapes is Universal Music Group,}
which is located in Santa Monica, California.
}
\end{tabular}
\noindent\textbf{(No further repair)}

\caption{Qualitative data-to-text improvement example (2).}
\label{fig:appendix_d2t_case_400}
\end{figure*}

\begin{figure*}[!t]
\small
\textbf{Source text.}

\begin{quote}
Olga Bondareva, known for the Bondareva-Shapley theorem, died on the 9th of December, 1991, in Saint Petersburg. Her alma mater is Leningrad State University.
\end{quote}

\textbf{Initial data output.} \hfill Exact P/R/F1: 0.75 / 0.75 / 0.75

\softbox{%
\ttfamily\footnotesize
Olga Bondareva | death place | Saint Petersburg\\
Olga Bondareva | death date | 1991-12-09\\
\bad{Olga Bondareva | birth place | Bondareva-Shapley}\\
Olga Bondareva | alma mater | Leningrad State University
}

\vspace{0.6em}
\textbf{Round 1}

\noindent
\begin{tabular}{@{}p{0.495\linewidth}@{\hspace{0.01\linewidth}}p{0.495\linewidth}@{}}
\textbf{Feedback} &
\textbf{Output} \hfill Exact P/R/F1: 1.00 / 0.75 / 0.86 \\[0.15em]
\softbox{%
\ttfamily\footnotesize
incorrect:\\
\bad{Olga Bondareva | birth place | Bondareva-Shapley}
}
&
\softbox{%
\ttfamily\footnotesize
Olga Bondareva | death place | Saint Petersburg\\
Olga Bondareva | death date | 1991-12-09\\
Olga Bondareva | alma mater | Leningrad State University
}
\end{tabular}

\textbf{Round 2}

\noindent
\begin{tabular}{@{}p{0.495\linewidth}@{\hspace{0.01\linewidth}}p{0.495\linewidth}@{}}
\textbf{Feedback} &
\textbf{Output} \hfill Exact P/R/F1: 1.00 / 1.00 / 1.00 \\[0.15em]
\softbox{%
\ttfamily\footnotesize
omitted:\\
\fix{Olga Bondareva | known for | Bondareva--Shapley theorem}
}
&
\softbox{%
\ttfamily\footnotesize
Olga Bondareva | death place | Saint Petersburg\\
Olga Bondareva | death date | 1991-12-09\\
Olga Bondareva | alma mater | Leningrad State University\\
\fix{Olga Bondareva | known for | Bondareva--Shapley theorem}
}
\end{tabular}
\noindent\textbf{(No further repair)}

\vspace{0.3em}
\textbf{Gold triples.}

\softbox{%
\ttfamily\footnotesize
Olga Bondareva | known for | Bondareva--Shapley theorem\\
Olga Bondareva | death date | 1991-12-09\\
Olga Bondareva | death place | Saint Petersburg\\
Olga Bondareva | alma mater | Leningrad State University
}

\captionof{figure}{Qualitative text-to-data improvement example (1). }
\label{fig:appendix_t2d_case_id1287}
\end{figure*}

\begin{figure*}[!t]
\small
\textbf{Source text.}

\begin{quote}
There are 27400 students at the University of Burgundy of which 16,800 are undergraduates.
\end{quote}

\textbf{Initial data output.} \hfill Exact P/R/F1: 0.00 / 0.00 / 0.00

\softbox{%
\ttfamily\footnotesize
\bad{University of Burgundy | number of undergraduate students | 27400}\\
\bad{University of Burgundy | affiliation | University of Burgundy}
}

\vspace{0.6em}
\textbf{Round 1}

\noindent
\begin{tabular}{@{}p{0.495\linewidth}@{\hspace{0.01\linewidth}}p{0.495\linewidth}@{}}
\textbf{Feedback} &
\textbf{Output} \hfill Exact P/R/F1: 0.50 / 0.50 / 0.50 \\[0.15em]
\softbox{%
\ttfamily\footnotesize
incorrect:\\
\bad{University of Burgundy | number of undergraduate students | 27400}\\
\bad{University of Burgundy | affiliation | University of Burgundy}
}
&
\softbox{%
\ttfamily\footnotesize
\fix{University of Burgundy | number of undergraduate students | 16800}\\
\bad{University of Burgundy | affiliation | University of Burgundy}
}
\end{tabular}

\vspace{0.6em}
\textbf{Round 2}

\noindent
\begin{tabular}{@{}p{0.495\linewidth}@{\hspace{0.01\linewidth}}p{0.495\linewidth}@{}}
\textbf{Feedback} &
\textbf{Output} \hfill Exact P/R/F1: 0.67 / 1.00 / 0.80 \\[0.15em]
\softbox{%
\ttfamily\footnotesize
incorrect:\\
\bad{University of Burgundy | affiliation | University of Burgundy}\\[0.15em]
missing:\\
\fix{University of Burgundy | number of students | 27400}
}
&
\softbox{%
\ttfamily\footnotesize
University of Burgundy | number of undergraduate students | 16800\\
\bad{University of Burgundy | affiliation | University of Burgundy}\\
\fix{University of Burgundy | number of students | 27400}
}
\end{tabular}

\vspace{0.6em}
\textbf{Round 3}

\noindent
\begin{tabular}{@{}p{0.495\linewidth}@{\hspace{0.01\linewidth}}p{0.495\linewidth}@{}}
\textbf{Feedback} &
\textbf{Output} \hfill Exact P/R/F1: 1.00 / 1.00 / 1.00 \\[0.15em]
\softbox{%
\ttfamily\footnotesize
extra:\\
\bad{University of Burgundy | affiliation | University of Burgundy}
}
&
\softbox{%
\ttfamily\footnotesize
University of Burgundy | number of undergraduate students | 16800\\
\fix{University of Burgundy | number of students | 27400}
}
\end{tabular}

\vspace{0.6em}
\textbf{Gold triples.}

\softbox{%
\ttfamily\footnotesize
University of Burgundy | number of undergraduate students | 16800\\
University of Burgundy | number of students | 27400
}

\captionof{figure}{Qualitative text-to-data improvement example (2). }
\label{fig:appendix_t2d_case_id1296}
\end{figure*}

\begin{table*}[t]
\centering
\footnotesize
\begin{adjustbox}{max width=\textwidth, max totalheight=0.92\textheight}
\begin{tabular}{lllccccccccc}
\toprule
Metric & Method & Round & G3-4B & G3-12B & G3-27B & L3.1-8B & L3.3-70B & Q3-4B & Q3-8B & Q3-14B & Q3-32B \\
\midrule
\multirow{16}{*}{Exact} & Baseline & - & 0.6747 & 0.6747 & 0.6747 & 0.6747 & 0.6747 & 0.6747 & 0.6747 & 0.6747 & 0.6747 \\
\cline{2-12}
& \multirow{3}{*}{XQDT} & Round1 & 0.6851 & 0.7201 & 0.7197 & 0.6960 & 0.7273 & 0.7031 & 0.7164 & 0.7230 & 0.7155 \\
&  & Round2 & 0.7031 & 0.7237 & 0.7234 & 0.7049 & 0.7299 & 0.7119 & 0.7225 & 0.7392 & 0.7251 \\
&  & Round3 & \textbf{0.7077} & \textbf{0.7249} & \textbf{0.7240} & \textbf{0.7065} & \textbf{0.7308} & \textbf{0.7141} & \textbf{0.7234} & \textbf{0.7407} & \textbf{0.7266} \\
\cline{2-12}
& \multirow{3}{*}{FactSpotter} & Round1 & 0.6753 & 0.6546 & 0.6966 & 0.6738 & 0.6903 & 0.6425 & 0.6420 & 0.6483 & 0.6406 \\
&  & Round2 & 0.6726 & 0.6511 & 0.6946 & 0.6700 & 0.6893 & 0.6398 & 0.6384 & 0.6449 & 0.6401 \\
&  & Round3 & 0.6724 & 0.6508 & 0.6949 & 0.6698 & 0.6894 & 0.6396 & 0.6384 & 0.6450 & 0.6401 \\
\cline{2-12}
& \multirow{3}{*}{NLI} & Round1 & 0.6571 & 0.6807 & 0.6898 & 0.6609 & 0.6926 & 0.6453 & 0.6506 & 0.6621 & 0.6491 \\
&  & Round2 & 0.6543 & 0.6854 & 0.6920 & 0.6559 & 0.6917 & 0.6437 & 0.6710 & 0.6679 & 0.6595 \\
&  & Round3 & 0.6539 & 0.6881 & 0.6922 & 0.6577 & 0.6928 & 0.6436 & 0.6720 & 0.6689 & 0.6610 \\
\cline{2-12}
& \multirow{3}{*}{PiVe} & Round1 & 0.6763 & 0.6763 & 0.6747 & 0.6752 & 0.6747 & 0.6763 & 0.6763 & 0.6763 & 0.6747 \\
&  & Round2 & 0.6747 & 0.6747 & 0.6747 & 0.6736 & 0.6747 & 0.6747 & 0.6747 & 0.6747 & 0.6747 \\
&  & Round3 & 0.6747 & 0.6747 & 0.6747 & 0.6734 & 0.6747 & 0.6747 & 0.6747 & 0.6747 & 0.6747 \\
\cline{2-12}
& \multirow{3}{*}{Self-Refine} & Round1 & 0.5666 & 0.5829 & 0.6437 & 0.6201 & 0.6656 & 0.6179 & 0.5942 & 0.6204 & 0.6743 \\
&  & Round2 & 0.5382 & 0.5595 & 0.6262 & 0.5889 & 0.6501 & 0.5920 & 0.5546 & 0.5967 & 0.6733 \\
&  & Round3 & 0.5326 & 0.5500 & 0.6239 & 0.5790 & 0.6468 & 0.5881 & 0.5485 & 0.5866 & 0.6734 \\
\midrule
\multirow{16}{*}{Strict} & Baseline & - & 0.6663 & 0.6663 & 0.6663 & 0.6663 & 0.6663 & 0.6663 & 0.6663 & 0.6663 & 0.6663 \\
\cline{2-12}
& \multirow{3}{*}{XQDT} & Round1 & 0.6674 & 0.7064 & 0.7092 & 0.6852 & 0.7193 & 0.6946 & 0.7080 & 0.7141 & 0.7078 \\
&  & Round2 & 0.6848 & 0.7091 & 0.7125 & 0.6941 & 0.7229 & 0.7038 & 0.7141 & 0.7308 & 0.7184 \\
&  & Round3 & \textbf{0.6880} & \textbf{0.7098} & \textbf{0.7132} & \textbf{0.6954} & \textbf{0.7240} & \textbf{0.7058} & \textbf{0.7150} & \textbf{0.7322} & \textbf{0.7199} \\
\cline{2-12}
& \multirow{3}{*}{FactSpotter} & Round1 & 0.6593 & 0.6461 & 0.6851 & 0.6651 & 0.6836 & 0.6375 & 0.6355 & 0.6427 & 0.6365 \\
&  & Round2 & 0.6574 & 0.6434 & 0.6842 & 0.6617 & 0.6827 & 0.6350 & 0.6332 & 0.6403 & 0.6360 \\
&  & Round3 & 0.6574 & 0.6433 & 0.6843 & 0.6614 & 0.6828 & 0.6350 & 0.6332 & 0.6404 & 0.6360 \\
\cline{2-12}
& \multirow{3}{*}{NLI} & Round1 & 0.6391 & 0.6639 & 0.6769 & 0.6519 & 0.6829 & 0.6392 & 0.6423 & 0.6547 & 0.6433 \\
&  & Round2 & 0.6341 & 0.6671 & 0.6792 & 0.6455 & 0.6815 & 0.6371 & 0.6617 & 0.6596 & 0.6528 \\
&  & Round3 & 0.6335 & 0.6691 & 0.6784 & 0.6462 & 0.6825 & 0.6365 & 0.6621 & 0.6605 & 0.6544 \\
\cline{2-12}
& \multirow{3}{*}{PiVe} & Round1 & 0.6679 & 0.6679 & 0.6663 & 0.6668 & 0.6663 & 0.6679 & 0.6679 & 0.6679 & 0.6663 \\
&  & Round2 & 0.6663 & 0.6663 & 0.6663 & 0.6652 & 0.6663 & 0.6663 & 0.6663 & 0.6663 & 0.6663 \\
&  & Round3 & 0.6663 & 0.6663 & 0.6663 & 0.6650 & 0.6663 & 0.6663 & 0.6663 & 0.6663 & 0.6663 \\
\cline{2-12}
& \multirow{3}{*}{Self-Refine} & Round1 & 0.5435 & 0.5623 & 0.6237 & 0.6092 & 0.6481 & 0.6043 & 0.5787 & 0.6065 & 0.6651 \\
&  & Round2 & 0.5087 & 0.5354 & 0.6010 & 0.5764 & 0.6294 & 0.5751 & 0.5378 & 0.5796 & 0.6622 \\
&  & Round3 & 0.4971 & 0.5247 & 0.5961 & 0.5643 & 0.6245 & 0.5693 & 0.5298 & 0.5672 & 0.6617 \\
\midrule
\multirow{16}{*}{Partial} & Baseline & - & 0.7053 & 0.7053 & 0.7053 & 0.7053 & 0.7053 & 0.7053 & 0.7053 & 0.7053 & 0.7053 \\
\cline{2-12}
& \multirow{3}{*}{XQDT} & Round1 & 0.7174 & 0.7526 & \textbf{0.7592} & 0.7287 & \textbf{0.7666} & 0.7339 & 0.7474 & 0.7534 & 0.7519 \\
&  & Round2 & 0.7370 & 0.7565 & 0.7573 & 0.7392 & 0.7638 & 0.7431 & 0.7540 & 0.7701 & 0.7564 \\
&  & Round3 & \textbf{0.7418} & \textbf{0.7578} & 0.7580 & \textbf{0.7413} & 0.7650 & \textbf{0.7453} & \textbf{0.7548} & \textbf{0.7718} & \textbf{0.7579} \\
\cline{2-12}
& \multirow{3}{*}{FactSpotter} & Round1 & 0.7097 & 0.6826 & 0.7274 & 0.7062 & 0.7195 & 0.6692 & 0.6683 & 0.6751 & 0.6638 \\
&  & Round2 & 0.7021 & 0.6752 & 0.7253 & 0.6980 & 0.7184 & 0.6622 & 0.6612 & 0.6680 & 0.6632 \\
&  & Round3 & 0.7018 & 0.6749 & 0.7256 & 0.6977 & 0.7185 & 0.6620 & 0.6611 & 0.6680 & 0.6632 \\
\cline{2-12}
& \multirow{3}{*}{NLI} & Round1 & 0.6949 & 0.7155 & 0.7217 & 0.6963 & 0.7250 & 0.6760 & 0.6813 & 0.6932 & 0.6749 \\
&  & Round2 & 0.6861 & 0.7166 & 0.7244 & 0.6855 & 0.7243 & 0.6691 & 0.6995 & 0.6970 & 0.6876 \\
&  & Round3 & 0.6859 & 0.7198 & 0.7246 & 0.6875 & 0.7254 & 0.6689 & 0.7006 & 0.6981 & 0.6892 \\
\cline{2-12}
& \multirow{3}{*}{PiVe} & Round1 & 0.7125 & 0.7125 & 0.7053 & 0.7115 & 0.7053 & 0.7125 & 0.7125 & 0.7125 & 0.7053 \\
&  & Round2 & 0.7053 & 0.7053 & 0.7053 & 0.7042 & 0.7053 & 0.7053 & 0.7053 & 0.7053 & 0.7053 \\
&  & Round3 & 0.7053 & 0.7053 & 0.7053 & 0.7039 & 0.7053 & 0.7053 & 0.7053 & 0.7053 & 0.7053 \\
\cline{2-12}
& \multirow{3}{*}{Self-Refine} & Round1 & 0.5973 & 0.6119 & 0.6764 & 0.6526 & 0.6998 & 0.6481 & 0.6254 & 0.6502 & 0.7040 \\
&  & Round2 & 0.5705 & 0.5888 & 0.6606 & 0.6226 & 0.6848 & 0.6230 & 0.5864 & 0.6287 & 0.7043 \\
&  & Round3 & 0.5662 & 0.5804 & 0.6589 & 0.6134 & 0.6825 & 0.6191 & 0.5803 & 0.6202 & 0.7047 \\
\bottomrule
\end{tabular}
\end{adjustbox}
\caption{Extraction repair results on WebNLG with BT5 as the base extractor across repair rounds. Best repair values within each metric block for each model column are boldfaced.}
\label{tab:appendix_webnlg_bt5_r123}
\end{table*}

\begin{table*}[t]
\centering
\footnotesize
\begin{adjustbox}{max width=\textwidth, max totalheight=0.92\textheight}
\begin{tabular}{lllccccccccc}
\toprule
Metric & Method & Round & G3-4B & G3-12B & G3-27B & L3.1-8B & L3.3-70B & Q3-4B & Q3-8B & Q3-14B & Q3-32B \\
\midrule
\multirow{16}{*}{Exact} & Baseline & - & 0.7315 & 0.7315 & 0.7315 & 0.7315 & 0.7315 & 0.7315 & 0.7315 & 0.7315 & 0.7315 \\
\cline{2-12}
& \multirow{3}{*}{XQDT} & Round1 & 0.7409 & 0.7498 & 0.7526 & 0.7421 & \textbf{0.7540} & 0.7434 & 0.7476 & 0.7533 & 0.7525 \\
&  & Round2 & 0.7459 & 0.7524 & 0.7525 & \textbf{0.7450} & 0.7528 & 0.7462 & 0.7500 & 0.7565 & 0.7533 \\
&  & Round3 & \textbf{0.7484} & \textbf{0.7534} & \textbf{0.7530} & 0.7447 & 0.7535 & \textbf{0.7471} & \textbf{0.7506} & \textbf{0.7567} & \textbf{0.7538} \\
\cline{2-12}
& \multirow{3}{*}{FactSpotter} & Round1 & 0.7332 & 0.7306 & 0.7386 & 0.7359 & 0.7414 & 0.7270 & 0.7252 & 0.7323 & 0.7267 \\
&  & Round2 & 0.7309 & 0.7287 & 0.7384 & 0.7327 & 0.7407 & 0.7253 & 0.7232 & 0.7299 & 0.7265 \\
&  & Round3 & 0.7306 & 0.7287 & 0.7383 & 0.7323 & 0.7406 & 0.7253 & 0.7232 & 0.7299 & 0.7265 \\
\cline{2-12}
& \multirow{3}{*}{NLI} & Round1 & 0.7105 & 0.7235 & 0.7246 & 0.7185 & 0.7314 & 0.7136 & 0.7053 & 0.7198 & 0.7172 \\
&  & Round2 & 0.7048 & 0.7279 & 0.7273 & 0.7105 & 0.7289 & 0.7102 & 0.7153 & 0.7200 & 0.7191 \\
&  & Round3 & 0.7048 & 0.7274 & 0.7268 & 0.7106 & 0.7277 & 0.7102 & 0.7149 & 0.7198 & 0.7190 \\
\cline{2-12}
& \multirow{3}{*}{PiVe} & Round1 & 0.7337 & 0.7337 & 0.7315 & 0.7323 & 0.7315 & 0.7337 & 0.7337 & 0.7337 & 0.7315 \\
&  & Round2 & 0.7315 & 0.7315 & 0.7315 & 0.7290 & 0.7315 & 0.7315 & 0.7315 & 0.7315 & 0.7315 \\
&  & Round3 & 0.7315 & 0.7315 & 0.7315 & 0.7290 & 0.7315 & 0.7315 & 0.7315 & 0.7315 & 0.7315 \\
\cline{2-12}
& \multirow{3}{*}{Self-Refine} & Round1 & 0.6005 & 0.6085 & 0.6562 & 0.6600 & 0.6777 & 0.6542 & 0.6178 & 0.6550 & 0.7120 \\
&  & Round2 & 0.5644 & 0.5765 & 0.6413 & 0.6192 & 0.6674 & 0.6270 & 0.5698 & 0.6103 & 0.7062 \\
&  & Round3 & 0.5566 & 0.5648 & 0.6349 & 0.6041 & 0.6622 & 0.6147 & 0.5561 & 0.6012 & 0.7035 \\
\midrule
\multirow{16}{*}{Strict} & Baseline & - & 0.7278 & 0.7278 & 0.7278 & 0.7278 & 0.7278 & 0.7278 & 0.7278 & 0.7278 & 0.7278 \\
\cline{2-12}
& \multirow{3}{*}{XQDT} & Round1 & 0.7346 & 0.7453 & \textbf{0.7486} & 0.7371 & \textbf{0.7501} & 0.7392 & 0.7436 & 0.7499 & 0.7491 \\
&  & Round2 & 0.7389 & 0.7478 & 0.7479 & \textbf{0.7397} & 0.7489 & 0.7419 & 0.7460 & 0.7529 & 0.7497 \\
&  & Round3 & \textbf{0.7409} & \textbf{0.7487} & 0.7486 & 0.7396 & 0.7496 & \textbf{0.7431} & \textbf{0.7465} & \textbf{0.7532} & \textbf{0.7501} \\
\cline{2-12}
& \multirow{3}{*}{FactSpotter} & Round1 & 0.7272 & 0.7267 & 0.7340 & 0.7318 & 0.7375 & 0.7239 & 0.7219 & 0.7289 & 0.7231 \\
&  & Round2 & 0.7249 & 0.7246 & 0.7339 & 0.7283 & 0.7369 & 0.7218 & 0.7197 & 0.7263 & 0.7229 \\
&  & Round3 & 0.7246 & 0.7246 & 0.7338 & 0.7280 & 0.7368 & 0.7218 & 0.7197 & 0.7263 & 0.7229 \\
\cline{2-12}
& \multirow{3}{*}{NLI} & Round1 & 0.7033 & 0.7168 & 0.7193 & 0.7132 & 0.7265 & 0.7103 & 0.7017 & 0.7165 & 0.7139 \\
&  & Round2 & 0.6972 & 0.7183 & 0.7212 & 0.7045 & 0.7234 & 0.7067 & 0.7107 & 0.7164 & 0.7156 \\
&  & Round3 & 0.6970 & 0.7179 & 0.7205 & 0.7044 & 0.7221 & 0.7067 & 0.7105 & 0.7160 & 0.7154 \\
\cline{2-12}
& \multirow{3}{*}{PiVe} & Round1 & 0.7303 & 0.7303 & 0.7278 & 0.7289 & 0.7278 & 0.7303 & 0.7303 & 0.7303 & 0.7278 \\
&  & Round2 & 0.7278 & 0.7278 & 0.7278 & 0.7252 & 0.7278 & 0.7278 & 0.7278 & 0.7278 & 0.7278 \\
&  & Round3 & 0.7278 & 0.7278 & 0.7278 & 0.7252 & 0.7278 & 0.7278 & 0.7278 & 0.7278 & 0.7278 \\
\cline{2-12}
& \multirow{3}{*}{Self-Refine} & Round1 & 0.5865 & 0.5945 & 0.6391 & 0.6511 & 0.6626 & 0.6449 & 0.6071 & 0.6449 & 0.7054 \\
&  & Round2 & 0.5439 & 0.5585 & 0.6182 & 0.6091 & 0.6500 & 0.6161 & 0.5570 & 0.5980 & 0.6984 \\
&  & Round3 & 0.5310 & 0.5450 & 0.6080 & 0.5933 & 0.6444 & 0.6024 & 0.5424 & 0.5882 & 0.6956 \\
\midrule
\multirow{16}{*}{Partial} & Baseline & - & 0.7666 & 0.7666 & 0.7666 & 0.7666 & 0.7666 & 0.7666 & 0.7666 & 0.7666 & 0.7666 \\
\cline{2-12}
& \multirow{3}{*}{XQDT} & Round1 & 0.7720 & 0.7816 & \textbf{0.7878} & 0.7740 & \textbf{0.7895} & 0.7757 & 0.7798 & 0.7848 & \textbf{0.7873} \\
&  & Round2 & 0.7776 & 0.7845 & 0.7845 & \textbf{0.7771} & 0.7850 & 0.7780 & 0.7822 & 0.7880 & 0.7855 \\
&  & Round3 & \textbf{0.7802} & \textbf{0.7854} & 0.7849 & 0.7770 & 0.7857 & \textbf{0.7790} & \textbf{0.7828} & \textbf{0.7882} & 0.7860 \\
\cline{2-12}
& \multirow{3}{*}{FactSpotter} & Round1 & 0.7672 & 0.7629 & 0.7705 & 0.7700 & 0.7727 & 0.7597 & 0.7577 & 0.7650 & 0.7580 \\
&  & Round2 & 0.7621 & 0.7589 & 0.7702 & 0.7636 & 0.7720 & 0.7559 & 0.7537 & 0.7604 & 0.7577 \\
&  & Round3 & 0.7617 & 0.7589 & 0.7701 & 0.7632 & 0.7719 & 0.7559 & 0.7537 & 0.7604 & 0.7577 \\
\cline{2-12}
& \multirow{3}{*}{NLI} & Round1 & 0.7464 & 0.7570 & 0.7570 & 0.7534 & 0.7634 & 0.7468 & 0.7378 & 0.7528 & 0.7487 \\
&  & Round2 & 0.7375 & 0.7600 & 0.7603 & 0.7420 & 0.7617 & 0.7408 & 0.7470 & 0.7513 & 0.7508 \\
&  & Round3 & 0.7376 & 0.7597 & 0.7600 & 0.7424 & 0.7605 & 0.7408 & 0.7465 & 0.7512 & 0.7507 \\
\cline{2-12}
& \multirow{3}{*}{PiVe} & Round1 & 0.7696 & 0.7696 & 0.7666 & 0.7682 & 0.7666 & 0.7696 & 0.7696 & 0.7696 & 0.7666 \\
&  & Round2 & 0.7666 & 0.7666 & 0.7666 & 0.7639 & 0.7666 & 0.7666 & 0.7666 & 0.7666 & 0.7666 \\
&  & Round3 & 0.7666 & 0.7666 & 0.7666 & 0.7639 & 0.7666 & 0.7666 & 0.7666 & 0.7666 & 0.7666 \\
\cline{2-12}
& \multirow{3}{*}{Self-Refine} & Round1 & 0.6302 & 0.6375 & 0.6887 & 0.6921 & 0.7102 & 0.6853 & 0.6493 & 0.6851 & 0.7435 \\
&  & Round2 & 0.5949 & 0.6044 & 0.6736 & 0.6519 & 0.7007 & 0.6581 & 0.6009 & 0.6411 & 0.7379 \\
&  & Round3 & 0.5881 & 0.5925 & 0.6684 & 0.6369 & 0.6961 & 0.6458 & 0.5875 & 0.6331 & 0.7358 \\
\bottomrule
\end{tabular}
\end{adjustbox}
\caption{Extraction repair results on WebNLG with ReGen as the base extractor across repair rounds.}
\label{tab:appendix_webnlg_regen_r123}
\end{table*}

\begin{table*}[t]
\centering
\footnotesize
\begin{adjustbox}{max width=\textwidth, max totalheight=0.92\textheight}
\begin{tabular}{lllccccccccc}
\toprule
Metric & Method & Round & G3-4B & G3-12B & G3-27B & L3.1-8B & L3.3-70B & Q3-4B & Q3-8B & Q3-14B & Q3-32B \\
\midrule
\multirow{16}{*}{Exact} & Baseline & - & 0.4089 & 0.4089 & 0.4089 & 0.4089 & 0.4089 & 0.4089 & 0.4089 & 0.4089 & 0.4089 \\
\cline{2-12}
& \multirow{3}{*}{XQDT} & Round1 & 0.4335 & 0.4860 & 0.4881 & 0.4465 & 0.4619 & 0.4289 & 0.4456 & 0.4326 & 0.4315 \\
&  & Round2 & 0.4709 & 0.5021 & 0.5077 & 0.4748 & 0.4946 & 0.4392 & 0.4592 & 0.4654 & 0.4627 \\
&  & Round3 & \textbf{0.4765} & \textbf{0.5079} & 0.5093 & \textbf{0.4780} & \textbf{0.4960} & \textbf{0.4454} & \textbf{0.4618} & \textbf{0.4708} & \textbf{0.4678} \\
\cline{2-12}
& \multirow{3}{*}{FactSpotter} & Round1 & 0.4085 & 0.3773 & 0.4711 & 0.4013 & 0.3852 & 0.3312 & 0.3344 & 0.3367 & 0.3342 \\
&  & Round2 & 0.4064 & 0.3704 & 0.4616 & 0.3909 & 0.3797 & 0.3264 & 0.3321 & 0.3321 & 0.3315 \\
&  & Round3 & 0.4051 & 0.3696 & 0.4612 & 0.3892 & 0.3797 & 0.3264 & 0.3321 & 0.3310 & 0.3315 \\
\cline{2-12}
& \multirow{3}{*}{NLI} & Round1 & 0.4290 & 0.4704 & 0.5114 & 0.4109 & 0.4543 & 0.3613 & 0.3831 & 0.3833 & 0.3705 \\
&  & Round2 & 0.4330 & 0.4892 & 0.5178 & 0.4142 & 0.4776 & 0.3633 & 0.3943 & 0.3997 & 0.3881 \\
&  & Round3 & 0.4333 & 0.4936 & \textbf{0.5234} & 0.4158 & 0.4814 & 0.3628 & 0.3980 & 0.4014 & 0.3914 \\
\cline{2-12}
& \multirow{3}{*}{PiVe} & Round1 & 0.4073 & 0.4073 & 0.4089 & 0.4066 & 0.4089 & 0.4073 & 0.4073 & 0.4073 & 0.4089 \\
&  & Round2 & 0.4089 & 0.4089 & 0.4089 & 0.4083 & 0.4089 & 0.4089 & 0.4089 & 0.4089 & 0.4089 \\
&  & Round3 & 0.4089 & 0.4089 & 0.4089 & 0.4083 & 0.4089 & 0.4089 & 0.4089 & 0.4089 & 0.4089 \\
\cline{2-12}
& \multirow{3}{*}{Self-Refine} & Round1 & 0.4069 & 0.4213 & 0.4979 & 0.4026 & 0.4751 & 0.3939 & 0.4208 & 0.4149 & 0.4261 \\
&  & Round2 & 0.4084 & 0.4206 & 0.5024 & 0.3960 & 0.4648 & 0.3895 & 0.3943 & 0.4026 & 0.4369 \\
&  & Round3 & 0.4089 & 0.4196 & 0.5040 & 0.4001 & 0.4669 & 0.3794 & 0.3894 & 0.4027 & 0.4391 \\
\midrule
\multirow{16}{*}{Strict} & Baseline & - & 0.3904 & 0.3904 & 0.3904 & 0.3904 & 0.3904 & 0.3904 & 0.3904 & 0.3904 & 0.3904 \\
\cline{2-12}
& \multirow{3}{*}{XQDT} & Round1 & 0.4134 & 0.4693 & 0.4721 & 0.4327 & 0.4474 & 0.4143 & 0.4321 & 0.4187 & 0.4169 \\
&  & Round2 & 0.4509 & 0.4864 & 0.4930 & 0.4593 & 0.4805 & 0.4253 & 0.4457 & 0.4509 & 0.4488 \\
&  & Round3 & \textbf{0.4561} & \textbf{0.4928} & 0.4949 & \textbf{0.4634} & \textbf{0.4821} & \textbf{0.4310} & \textbf{0.4485} & \textbf{0.4565} & \textbf{0.4542} \\
\cline{2-12}
& \multirow{3}{*}{FactSpotter} & Round1 & 0.3904 & 0.3651 & 0.4556 & 0.3875 & 0.3752 & 0.3212 & 0.3254 & 0.3270 & 0.3242 \\
&  & Round2 & 0.3892 & 0.3596 & 0.4469 & 0.3776 & 0.3700 & 0.3172 & 0.3241 & 0.3233 & 0.3216 \\
&  & Round3 & 0.3880 & 0.3588 & 0.4465 & 0.3759 & 0.3699 & 0.3172 & 0.3242 & 0.3223 & 0.3216 \\
\cline{2-12}
& \multirow{3}{*}{NLI} & Round1 & 0.4087 & 0.4520 & 0.4948 & 0.3965 & 0.4397 & 0.3506 & 0.3721 & 0.3730 & 0.3598 \\
&  & Round2 & 0.4140 & 0.4718 & 0.5022 & 0.3999 & 0.4625 & 0.3528 & 0.3834 & 0.3886 & 0.3764 \\
&  & Round3 & 0.4141 & 0.4754 & \textbf{0.5061} & 0.4019 & 0.4671 & 0.3524 & 0.3879 & 0.3905 & 0.3790 \\
\cline{2-12}
& \multirow{3}{*}{PiVe} & Round1 & 0.3889 & 0.3889 & 0.3904 & 0.3882 & 0.3904 & 0.3889 & 0.3889 & 0.3889 & 0.3904 \\
&  & Round2 & 0.3904 & 0.3904 & 0.3904 & 0.3898 & 0.3904 & 0.3904 & 0.3904 & 0.3904 & 0.3904 \\
&  & Round3 & 0.3904 & 0.3904 & 0.3904 & 0.3898 & 0.3904 & 0.3904 & 0.3904 & 0.3904 & 0.3904 \\
\cline{2-12}
& \multirow{3}{*}{Self-Refine} & Round1 & 0.3891 & 0.4065 & 0.4819 & 0.3884 & 0.4636 & 0.3817 & 0.4104 & 0.4048 & 0.4117 \\
&  & Round2 & 0.3926 & 0.4057 & 0.4876 & 0.3829 & 0.4538 & 0.3775 & 0.3841 & 0.3930 & 0.4229 \\
&  & Round3 & 0.3933 & 0.4044 & 0.4906 & 0.3870 & 0.4554 & 0.3675 & 0.3802 & 0.3926 & 0.4244 \\
\midrule
\multirow{16}{*}{Partial} & Baseline & - & 0.4343 & 0.4343 & 0.4343 & 0.4343 & 0.4343 & 0.4343 & 0.4343 & 0.4343 & 0.4343 \\
\cline{2-12}
& \multirow{3}{*}{XQDT} & Round1 & 0.4555 & 0.5074 & 0.5295 & 0.4675 & 0.5004 & 0.4518 & 0.4667 & 0.4521 & 0.4722 \\
&  & Round2 & 0.4935 & 0.5230 & 0.5295 & 0.4974 & 0.5146 & 0.4623 & 0.4798 & 0.4867 & 0.4841 \\
&  & Round3 & \textbf{0.4996} & \textbf{0.5287} & 0.5312 & \textbf{0.5010} & \textbf{0.5166} & \textbf{0.4692} & \textbf{0.4823} & \textbf{0.4929} & \textbf{0.4892} \\
\cline{2-12}
& \multirow{3}{*}{FactSpotter} & Round1 & 0.4449 & 0.4094 & 0.4900 & 0.4385 & 0.4024 & 0.3621 & 0.3655 & 0.3669 & 0.3512 \\
&  & Round2 & 0.4269 & 0.3867 & 0.4805 & 0.4086 & 0.3967 & 0.3422 & 0.3480 & 0.3475 & 0.3484 \\
&  & Round3 & 0.4253 & 0.3859 & 0.4800 & 0.4067 & 0.3966 & 0.3422 & 0.3481 & 0.3465 & 0.3484 \\
\cline{2-12}
& \multirow{3}{*}{NLI} & Round1 & 0.4663 & 0.5070 & 0.5301 & 0.4480 & 0.4720 & 0.3963 & 0.4182 & 0.4176 & 0.3870 \\
&  & Round2 & 0.4535 & 0.5077 & 0.5363 & 0.4319 & 0.4959 & 0.3812 & 0.4128 & 0.4156 & 0.4046 \\
&  & Round3 & 0.4537 & 0.5122 & \textbf{0.5418} & 0.4340 & 0.4999 & 0.3806 & 0.4165 & 0.4176 & 0.4078 \\
\cline{2-12}
& \multirow{3}{*}{PiVe} & Round1 & 0.4554 & 0.4554 & 0.4343 & 0.4547 & 0.4343 & 0.4554 & 0.4554 & 0.4554 & 0.4343 \\
&  & Round2 & 0.4343 & 0.4343 & 0.4343 & 0.4336 & 0.4343 & 0.4343 & 0.4343 & 0.4343 & 0.4343 \\
&  & Round3 & 0.4343 & 0.4343 & 0.4343 & 0.4336 & 0.4343 & 0.4343 & 0.4343 & 0.4343 & 0.4343 \\
\cline{2-12}
& \multirow{3}{*}{Self-Refine} & Round1 & 0.4242 & 0.4379 & 0.5160 & 0.4213 & 0.4920 & 0.4129 & 0.4406 & 0.4307 & 0.4452 \\
&  & Round2 & 0.4250 & 0.4351 & 0.5207 & 0.4142 & 0.4816 & 0.4073 & 0.4110 & 0.4163 & 0.4556 \\
&  & Round3 & 0.4267 & 0.4335 & 0.5211 & 0.4185 & 0.4842 & 0.3968 & 0.4060 & 0.4167 & 0.4574 \\
\bottomrule
\end{tabular}
\end{adjustbox}
\caption{Extraction repair results on GenWiki with BT5 as the base extractor across repair rounds.}
\label{tab:appendix_genwiki_bt5_r123}
\end{table*}

\begin{table*}[t]
\centering
\footnotesize
\begin{adjustbox}{max width=\textwidth, max totalheight=0.92\textheight}
\begin{tabular}{lllccccccccc}
\toprule
Metric & Method & Round & G3-4B & G3-12B & G3-27B & L3.1-8B & L3.3-70B & Q3-4B & Q3-8B & Q3-14B & Q3-32B \\
\midrule
\multirow{16}{*}{Exact} & Baseline & - & 0.4753 & 0.4753 & 0.4753 & 0.4753 & 0.4753 & 0.4753 & 0.4753 & 0.4753 & 0.4753 \\
\cline{2-12}
& \multirow{3}{*}{XQDT} & Round1 & 0.4859 & 0.5067 & 0.5135 & 0.4967 & 0.5047 & 0.4853 & 0.4917 & 0.4845 & 0.4863 \\
&  & Round2 & 0.4960 & 0.5173 & 0.5189 & \textbf{0.5056} & 0.5168 & \textbf{0.4878} & 0.4946 & 0.4928 & 0.4985 \\
&  & Round3 & \textbf{0.4984} & \textbf{0.5195} & 0.5208 & 0.5048 & \textbf{0.5194} & 0.4876 & \textbf{0.4975} & \textbf{0.4948} & \textbf{0.5002} \\
\cline{2-12}
& \multirow{3}{*}{FactSpotter} & Round1 & 0.4674 & 0.4456 & 0.4802 & 0.4649 & 0.4564 & 0.4282 & 0.4293 & 0.4341 & 0.4309 \\
&  & Round2 & 0.4670 & 0.4422 & 0.4785 & 0.4565 & 0.4519 & 0.4279 & 0.4279 & 0.4325 & 0.4282 \\
&  & Round3 & 0.4667 & 0.4424 & 0.4776 & 0.4557 & 0.4516 & 0.4289 & 0.4278 & 0.4318 & 0.4282 \\
\cline{2-12}
& \multirow{3}{*}{NLI} & Round1 & 0.4740 & 0.4935 & 0.5169 & 0.4671 & 0.4906 & 0.4512 & 0.4488 & 0.4533 & 0.4523 \\
&  & Round2 & 0.4775 & 0.5110 & 0.5243 & 0.4628 & 0.5033 & 0.4502 & 0.4534 & 0.4561 & 0.4563 \\
&  & Round3 & 0.4781 & 0.5125 & \textbf{0.5256} & 0.4637 & 0.5093 & 0.4498 & 0.4524 & 0.4569 & 0.4565 \\
\cline{2-12}
& \multirow{3}{*}{PiVe} & Round1 & 0.4734 & 0.4734 & 0.4753 & 0.4725 & 0.4753 & 0.4734 & 0.4734 & 0.4734 & 0.4753 \\
&  & Round2 & 0.4753 & 0.4753 & 0.4753 & 0.4735 & 0.4753 & 0.4753 & 0.4753 & 0.4753 & 0.4753 \\
&  & Round3 & 0.4753 & 0.4753 & 0.4753 & 0.4735 & 0.4753 & 0.4753 & 0.4753 & 0.4753 & 0.4753 \\
\cline{2-12}
& \multirow{3}{*}{Self-Refine} & Round1 & 0.4408 & 0.4408 & 0.5070 & 0.4718 & 0.4894 & 0.4291 & 0.4317 & 0.4423 & 0.4676 \\
&  & Round2 & 0.4363 & 0.4264 & 0.5025 & 0.4616 & 0.4822 & 0.4136 & 0.4122 & 0.4093 & 0.4654 \\
&  & Round3 & 0.4360 & 0.4379 & 0.5113 & 0.4556 & 0.4803 & 0.4127 & 0.4012 & 0.4101 & 0.4646 \\
\midrule
\multirow{16}{*}{Strict} & Baseline & - & 0.4645 & 0.4645 & 0.4645 & 0.4645 & 0.4645 & 0.4645 & 0.4645 & 0.4645 & 0.4645 \\
\cline{2-12}
& \multirow{3}{*}{XQDT} & Round1 & 0.4714 & 0.4932 & 0.4994 & 0.4846 & 0.4921 & 0.4740 & 0.4803 & 0.4737 & 0.4755 \\
&  & Round2 & 0.4809 & 0.5020 & 0.5052 & \textbf{0.4934} & 0.5047 & 0.4768 & 0.4831 & 0.4816 & 0.4864 \\
&  & Round3 & \textbf{0.4836} & \textbf{0.5042} & 0.5072 & 0.4934 & \textbf{0.5076} & \textbf{0.4770} & \textbf{0.4857} & \textbf{0.4837} & \textbf{0.4884} \\
\cline{2-12}
& \multirow{3}{*}{FactSpotter} & Round1 & 0.4534 & 0.4352 & 0.4669 & 0.4535 & 0.4455 & 0.4193 & 0.4201 & 0.4244 & 0.4213 \\
&  & Round2 & 0.4534 & 0.4322 & 0.4652 & 0.4455 & 0.4409 & 0.4185 & 0.4189 & 0.4231 & 0.4187 \\
&  & Round3 & 0.4529 & 0.4324 & 0.4644 & 0.4442 & 0.4407 & 0.4194 & 0.4188 & 0.4223 & 0.4187 \\
\cline{2-12}
& \multirow{3}{*}{NLI} & Round1 & 0.4570 & 0.4789 & 0.5034 & 0.4561 & 0.4793 & 0.4408 & 0.4392 & 0.4432 & 0.4425 \\
&  & Round2 & 0.4597 & 0.4963 & 0.5091 & 0.4516 & 0.4914 & 0.4402 & 0.4436 & 0.4459 & 0.4463 \\
&  & Round3 & 0.4601 & 0.4979 & \textbf{0.5106} & 0.4525 & 0.4971 & 0.4399 & 0.4426 & 0.4467 & 0.4469 \\
\cline{2-12}
& \multirow{3}{*}{PiVe} & Round1 & 0.4630 & 0.4630 & 0.4645 & 0.4621 & 0.4645 & 0.4630 & 0.4630 & 0.4630 & 0.4645 \\
&  & Round2 & 0.4645 & 0.4645 & 0.4645 & 0.4628 & 0.4645 & 0.4645 & 0.4645 & 0.4645 & 0.4645 \\
&  & Round3 & 0.4645 & 0.4645 & 0.4645 & 0.4628 & 0.4645 & 0.4645 & 0.4645 & 0.4645 & 0.4645 \\
\cline{2-12}
& \multirow{3}{*}{Self-Refine} & Round1 & 0.4282 & 0.4302 & 0.4942 & 0.4612 & 0.4784 & 0.4184 & 0.4220 & 0.4319 & 0.4558 \\
&  & Round2 & 0.4229 & 0.4148 & 0.4899 & 0.4509 & 0.4717 & 0.4031 & 0.4022 & 0.3996 & 0.4539 \\
&  & Round3 & 0.4225 & 0.4245 & 0.4986 & 0.4454 & 0.4685 & 0.4037 & 0.3921 & 0.4004 & 0.4532 \\
\midrule
\multirow{16}{*}{Partial} & Baseline & - & 0.5025 & 0.5025 & 0.5025 & 0.5025 & 0.5025 & 0.5025 & 0.5025 & 0.5025 & 0.5025 \\
\cline{2-12}
& \multirow{3}{*}{XQDT} & Round1 & 0.5092 & 0.5316 & \textbf{0.5565} & 0.5209 & \textbf{0.5464} & 0.5107 & 0.5159 & 0.5089 & \textbf{0.5290} \\
&  & Round2 & 0.5205 & 0.5424 & 0.5427 & \textbf{0.5294} & 0.5409 & 0.5129 & 0.5184 & 0.5173 & 0.5227 \\
&  & Round3 & \textbf{0.5235} & \textbf{0.5448} & 0.5443 & 0.5290 & 0.5433 & 0.5130 & 0.5211 & 0.5196 & 0.5243 \\
\cline{2-12}
& \multirow{3}{*}{FactSpotter} & Round1 & 0.5063 & 0.4825 & 0.5031 & 0.5039 & 0.4767 & 0.4653 & 0.4650 & 0.4719 & 0.4515 \\
&  & Round2 & 0.4888 & 0.4619 & 0.5011 & 0.4770 & 0.4721 & 0.4480 & 0.4474 & 0.4531 & 0.4488 \\
&  & Round3 & 0.4884 & 0.4621 & 0.5001 & 0.4761 & 0.4718 & 0.4489 & 0.4473 & 0.4523 & 0.4488 \\
\cline{2-12}
& \multirow{3}{*}{NLI} & Round1 & 0.5140 & 0.5326 & 0.5406 & 0.5069 & 0.5128 & 0.4910 & 0.4869 & 0.4929 & 0.4742 \\
&  & Round2 & 0.5004 & 0.5331 & 0.5478 & 0.4845 & 0.5249 & 0.4722 & 0.4747 & 0.4783 & 0.4772 \\
&  & Round3 & 0.5014 & 0.5347 & 0.5491 & 0.4854 & 0.5312 & 0.4716 & 0.4735 & 0.4790 & 0.4772 \\
\cline{2-12}
& \multirow{3}{*}{PiVe} & Round1 & 0.5214 & 0.5214 & 0.5025 & 0.5205 & 0.5025 & \textbf{0.5214} & \textbf{0.5214} & \textbf{0.5214} & 0.5025 \\
&  & Round2 & 0.5025 & 0.5025 & 0.5025 & 0.5007 & 0.5025 & 0.5025 & 0.5025 & 0.5025 & 0.5025 \\
&  & Round3 & 0.5025 & 0.5025 & 0.5025 & 0.5007 & 0.5025 & 0.5025 & 0.5025 & 0.5025 & 0.5025 \\
\cline{2-12}
& \multirow{3}{*}{Self-Refine} & Round1 & 0.4588 & 0.4577 & 0.5268 & 0.4912 & 0.5078 & 0.4487 & 0.4500 & 0.4604 & 0.4864 \\
&  & Round2 & 0.4536 & 0.4414 & 0.5214 & 0.4798 & 0.4992 & 0.4307 & 0.4272 & 0.4246 & 0.4846 \\
&  & Round3 & 0.4525 & 0.4519 & 0.5299 & 0.4733 & 0.4970 & 0.4299 & 0.4163 & 0.4243 & 0.4830 \\
\bottomrule
\end{tabular}
\end{adjustbox}
\caption{Extraction repair results on GenWiki with ReGen as the base extractor across repair rounds.}
\label{tab:appendix_genwiki_regen_r123}
\end{table*}

\begin{table*}[t]
\centering
\footnotesize
\caption{WebNLG repair results on the Control Prefix baseline across repair rounds.}
\label{tab:appendix_webnlg_control_prefix_r13}
\begin{adjustbox}{max width=\textwidth, max totalheight=.95\textheight}
\begin{tabular}{lllccccccccc}
\toprule
Metric & Method & Round & G3-4B & G3-12B & G3-27B & L3.1-8B & L3.3-70B & Q3-4B & Q3-8B & Q3-14B & Q3-32B \\
\midrule
\multirow{9}{*}{BLEU} & Control Prefix & - & \textbf{53.81} & 53.81 & 53.81 & 53.81 & 53.81 & 53.81 & 53.81 & 53.81 & 53.81 \\
\cmidrule(lr){2-12}
& Self-Refine & R1 & 51.99 & 53.72 & 53.66 & 48.51 & 54.17 & 52.36 & 52.68 & 52.52 & 53.67 \\
& Self-Refine & R3 & 51.54 & 53.65 & 53.73 & 47.26 & \textbf{54.21} & 51.84 & 52.28 & 52.55 & 53.67 \\
\cmidrule(lr){2-12}
& FS & R1 & 53.43 & 53.66 & 53.75 & 53.52 & 53.63 & 53.81 & \textbf{54.05} & 53.78 & 53.96 \\
& FS & R3 & 53.35 & 53.63 & 53.66 & 53.45 & 53.58 & 53.78 & 53.97 & 53.73 & 53.92 \\
\cmidrule(lr){2-12}
& NLI & R1 & 52.67 & 53.27 & 53.54 & 53.05 & 53.21 & 53.56 & 53.64 & 53.62 & 53.78 \\
& NLI & R3 & 52.48 & 53.07 & 53.48 & 53.08 & 53.10 & 53.45 & 53.66 & 53.61 & 53.77 \\
\cmidrule(lr){2-12}
& XQDT & R1 & 53.73 & \textbf{53.87} & \textbf{53.90} & \textbf{53.93} & 53.97 & \textbf{53.83} & 53.86 & 53.90 & 53.98 \\
& XQDT & R3 & 53.69 & 53.86 & 53.90 & 53.88 & 53.98 & 53.80 & 53.85 & \textbf{53.90} & \textbf{53.99} \\
\midrule
\multirow{9}{*}{chrF++} & Control Prefix & - & 69.18 & 69.18 & 69.18 & 69.18 & 69.18 & 69.18 & 69.18 & 69.18 & 69.18 \\
\cmidrule(lr){2-12}
& Self-Refine & R1 & 67.95 & 69.72 & 69.37 & 66.47 & 69.74 & 69.38 & 69.07 & 68.98 & 69.37 \\
& Self-Refine & R3 & 67.62 & \textbf{69.84} & 69.48 & 65.98 & \textbf{69.81} & 69.51 & 69.25 & 69.36 & 69.60 \\
\cmidrule(lr){2-12}
& FS & R1 & 69.05 & 69.48 & 69.58 & 68.88 & 69.71 & 69.56 & 69.74 & 69.54 & 69.80 \\
& FS & R3 & 69.11 & 69.59 & \textbf{69.60} & 69.21 & 69.69 & \textbf{69.65} & \textbf{69.83} & \textbf{69.61} & \textbf{69.86} \\
\cmidrule(lr){2-12}
& NLI & R1 & 68.38 & 69.28 & 69.43 & 68.77 & 69.43 & 69.43 & 69.62 & 69.36 & 69.54 \\
& NLI & R3 & 68.38 & 69.21 & 69.43 & 69.02 & 69.43 & 69.50 & 69.70 & 69.53 & 69.60 \\
\cmidrule(lr){2-12}
& XQDT & R1 & 69.45 & 69.57 & 69.59 & 69.48 & 69.63 & 69.53 & 69.59 & 69.59 & 69.59 \\
& XQDT & R3 & \textbf{69.49} & 69.57 & 69.58 & \textbf{69.55} & 69.64 & 69.53 & 69.58 & 69.59 & 69.59 \\
\midrule
\multirow{9}{*}{ROUGE-L} & Control Prefix & - & 69.15 & 69.15 & 69.15 & 69.15 & 69.15 & \textbf{69.15} & 69.15 & 69.15 & 69.15 \\
\cmidrule(lr){2-12}
& Self-Refine & R1 & 68.30 & 69.02 & 68.93 & 66.21 & 69.29 & 68.41 & 68.46 & 68.47 & 69.12 \\
& Self-Refine & R3 & 68.11 & 69.07 & 68.98 & 65.28 & \textbf{69.33} & 68.10 & 68.39 & 68.55 & 69.14 \\
\cmidrule(lr){2-12}
& FS & R1 & \textbf{69.15} & 69.20 & 69.20 & 68.94 & 69.10 & 69.09 & \textbf{69.35} & \textbf{69.19} & 69.27 \\
& FS & R3 & 69.15 & \textbf{69.22} & 69.15 & 68.86 & 69.05 & 69.09 & 69.32 & 69.18 & 69.23 \\
\cmidrule(lr){2-12}
& NLI & R1 & 68.77 & 69.01 & 69.21 & 68.71 & 68.93 & 69.04 & 69.21 & 69.17 & \textbf{69.27} \\
& NLI & R3 & 68.68 & 68.95 & \textbf{69.23} & 68.78 & 68.83 & 69.00 & 69.28 & 69.17 & 69.26 \\
\cmidrule(lr){2-12}
& XQDT & R1 & 69.09 & 69.09 & 69.14 & \textbf{69.16} & 69.17 & 69.10 & 69.18 & 69.14 & 69.20 \\
& XQDT & R3 & 69.06 & 69.09 & 69.15 & 69.13 & 69.18 & 69.08 & 69.16 & 69.14 & 69.21 \\
\midrule
\multirow{9}{*}{METEOR} & Control Prefix & - & 41.58 & 41.58 & 41.58 & 41.58 & 41.58 & 41.58 & 41.58 & 41.58 & 41.58 \\
\cmidrule(lr){2-12}
& Self-Refine & R1 & 40.98 & 41.96 & 41.74 & 39.86 & 42.16 & 41.84 & 41.65 & 41.54 & 41.81 \\
& Self-Refine & R3 & 40.82 & \textbf{42.08} & 41.82 & 39.72 & \textbf{42.20} & 41.92 & 41.79 & 41.81 & 41.95 \\
\cmidrule(lr){2-12}
& FS & R1 & 41.57 & 41.82 & 41.85 & 41.44 & 41.99 & 41.85 & 42.00 & 41.84 & 41.98 \\
& FS & R3 & 41.63 & 41.89 & 41.87 & 41.65 & 41.98 & \textbf{41.94} & \textbf{42.06} & \textbf{41.89} & \textbf{42.02} \\
\cmidrule(lr){2-12}
& NLI & R1 & 41.20 & 41.72 & 41.84 & 41.38 & 41.82 & 41.76 & 41.92 & 41.74 & 41.83 \\
& NLI & R3 & 41.22 & 41.67 & 41.81 & 41.60 & 41.82 & 41.84 & 41.98 & 41.84 & 41.87 \\
\cmidrule(lr){2-12}
& XQDT & R1 & 41.80 & 41.88 & \textbf{41.88} & 41.82 & 41.91 & 41.85 & 41.89 & 41.87 & 41.86 \\
& XQDT & R3 & \textbf{41.85} & 41.87 & 41.88 & \textbf{41.86} & 41.91 & 41.85 & 41.88 & 41.87 & 41.86 \\
\midrule
\multirow{9}{*}{PARENT-F1} & Control Prefix & - & 65.60 & 65.60 & 65.60 & 65.60 & 65.60 & 65.60 & 65.60 & 65.60 & 65.60 \\
\cmidrule(lr){2-12}
& Self-Refine & R1 & 65.70 & 66.10 & 65.95 & 62.93 & 66.62 & 65.82 & 65.95 & 65.80 & 65.88 \\
& Self-Refine & R3 & 65.65 & \textbf{66.37} & \textbf{66.13} & 62.65 & \textbf{66.69} & 65.67 & \textbf{65.95} & \textbf{65.90} & \textbf{65.89} \\
\cmidrule(lr){2-12}
& FS & R1 & 65.44 & 65.43 & 65.29 & 65.30 & 65.24 & 65.68 & 65.70 & 65.61 & 65.58 \\
& FS & R3 & 65.50 & 65.43 & 65.24 & 65.21 & 65.21 & 65.75 & 65.67 & 65.61 & 65.56 \\
\cmidrule(lr){2-12}
& NLI & R1 & 65.09 & 65.38 & 65.53 & 65.00 & 65.21 & 65.88 & 65.71 & 65.73 & 65.68 \\
& NLI & R3 & 65.18 & 65.31 & 65.58 & 65.17 & 65.18 & \textbf{65.95} & 65.75 & 65.79 & 65.71 \\
\cmidrule(lr){2-12}
& XQDT & R1 & 65.76 & 65.67 & 65.68 & \textbf{65.73} & 65.66 & 65.75 & 65.75 & 65.74 & 65.78 \\
& XQDT & R3 & \textbf{65.81} & 65.66 & 65.68 & 65.71 & 65.70 & 65.75 & 65.74 & 65.74 & 65.79 \\
\midrule
\multirow{9}{*}{BERTScore-F1} & Control Prefix & - & 95.72 & 95.72 & 95.72 & 95.72 & 95.72 & 95.72 & 95.72 & 95.72 & 95.72 \\
\cmidrule(lr){2-12}
& Self-Refine & R1 & 95.55 & 95.77 & 95.78 & 95.19 & 95.82 & 95.70 & 95.70 & 95.69 & 95.78 \\
& Self-Refine & R3 & 95.51 & 95.76 & \textbf{95.78} & 95.09 & \textbf{95.83} & 95.67 & 95.68 & 95.70 & \textbf{95.80} \\
\cmidrule(lr){2-12}
& FS & R1 & 95.73 & 95.77 & 95.76 & 95.71 & 95.77 & 95.74 & 95.80 & 95.77 & 95.78 \\
& FS & R3 & 95.72 & \textbf{95.77} & 95.76 & 95.72 & 95.76 & 95.75 & 95.80 & 95.77 & 95.78 \\
\cmidrule(lr){2-12}
& NLI & R1 & 95.64 & 95.77 & 95.78 & 95.71 & 95.76 & \textbf{95.75} & 95.80 & 95.77 & 95.79 \\
& NLI & R3 & 95.64 & 95.76 & 95.78 & 95.72 & 95.74 & 95.75 & \textbf{95.81} & \textbf{95.78} & 95.79 \\
\cmidrule(lr){2-12}
& XQDT & R1 & \textbf{95.73} & 95.75 & 95.75 & \textbf{95.75} & 95.75 & 95.74 & 95.76 & 95.75 & 95.76 \\
& XQDT & R3 & 95.73 & 95.75 & 95.75 & 95.75 & 95.75 & 95.74 & 95.76 & 95.75 & 95.76 \\
\midrule
\multirow{9}{*}{SBERT} & Control Prefix & - & 95.20 & 95.20 & 95.20 & 95.20 & 95.20 & 95.20 & 95.20 & 95.20 & 95.20 \\
\cmidrule(lr){2-12}
& Self-Refine & R1 & 95.01 & 95.44 & 95.41 & 94.11 & 95.49 & 95.41 & 95.43 & 95.15 & 95.42 \\
& Self-Refine & R3 & 94.99 & \textbf{95.51} & 95.46 & 94.08 & 95.52 & 95.53 & 95.44 & 95.30 & \textbf{95.46} \\
\cmidrule(lr){2-12}
& FS & R1 & 95.28 & 95.39 & 95.40 & 95.23 & 95.44 & 95.38 & 95.43 & 95.33 & 95.36 \\
& FS & R3 & \textbf{95.30} & 95.40 & 95.41 & 95.30 & 95.44 & 95.40 & 95.44 & 95.34 & 95.38 \\
\cmidrule(lr){2-12}
& NLI & R1 & 95.20 & 95.51 & 95.52 & 95.24 & \textbf{95.53} & 95.51 & 95.56 & 95.40 & 95.41 \\
& NLI & R3 & 95.25 & 95.51 & \textbf{95.55} & \textbf{95.36} & 95.53 & \textbf{95.55} & \textbf{95.58} & \textbf{95.45} & 95.44 \\
\cmidrule(lr){2-12}
& XQDT & R1 & 95.28 & 95.35 & 95.35 & 95.31 & 95.34 & 95.32 & 95.35 & 95.34 & 95.34 \\
& XQDT & R3 & 95.30 & 95.35 & 95.36 & 95.34 & 95.36 & 95.32 & 95.34 & 95.34 & 95.34 \\
\bottomrule
\end{tabular}
\end{adjustbox}
\end{table*}

\end{document}